\RequirePackage{fix-cm}
\documentclass{article}
\usepackage{configuration/iclr2027_conference,times}

\usepackage{hyperref}
\usepackage{url}
\usepackage{xcolor}
\usepackage{booktabs}

\definecolor{color1}{HTML}{406058}
\definecolor{color2}{HTML}{78A0C8}
\definecolor{color3}{HTML}{3850A0}
\definecolor{color4}{HTML}{6860A8}
\definecolor{color5}{HTML}{A03850}
\definecolor{color6}{HTML}{AE7694}
\definecolor{color7}{HTML}{8488B4}
\definecolor{utAccent}{HTML}{006D77}
\hypersetup{
    pdfauthor={Deyuan Liu, Yihao Hu, Jingxuan Zhang, Xingying Li, Jun Xie, Jiacheng Liu, Jungang Li, Yu Huang, Xuanyi Liu, Yue Ding, Zecheng Wang, Lei Zhao, Mingda Wang, Zhenglin Cheng, Peng Sun, Tao Lin},
    pdftitle={UltraText Bench: A Comprehensive Bilingual Benchmark for Evaluating Visual Text Rendering in Image Generation},
    colorlinks=true,
    linkcolor=black,
    citecolor=color1,
    urlcolor=color2,
    pdfborder={0 0 0}
}
\usepackage[T1]{fontenc}
\usepackage{CJKutf8}

\usepackage{amsmath,amssymb,amsthm}

\providecommand{\mv}{\boldsymbol{v}}

\newenvironment{talign*}
{\csname align*\endcsname}
{\endalign}

\usepackage[utf8]{inputenc}
\usepackage[T1]{fontenc}
\usepackage{url}
\usepackage{booktabs}
\usepackage{tabularx}
\usepackage{amsfonts}
\usepackage{nicefrac}
\usepackage{microtype}
\usepackage{xcolor}
\usepackage{algorithm}
\usepackage{algorithmic}
\usepackage{graphicx}
\usepackage{subcaption}
\usepackage[flushleft]{threeparttable}
\usepackage{float}
\usepackage{multirow}
\usepackage{xspace}
\usepackage{natbib}
\usepackage{enumitem}
\usepackage[font=small,labelfont=bf,textfont=small,justification=raggedright,
  singlelinecheck=false,skip=3pt]{caption}
\usepackage{autobreak}
\usepackage{sidecap}
\usepackage{wrapfig}
\usepackage{bbding}
\usepackage[toc, page, header]{appendix}
\usepackage{tikz}
\usetikzlibrary{arrows.meta,calc,fit,matrix,patterns,positioning,shapes.geometric}

\definecolor{utEN}{HTML}{0072B2}
\definecolor{utZH}{HTML}{C45100}
\definecolor{utText}{HTML}{7B3294}
\definecolor{utSpatial}{HTML}{008837}
\definecolor{utLone}{HTML}{6B7782}
\definecolor{utLtwo}{HTML}{596B7A}
\definecolor{utLthree}{HTML}{34495E}
\definecolor{utGarble}{HTML}{B2182B}
\definecolor{utOmit}{HTML}{B8860B}
\definecolor{utPosition}{HTML}{CC79A7}
\definecolor{utTexture}{HTML}{8C564B}
\definecolor{utLegacy}{HTML}{6B6B6B}
\definecolor{utNeutral}{HTML}{3F3F3F}
\definecolor{utAxis}{HTML}{555555}
\definecolor{utGrid}{HTML}{D9DDE2}
\definecolor{utLight}{HTML}{F4F5F7}

\tikzset{
  ut/figure body/.style={font=\footnotesize},
  ut/figure label/.style={font=\scriptsize},
  ut/figure label bold/.style={font=\scriptsize\bfseries},
  ut/figure note/.style={font=\scriptsize, text=utNeutral},
  ut/box/.style={draw=utNeutral!55, rounded corners=2pt, fill=white,
    line width=0.7pt, inner sep=5pt, align=center},
  ut/document/.style={draw=utAccent!65, rounded corners=1pt, fill=white,
    line width=0.8pt, inner sep=5pt, align=left},
  ut/metadata/.style={draw=utNeutral!55, rounded corners=1pt, fill=utLight,
    line width=0.7pt, inner sep=5pt, align=left},
  ut/process/.style={draw=utNeutral!65, rounded corners=5pt, fill=white,
    line width=0.8pt, inner sep=5pt, align=center},
  ut/result/.style={draw=utAccent!65, rounded corners=2pt, fill=utAccent!7,
    line width=0.8pt, inner sep=5pt, align=center},
  ut/flow/.style={-{Latex[length=2.2mm,width=1.4mm]}, line width=0.9pt,
    draw=utNeutral},
  ut/axis/.style={-{Latex[length=1.8mm,width=1.2mm]}, line width=0.65pt,
    draw=utAxis},
  ut/tick/.style={line width=0.5pt, draw=utAxis},
  ut/grid/.style={line width=0.35pt, draw=utGrid},
  ut/ref/.style={densely dashed, line width=0.5pt, draw=utNeutral!60},
  ut/panel/.style={draw=utNeutral!35, rounded corners=2pt, fill=utLight,
    line width=0.7pt, inner sep=5pt},
  ut/current panel/.style={ut/panel, draw=utAccent!65, fill=utAccent!4},
  ut/legacy panel/.style={ut/panel, draw=utLegacy!75, fill=utLegacy!4,
    dash pattern=on 2.2pt off 1.2pt},
  ut/panel title/.style={font=\small\bfseries, text=utNeutral},
  ut/badge/.style={draw=utNeutral!45, rounded corners=2pt, fill=white,
    line width=0.55pt, inner xsep=4pt, inner ysep=2pt,
    ut/figure label bold, text=utNeutral},
  ut/current badge/.style={ut/badge, draw=utAccent!70, fill=utAccent!5},
  ut/legacy badge/.style={draw=utLegacy!80, rounded corners=2pt,
    fill=utLegacy!8, line width=0.65pt, dash pattern=on 2.2pt off 1.2pt,
    inner xsep=4pt, inner ysep=2pt, ut/figure label bold, text=utLegacy},
  ut/en/.style={draw=utEN, text=utEN, line width=1.2pt},
  ut/zh/.style={draw=utZH, text=utZH, line width=1.2pt, dashed},
  ut/en marker/.style={circle, draw=utEN, fill=utEN, inner sep=0pt,
    minimum size=4.4pt},
  ut/zh marker/.style={rectangle, draw=utZH, fill=utZH, inner sep=0pt,
    minimum size=4.4pt},
  ut/text mark/.style={draw=utText!85, fill=utText!12,
    pattern=north east lines, pattern color=utText!65},
  ut/spatial mark/.style={draw=utSpatial!85, fill=utSpatial!12,
    pattern=dots, pattern color=utSpatial!65},
  ut/difficulty l1/.style={draw=utLone, text=utLone, line width=0.8pt,
    densely dotted},
  ut/difficulty l2/.style={draw=utLtwo, text=utLtwo, line width=1.0pt,
    dashed},
  ut/difficulty l3/.style={draw=utLthree, text=utLthree, line width=1.2pt},
  ut/failure garble/.style={draw=utGarble, text=utGarble,
    line width=1.0pt},
  ut/failure omit/.style={draw=utOmit, text=utOmit,
    line width=1.0pt, dashed},
  ut/failure position/.style={draw=utPosition, text=utPosition,
    line width=1.0pt, dash dot},
  ut/failure texture/.style={draw=utTexture, text=utTexture,
    line width=1.0pt, densely dotted},
}
\usepackage{placeins}
\usepackage{xcolor}
\usepackage{pifont}
\usepackage{mdframed}
\usepackage{colortbl}

\definecolor{coral}{RGB}{255,127,80}
\definecolor{darkgreen}{RGB}{0,100,0}
\definecolor{darkyellow}{RGB}{204,153,0}
\definecolor{salmon}{RGB}{250,128,114}
\definecolor{bauhiniapurple}{RGB}{193,84,193}
\definecolor{kleinblue}{RGB}{0,47,167}
\definecolor{darkred}{RGB}{150,0,0}

\DeclareRobustCommand{\darkredtext}[1]{{\color{darkred}#1}}

\DeclareRobustCommand{\thmref}[1]{\hyperref[#1]{\darkredtext{Thm.~\ref*{#1}}}}
\DeclareRobustCommand{\defref}[1]{\hyperref[#1]{\darkredtext{Def.~\ref*{#1}}}}
\DeclareRobustCommand{\propref}[1]{\hyperref[#1]{\darkredtext{Prop.~\ref*{#1}}}}
\DeclareRobustCommand{\assumpref}[1]{\hyperref[#1]{\darkredtext{Assump.~\ref*{#1}}}}
\DeclareRobustCommand{\remarkref}[1]{\hyperref[#1]{\darkredtext{Rem.~\ref*{#1}}}}
\DeclareRobustCommand{\hypref}[1]{\hyperref[#1]{\darkredtext{Hyp.~\ref*{#1}}}}
\DeclareRobustCommand{\conjref}[1]{\hyperref[#1]{\darkredtext{Conj.~\ref*{#1}}}}
\DeclareRobustCommand{\lemref}[1]{\hyperref[#1]{\darkredtext{Lem.~\ref*{#1}}}}
\DeclareRobustCommand{\corref}[1]{\hyperref[#1]{\darkredtext{Cor.~\ref*{#1}}}}
\DeclareRobustCommand{\noteref}[1]{\hyperref[#1]{\darkredtext{Nota.~\ref*{#1}}}}
\DeclareRobustCommand{\claimref}[1]{\hyperref[#1]{\darkredtext{Clm.~\ref*{#1}}}}
\DeclareRobustCommand{\obsref}[1]{\hyperref[#1]{\darkredtext{Obs.~\ref*{#1}}}}

\DeclareRobustCommand{\transparentteal}[1]{\tikz[baseline=(X.base)] \node[fill=teal, fill opacity=0.1, text opacity=1, inner sep=2pt, outer sep=0pt] (X) {#1};}
\DeclareRobustCommand{\figref}[1]{\hyperref[#1]{\transparentteal{Figure~\ref*{#1}}}}

\DeclareRobustCommand{\transparentdarkgreen}[1]{\tikz[baseline=(X.base)] \node[fill=darkgreen, fill opacity=0.1, text opacity=1, inner sep=2pt, outer sep=0pt] (X) {#1};}
\DeclareRobustCommand{\tabref}[1]{\hyperref[#1]{\transparentdarkgreen{Table~\ref*{#1}}}}

\DeclareRobustCommand{\transparentdarkyellow}[1]{\tikz[baseline=(X.base)] \node[fill=darkyellow, fill opacity=0.1, text opacity=1, inner sep=2pt, outer sep=0pt] (X) {#1};}
\DeclareRobustCommand{\secref}[1]{\hyperref[#1]{\transparentdarkyellow{Section~\ref*{#1}}}}

\DeclareRobustCommand{\transparentcoral}[1]{\tikz[baseline=(X.base)] \node[fill=coral, fill opacity=0.1, text opacity=1, inner sep=2pt, outer sep=0pt] (X) {#1};}
\DeclareRobustCommand{\appref}[1]{\hyperref[#1]{\transparentcoral{Appendix~\ref*{#1}}}}

\DeclareRobustCommand{\transparentkleinblue}[1]{\tikz[baseline=(X.base)] \node[fill=kleinblue, fill opacity=0.1, text opacity=1, inner sep=2pt, outer sep=0pt] (X) {#1};}
\DeclareRobustCommand{\algref}[1]{\hyperref[#1]{\transparentkleinblue{Algorithm~\ref*{#1}}}}

\DeclareRobustCommand{\transparentbauhiniapurple}[1]{\tikz[baseline=(X.base)] \node[fill=bauhiniapurple, fill opacity=0.1, text opacity=1, inner sep=2pt, outer sep=0pt] (X) {#1};}
\DeclareRobustCommand{\equref}[1]{\hyperref[#1]{\transparentbauhiniapurple{Equation~\ref*{#1}}}}

\newtheoremstyle{custom}
{1pt}
{1pt}
{\itshape}
{}
{\bfseries}
{}
{ }
{\thmname{#1} \thmnumber{#2} \thmnote{(#3)} . }

\theoremstyle{custom}

\newtheorem{innerdefinition}{Definition}
\newtheorem{innerproposition}{Proposition}
\newtheorem{innerassumption}{Assumption}
\newtheorem{innerremark}{Remark}
\newtheorem{innertheorem}{Theorem}
\newtheorem{innerhypothesis}{Hypothesis}
\newtheorem{innerconjecture}{Conjecture}
\newtheorem{innerlemma}{Lemma}
\newtheorem{innercorollary}{Corollary}
\newtheorem{innernotation}{Notation}
\newtheorem{innerclaim}{Claim}
\newtheorem{innerproblem}{Problem}

\newtheorem{innerobservation}{Observation}

\newmdenv[
  backgroundcolor=gray!10,
  linecolor=gray!100,
  linewidth=0.8pt,
  skipabove=2pt,
  skipbelow=2pt,
  innertopmargin=10pt,
  innerbottommargin=5pt,
  innerleftmargin=5pt,
  innerrightmargin=5pt,
]{definitionframe}

\newmdenv[
  backgroundcolor=blue!10,
  linecolor=blue!100,
  linewidth=0.8pt,
  skipabove=2pt,
  skipbelow=2pt,
  innertopmargin=10pt,
  innerbottommargin=5pt,
  innerleftmargin=5pt,
  innerrightmargin=5pt,
]{propositionframe}

\newmdenv[
  backgroundcolor=green!10,
  linecolor=green!100,
  linewidth=0.8pt,
  skipabove=2pt,
  skipbelow=2pt,
  innertopmargin=10pt,
  innerbottommargin=5pt,
  innerleftmargin=5pt,
  innerrightmargin=5pt,
]{assumptionframe}

\newmdenv[
  backgroundcolor=yellow!10,
  linecolor=yellow!100,
  linewidth=0.8pt,
  skipabove=2pt,
  skipbelow=2pt,
  innertopmargin=10pt,
  innerbottommargin=5pt,
  innerleftmargin=5pt,
  innerrightmargin=5pt,
]{remarkframe}

\newmdenv[
  backgroundcolor=red!10,
  linecolor=red!100,
  linewidth=0.8pt,
  skipabove=2pt,
  skipbelow=2pt,
  innertopmargin=10pt,
  innerbottommargin=5pt,
  innerleftmargin=5pt,
  innerrightmargin=5pt,
]{theoremframe}

\newmdenv[
  backgroundcolor=purple!10,
  linecolor=purple!100,
  linewidth=0.8pt,
  skipabove=2pt,
  skipbelow=2pt,
  innertopmargin=10pt,
  innerbottommargin=5pt,
  innerleftmargin=5pt,
  innerrightmargin=5pt,
]{hypothesisframe}

\newmdenv[
  backgroundcolor=orange!10,
  linecolor=orange!100,
  linewidth=0.8pt,
  skipabove=2pt,
  skipbelow=2pt,
  innertopmargin=10pt,
  innerbottommargin=5pt,
  innerleftmargin=5pt,
  innerrightmargin=5pt,
]{conjectureframe}

\newmdenv[
  backgroundcolor=cyan!10,
  linecolor=cyan!100,
  linewidth=0.8pt,
  skipabove=2pt,
  skipbelow=2pt,
  innertopmargin=10pt,
  innerbottommargin=5pt,
  innerleftmargin=5pt,
  innerrightmargin=5pt,
]{lemmaframe}

\newmdenv[
  backgroundcolor=magenta!10,
  linecolor=magenta!100,
  linewidth=0.8pt,
  skipabove=2pt,
  skipbelow=2pt,
  innertopmargin=10pt,
  innerbottommargin=5pt,
  innerleftmargin=5pt,
  innerrightmargin=5pt,
]{corollaryframe}

\newmdenv[
  backgroundcolor=pink!10,
  linecolor=pink!100,
  linewidth=0.8pt,
  skipabove=2pt,
  skipbelow=2pt,
  innertopmargin=10pt,
  innerbottommargin=5pt,
  innerleftmargin=5pt,
  innerrightmargin=5pt,
]{notationframe}

\newmdenv[
  backgroundcolor=violet!10,
  linecolor=violet!100,
  linewidth=0.8pt,
  skipabove=2pt,
  skipbelow=2pt,
  innertopmargin=10pt,
  innerbottommargin=5pt,
  innerleftmargin=5pt,
  innerrightmargin=5pt,
]{claimframe}

\newmdenv[
  backgroundcolor=salmon!10,
  linecolor=salmon!100,
  linewidth=0.8pt,
  skipabove=2pt,
  skipbelow=2pt,
  innertopmargin=10pt,
  innerbottommargin=5pt,
  innerleftmargin=5pt,
  innerrightmargin=5pt,
]{problemframe}

\newmdenv[
  backgroundcolor=lavender!10,
  linecolor=lavender!100,
  linewidth=0.8pt,
  skipabove=2pt,
  skipbelow=2pt,
  innertopmargin=10pt,
  innerbottommargin=5pt,
  innerleftmargin=5pt,
  innerrightmargin=5pt,
]{observationframe}

\usepackage{fvextra}
\usepackage{needspace}
\DefineVerbatimEnvironment{UTCode}{Verbatim}{fontsize=\footnotesize,
  breaklines=true,breakanywhere=false,breakafter={,},frame=single,framesep=5pt,
  formatcom=\color{black},
  framerule=0.3pt,rulecolor=\color{utGrid},tabsize=2}


\title{UltraText Bench: A Comprehensive Bilingual Benchmark for Evaluating Visual Text Rendering in Image Generation}

\author{\begin{minipage}{\dimexpr\textwidth-2\tabcolsep\relax}
  \centering\small
  Deyuan Liu$^{1,\dagger}$ \,
  Yihao Hu$^{1,2,\dagger}$ \,
  Jingxuan Zhang$^{1,\dagger}$ \,
  Xingying Li$^{1,\dagger}$ \,
  Jun Xie$^{1,3,4,\dagger}$ \\
  \bf
  Jiacheng Liu$^{8}$ \,
  Jungang Li$^{5}$ \,
  Yu Huang$^{9}$ \,
  Xuanyi Liu$^{10}$ \,
  Yue Ding$^{6}$ \,
  Zecheng Wang$^{7}$ \\
  \bf
  Lei Zhao$^{1}$ \,
  Mingda Wang$^{1}$ \,
  Zhenglin Cheng$^{1,3,4}$ \,
  Peng Sun$^{1,3}$ \,
  \textbf{Tao Lin}$^{1}$\thanks{Corresponding author. \quad $^{\dagger}$Equal contribution.} \\ [0.5ex]
  \footnotesize\normalfont
  $^{1}$Westlake University \quad
  $^{2}$Ant Group \quad
  $^{3}$Zhejiang University \quad
  $^{4}$Shanghai Innovation Institute \\
  $^{5}$HKUST \quad
  $^{6}$CASIA \quad
  $^{7}$Wechat AI \quad
  $^{8}$MBZUAI \quad
  $^{9}$CityU \quad
  $^{10}$Peking University
  \end{minipage}
}

\iclrpreprintcopy
\begin{document}
\begin{CJK*}{UTF8}{gbsn}

\maketitle
\begin{abstract}
  Dense visual text requires image generators to reproduce long strings across multiple regions with correct placement and legibility.
  As short-string rendering improves, evaluation must test sustained performance across more demanding scenes.
  We introduce \textcolor{color3}{UltraText Bench}, a bilingual benchmark for prompt-only generation of dense visual text.
  It contains 432 prompts spanning 24 real-world scene categories and three difficulty levels, split equally between English and Chinese.
  Each human-reviewed prompt supplies exact strings for four to twelve text regions, paired with structured references for their content, placement, and visual attributes.
  We use the Q-Judger vision-language model to assess each image against the complete reference, reporting text fidelity, text clarity, spatial quality, and scene quality.
  Across 24 model configurations, these dimensions reveal different strengths: Z-Image-Turbo gains 3.81 clarity points over Z-Image-Base while losing 14.76 fidelity points under the reported settings.
  Performance also varies with workload; Qwen-Image-2512's English composite falls from 86.50 at L1 to 42.86 at L3.
  Ten participants took part in human evaluation of the automatic scores.
  Repository: \url{https://github.com/LINs-lab/UltraText_Bench}.
\end{abstract}

\setlength{\parskip}{1pt plus1pt minus1pt}

\section{Introduction}
\label{sec:intro}

\begin{figure}[!t]
  \centering
  \setlength{\tabcolsep}{2pt}
  \begin{tabular}{@{}cccc@{}}
    \includegraphics[width=0.205\textwidth]{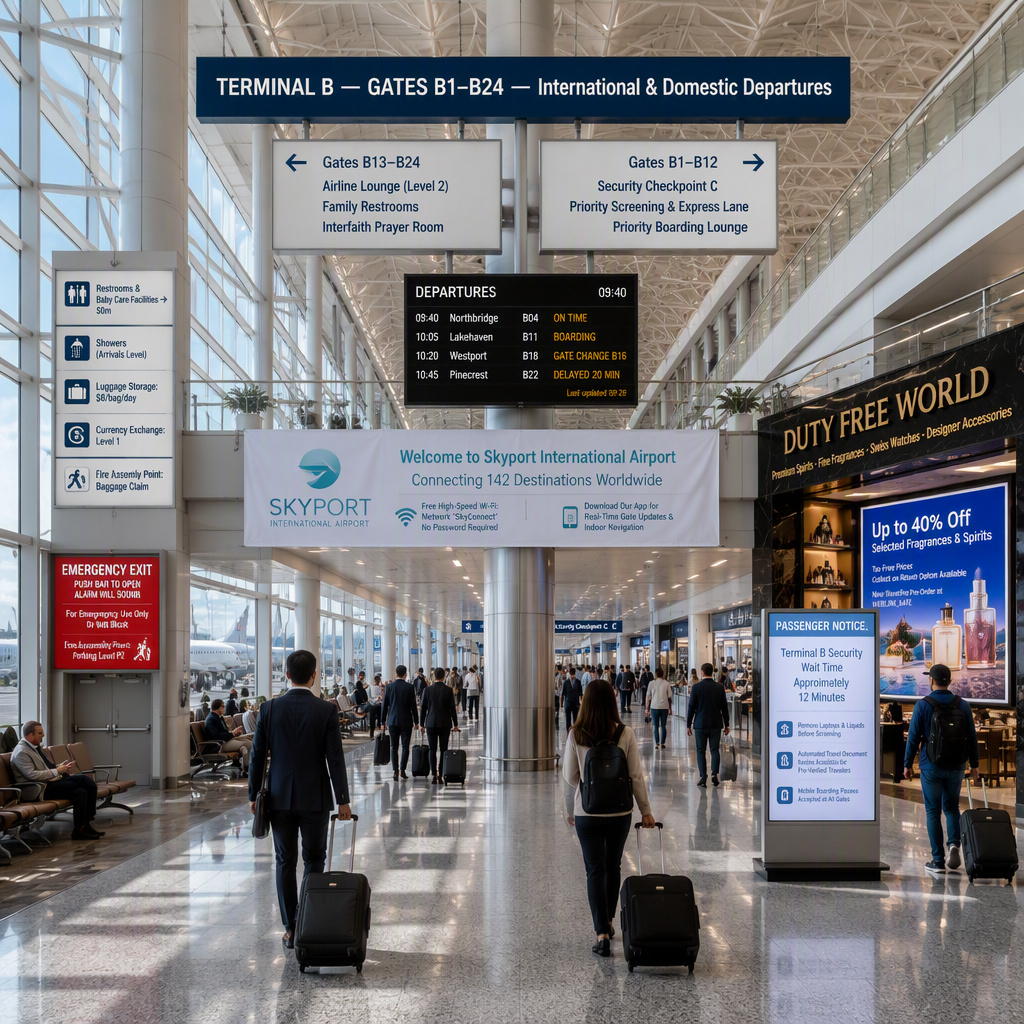} &
    \includegraphics[width=0.205\textwidth]{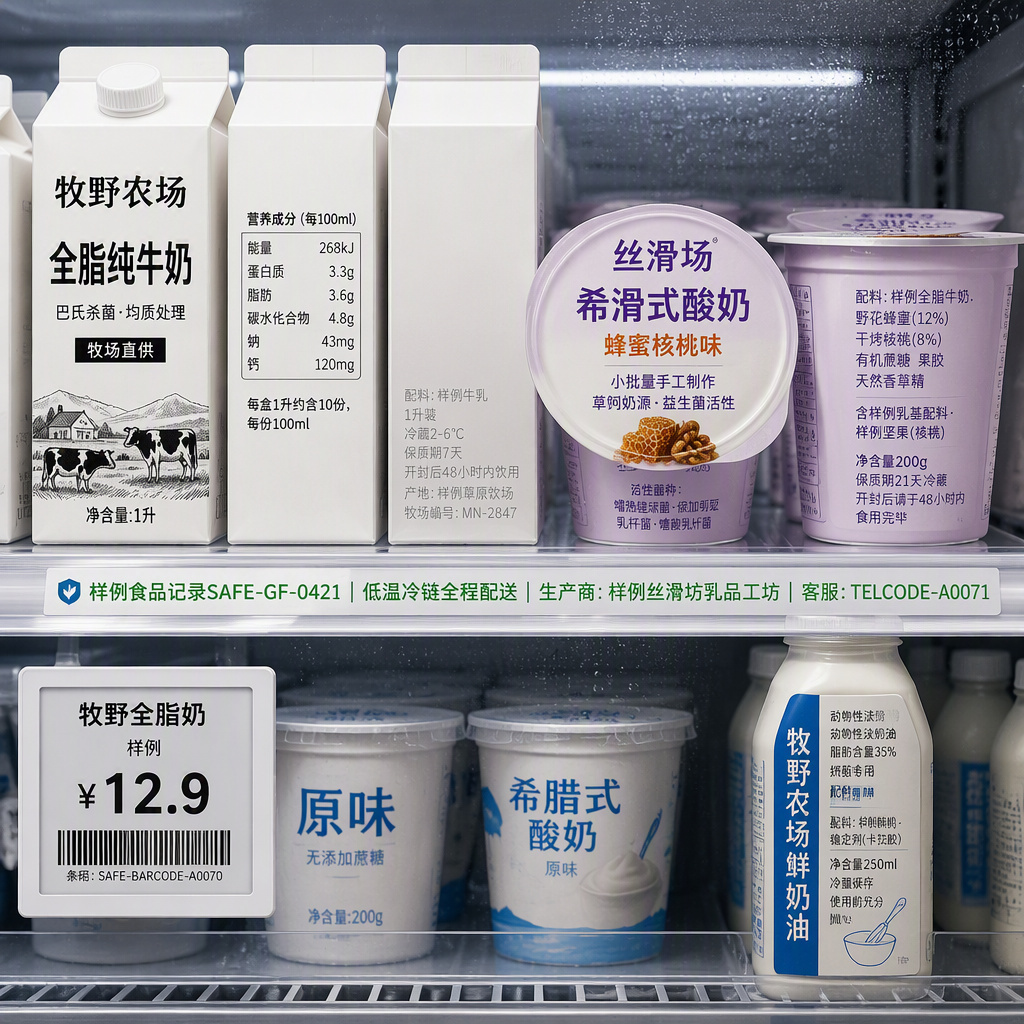} &
    \includegraphics[width=0.205\textwidth]{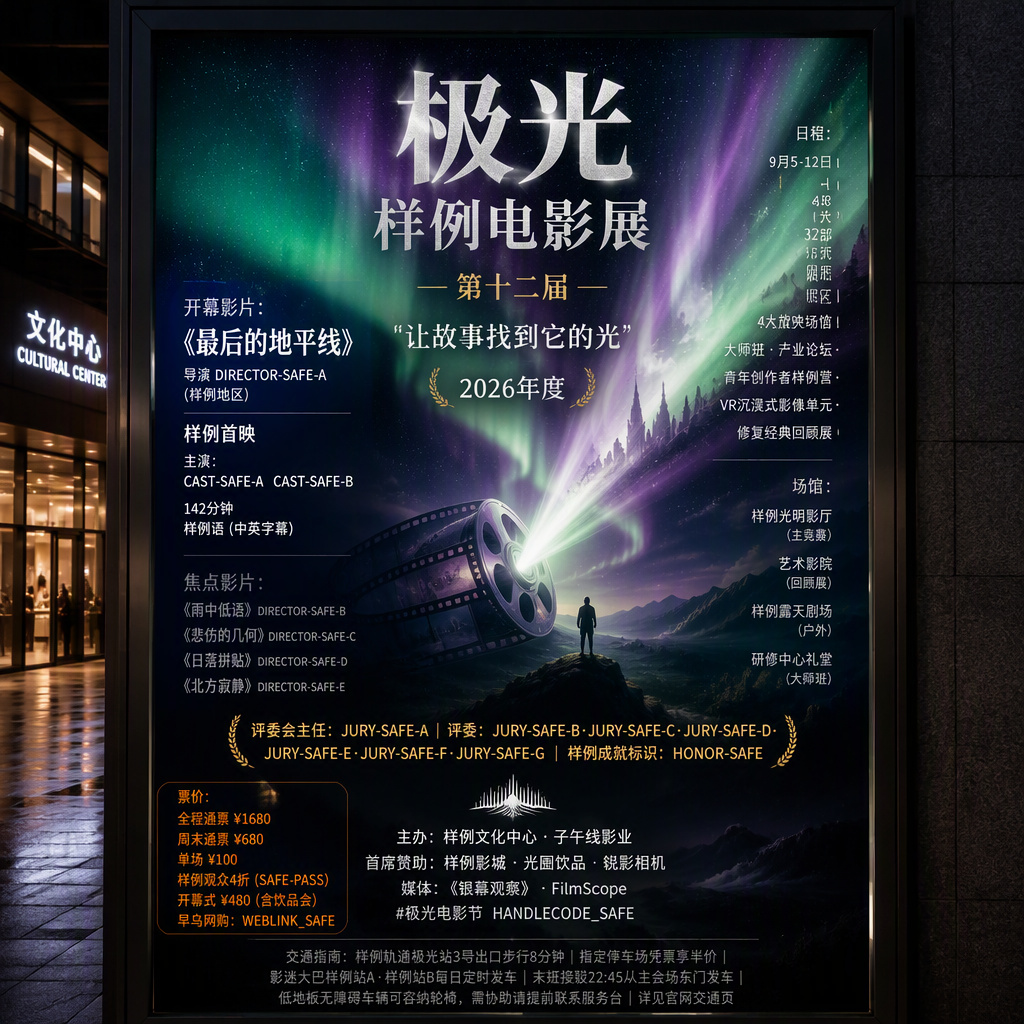} &
    \includegraphics[width=0.205\textwidth]{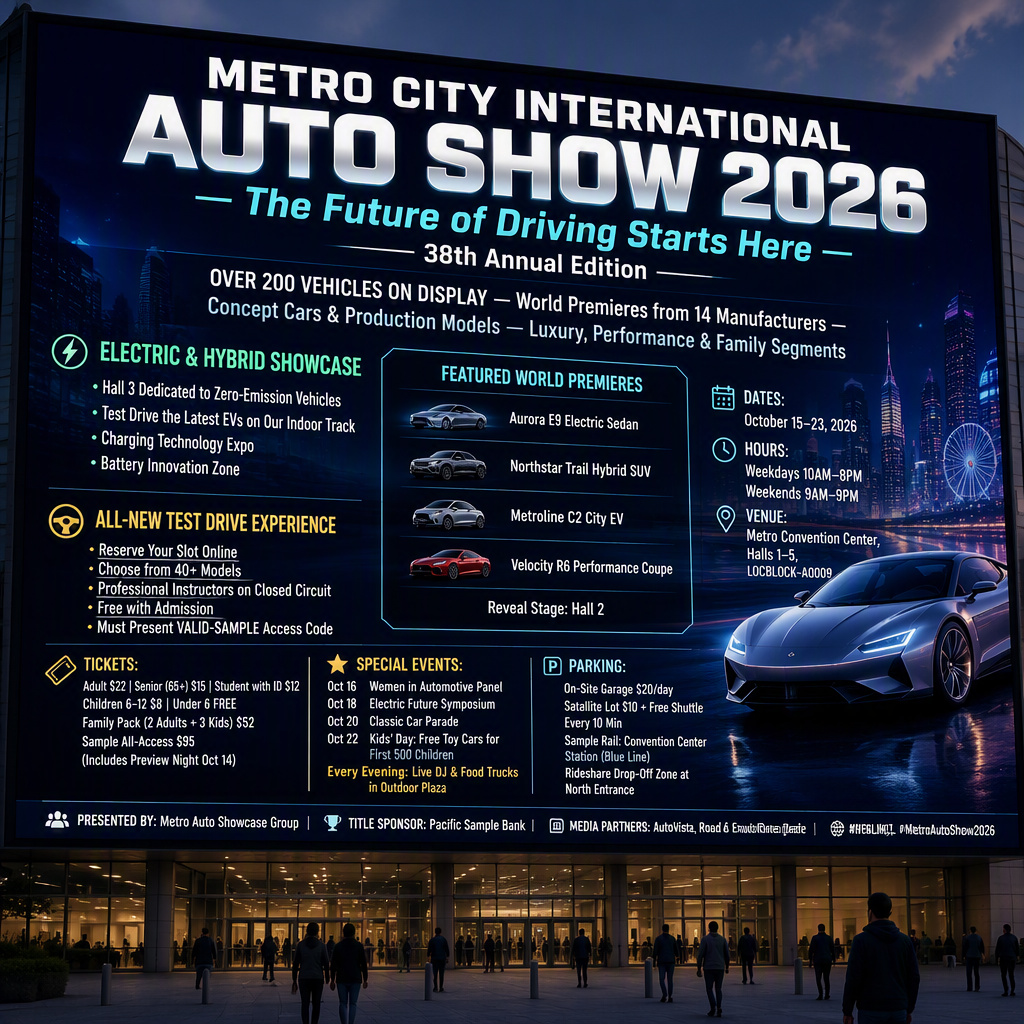} \\
    \scriptsize Sign (EN/L3) & \scriptsize Label (ZH/L3) & \scriptsize Poster (ZH/L3) & \scriptsize Billboard (EN/L3) \\[1pt]
    \includegraphics[width=0.205\textwidth]{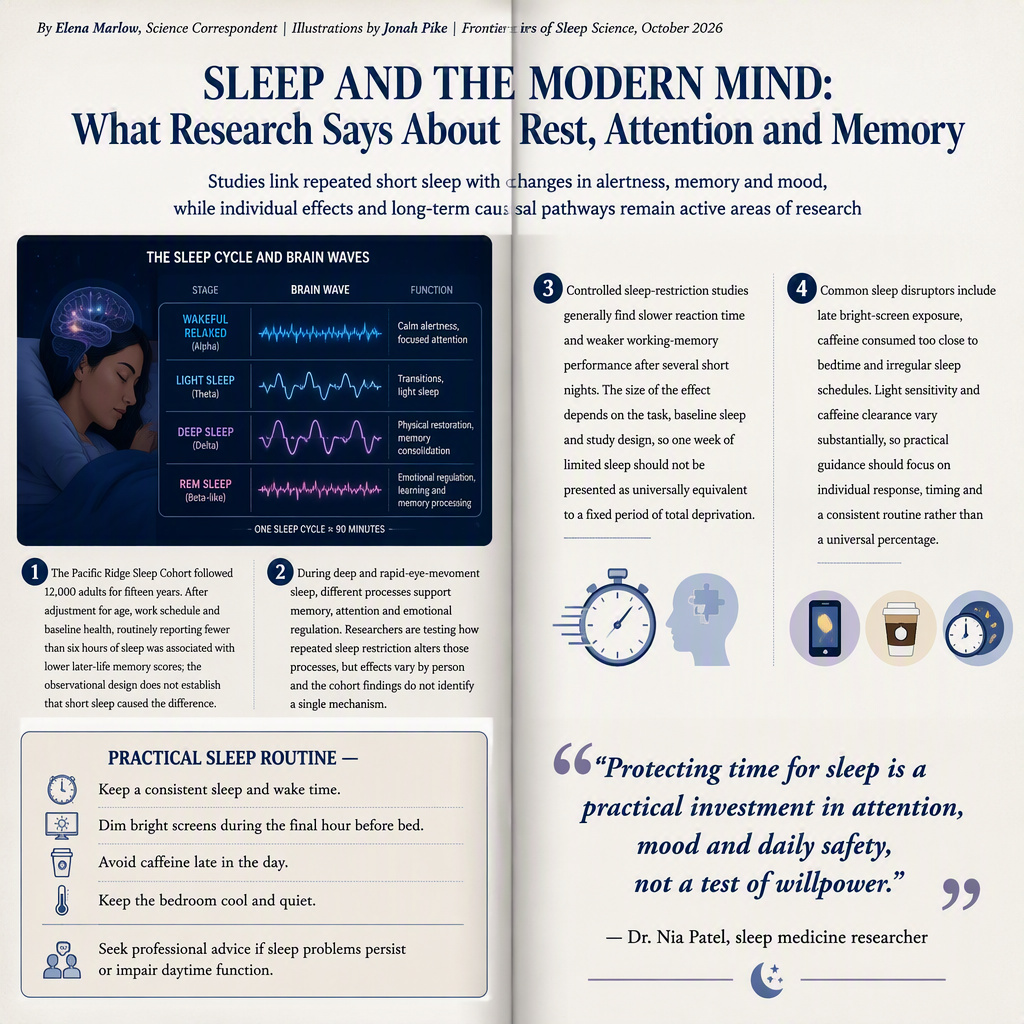} &
    \includegraphics[width=0.205\textwidth]{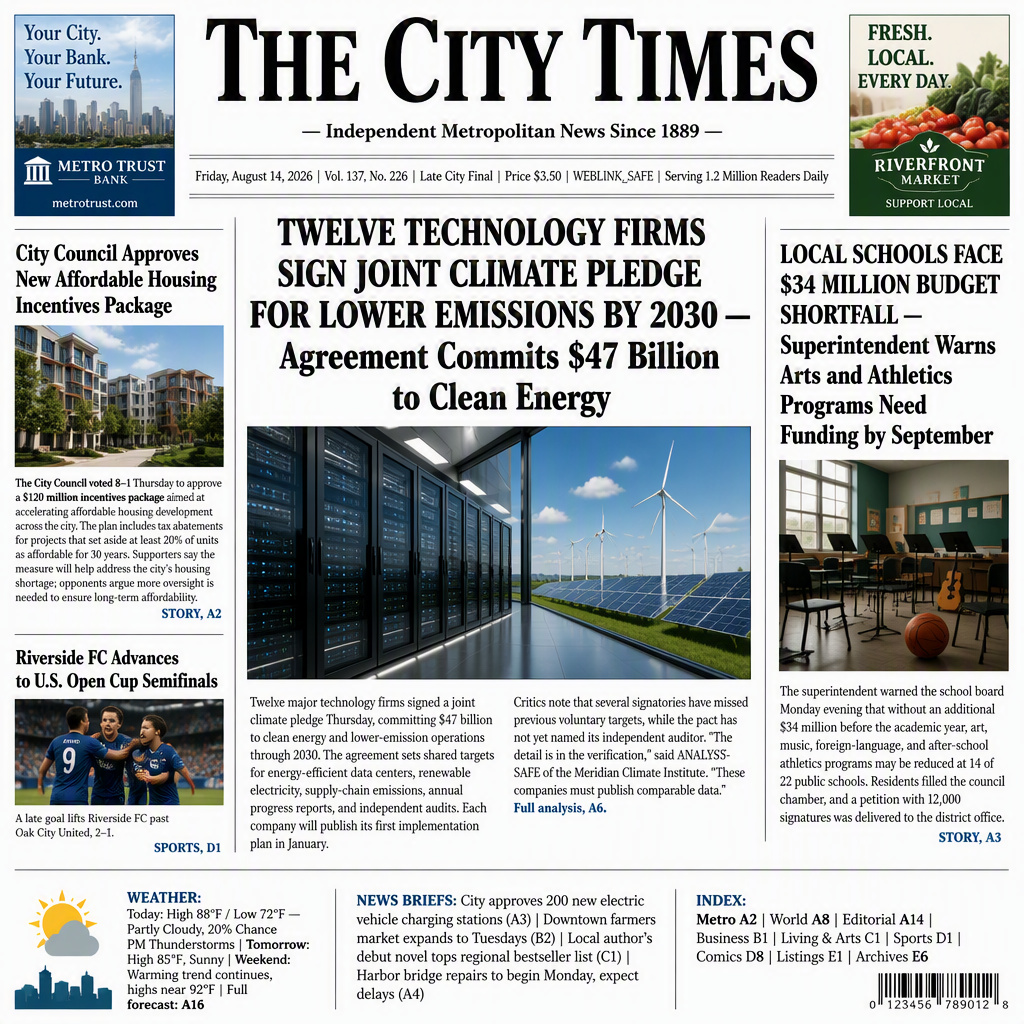} &
    \includegraphics[width=0.205\textwidth]{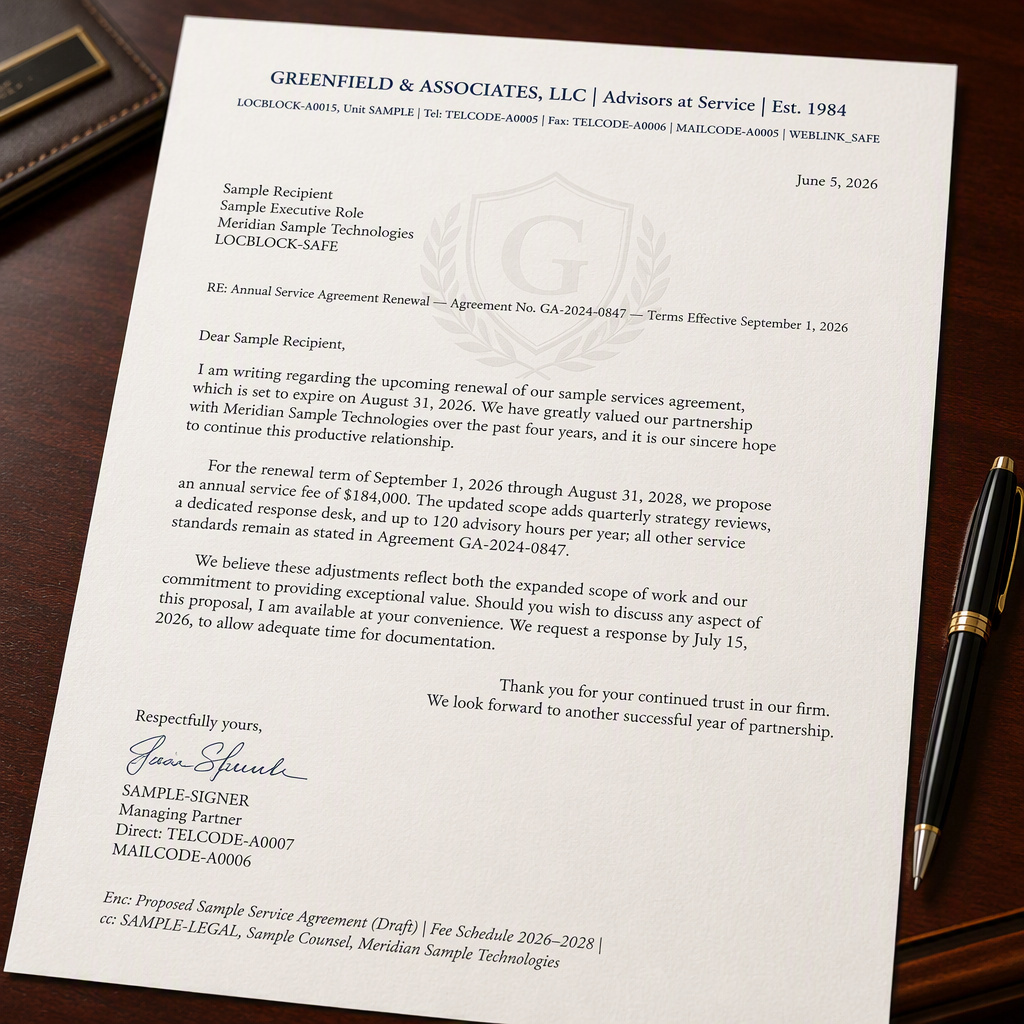} &
    \includegraphics[width=0.205\textwidth]{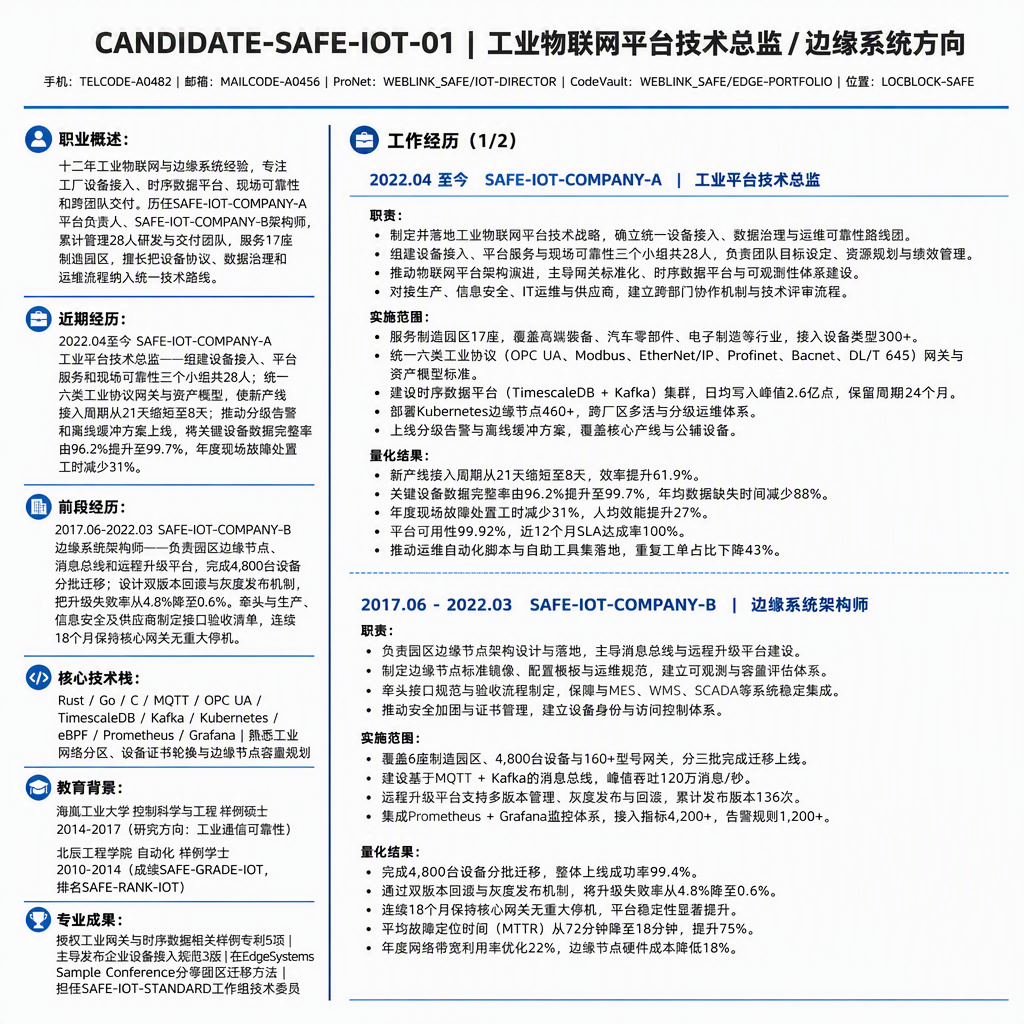} \\
    \scriptsize Article (EN/L3) & \scriptsize Newspaper (EN/L3) & \scriptsize Letter (EN/L3) & \scriptsize Resume (ZH/L3) \\[2pt]
    \includegraphics[width=0.205\textwidth]{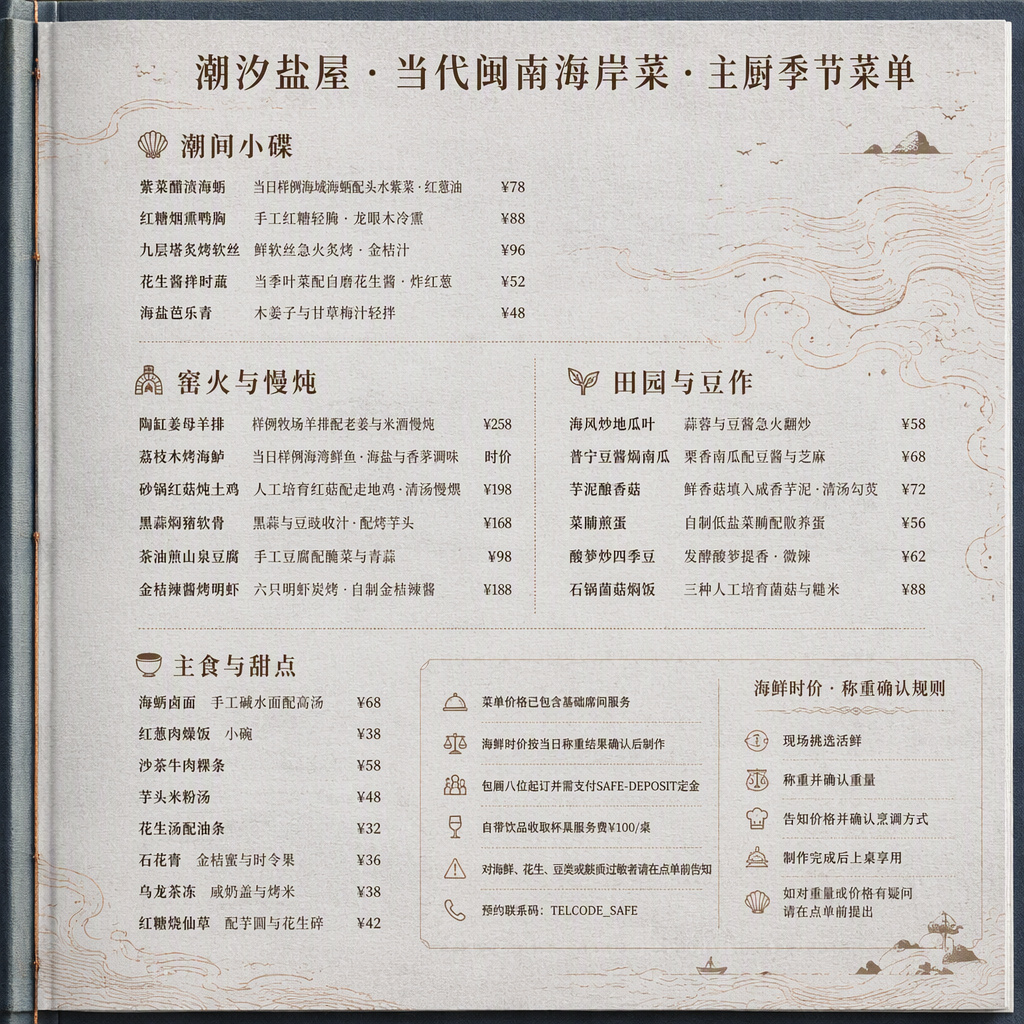} &
    \includegraphics[width=0.205\textwidth]{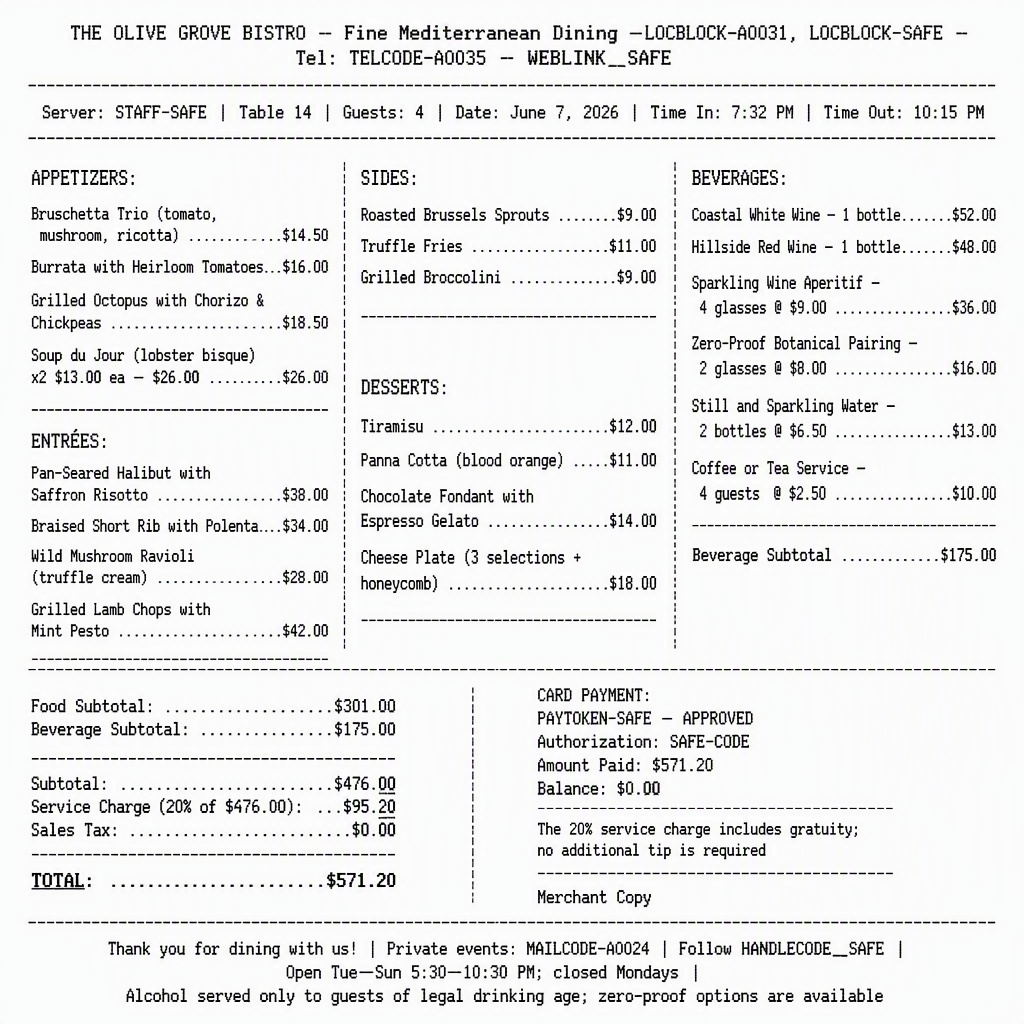} &
    \includegraphics[width=0.205\textwidth]{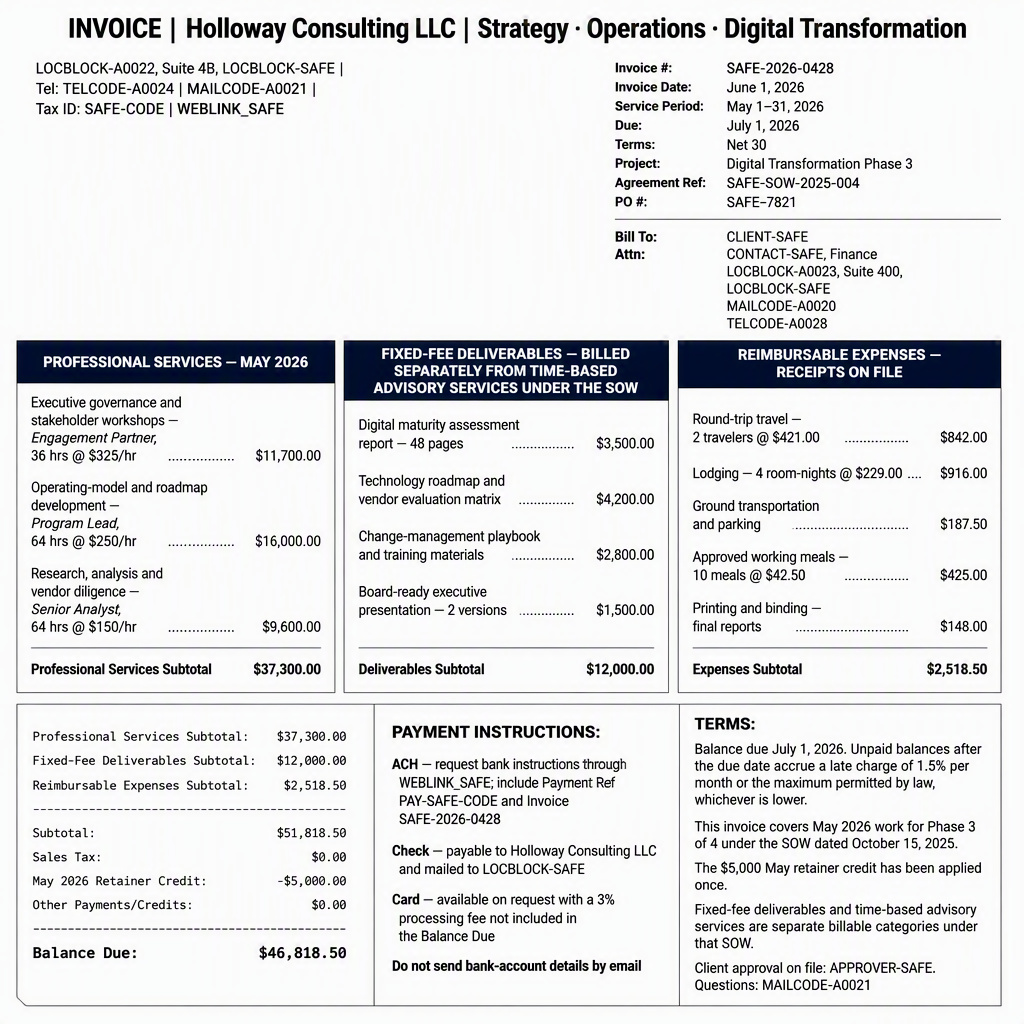} &
    \includegraphics[width=0.205\textwidth]{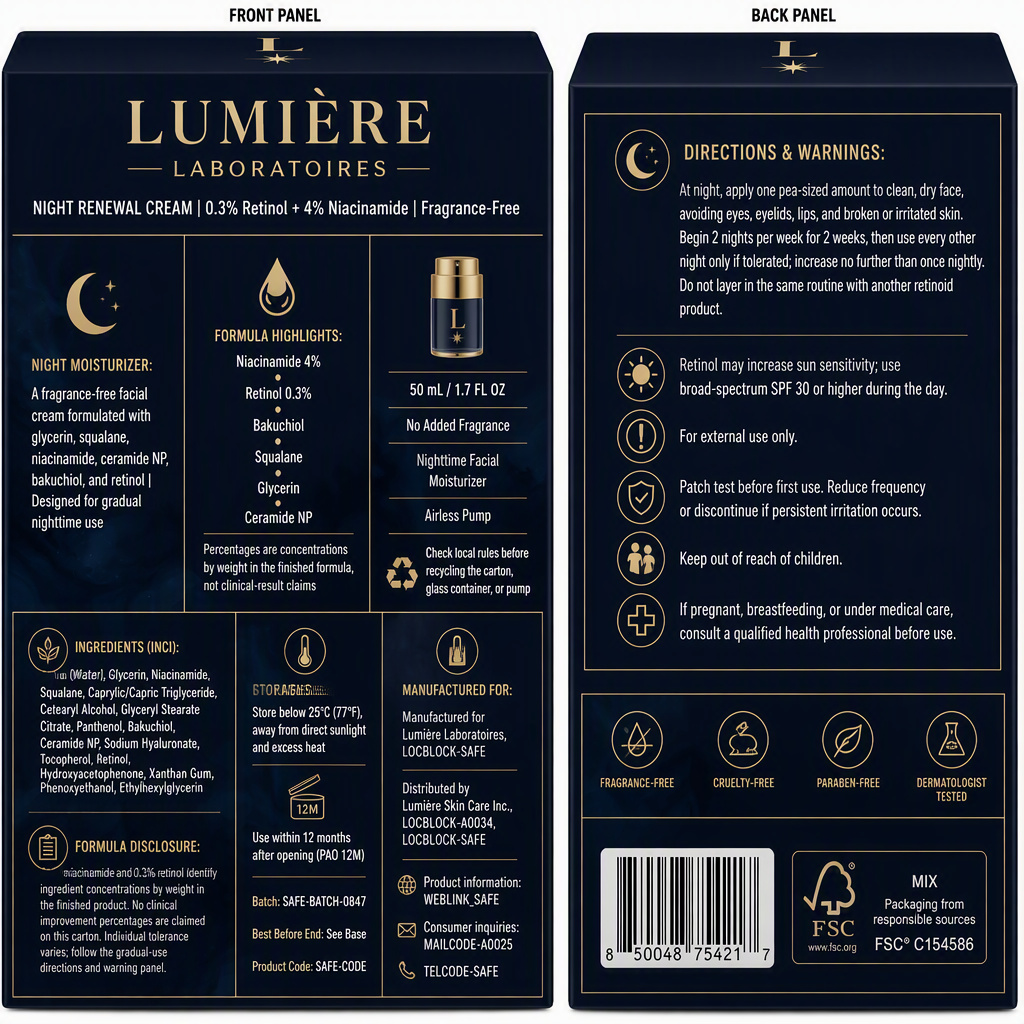} \\
    \scriptsize Menu (ZH/L3) & \scriptsize Receipt (EN/L3) & \scriptsize Invoice (EN/L3) & \scriptsize Packaging (EN/L3) \\[2pt]
    \includegraphics[width=0.205\textwidth]{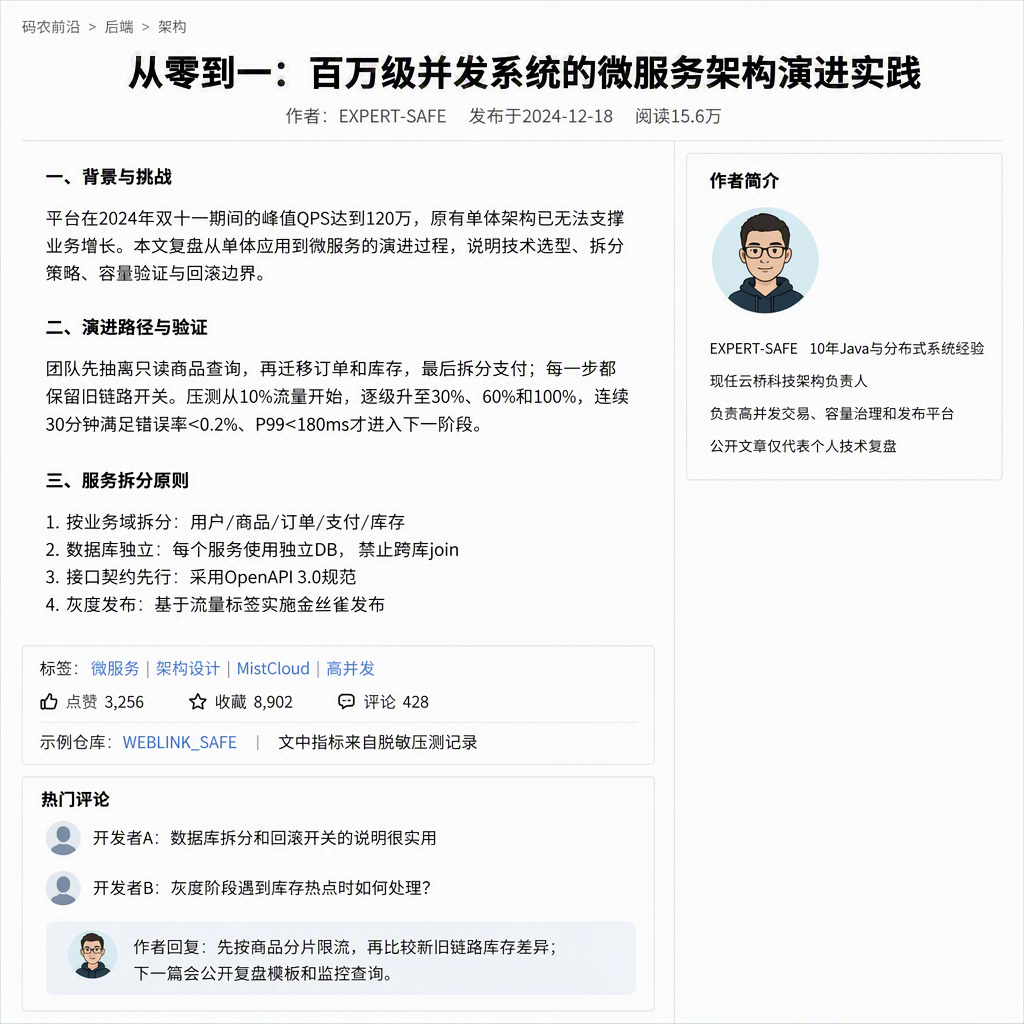} &
    \includegraphics[width=0.205\textwidth]{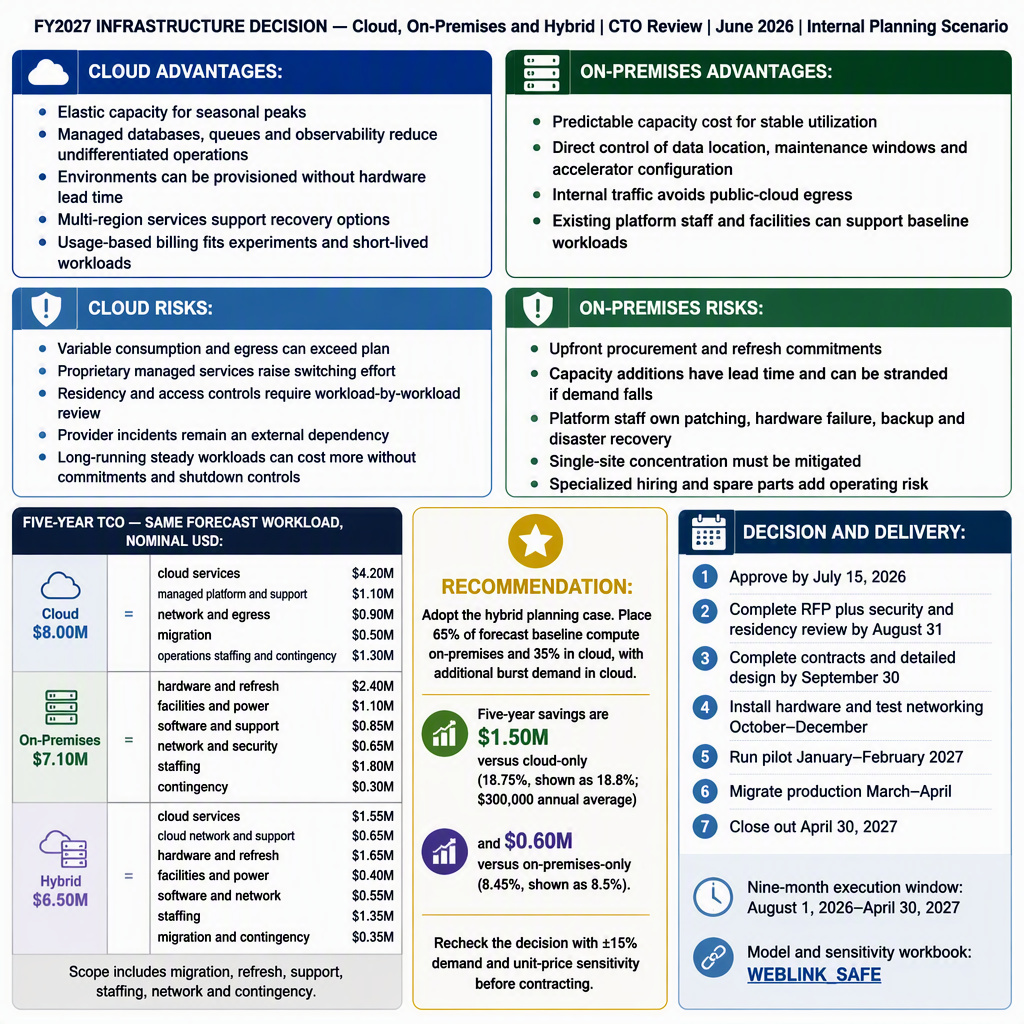} &
    \includegraphics[width=0.205\textwidth]{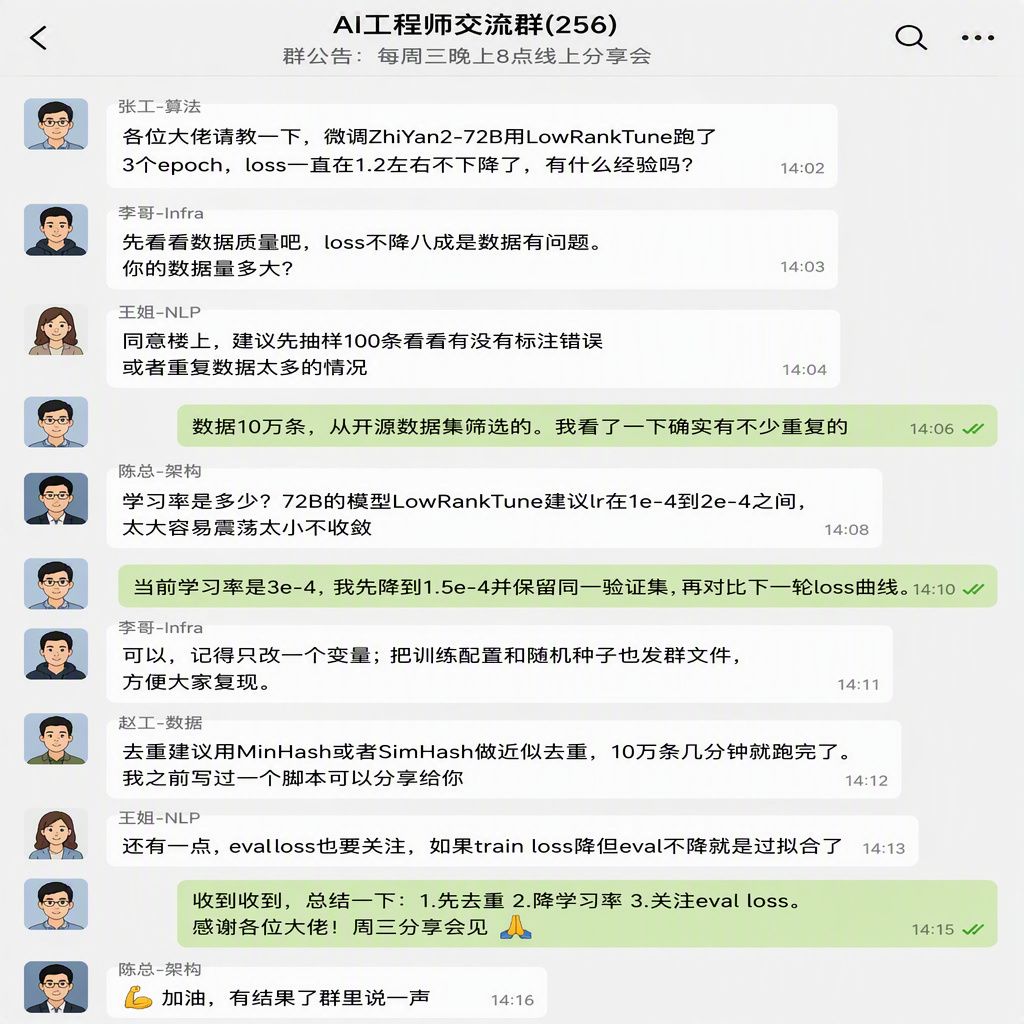} &
    \includegraphics[width=0.205\textwidth]{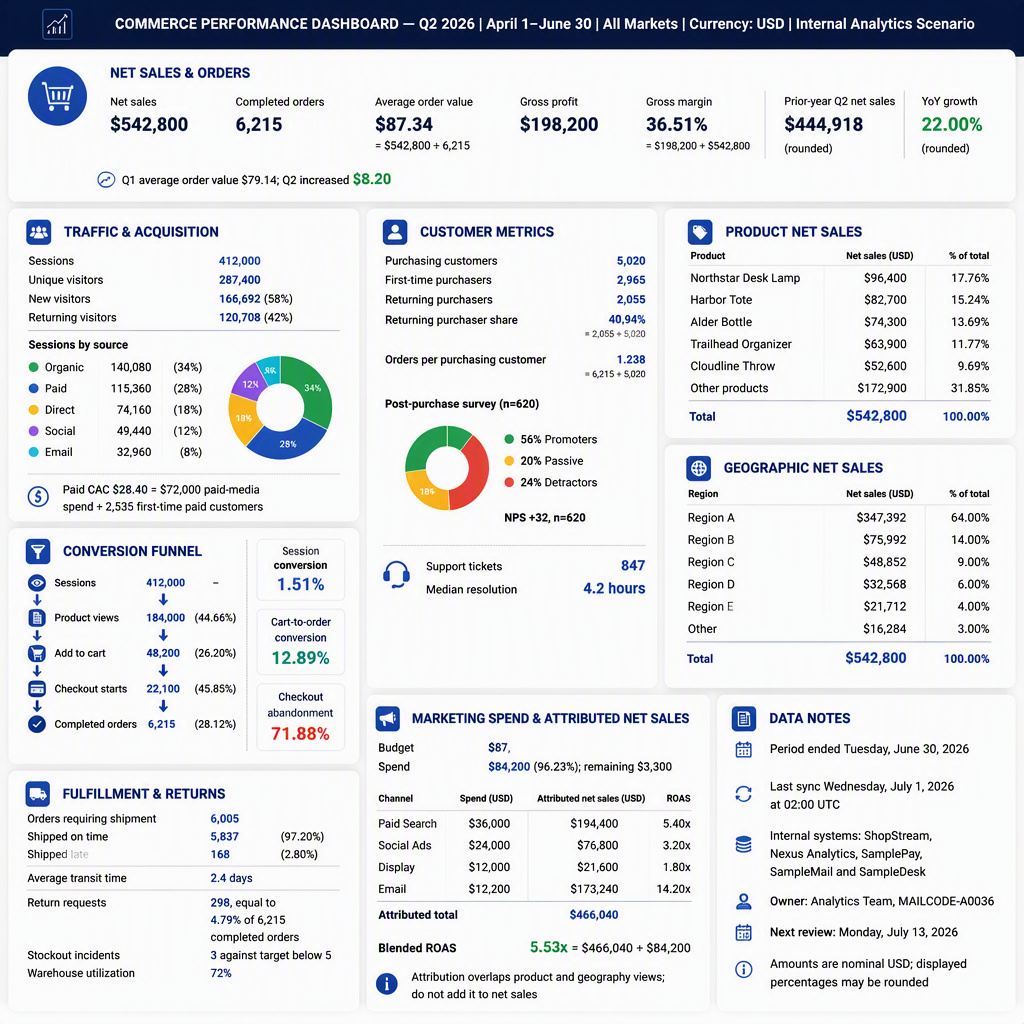} \\
    \scriptsize Webpage (ZH/L3) & \scriptsize Slide (EN/L3) & \scriptsize Social Media (ZH/L3) & \scriptsize Dashboard (EN/L3) \\[2pt]
    \includegraphics[width=0.205\textwidth]{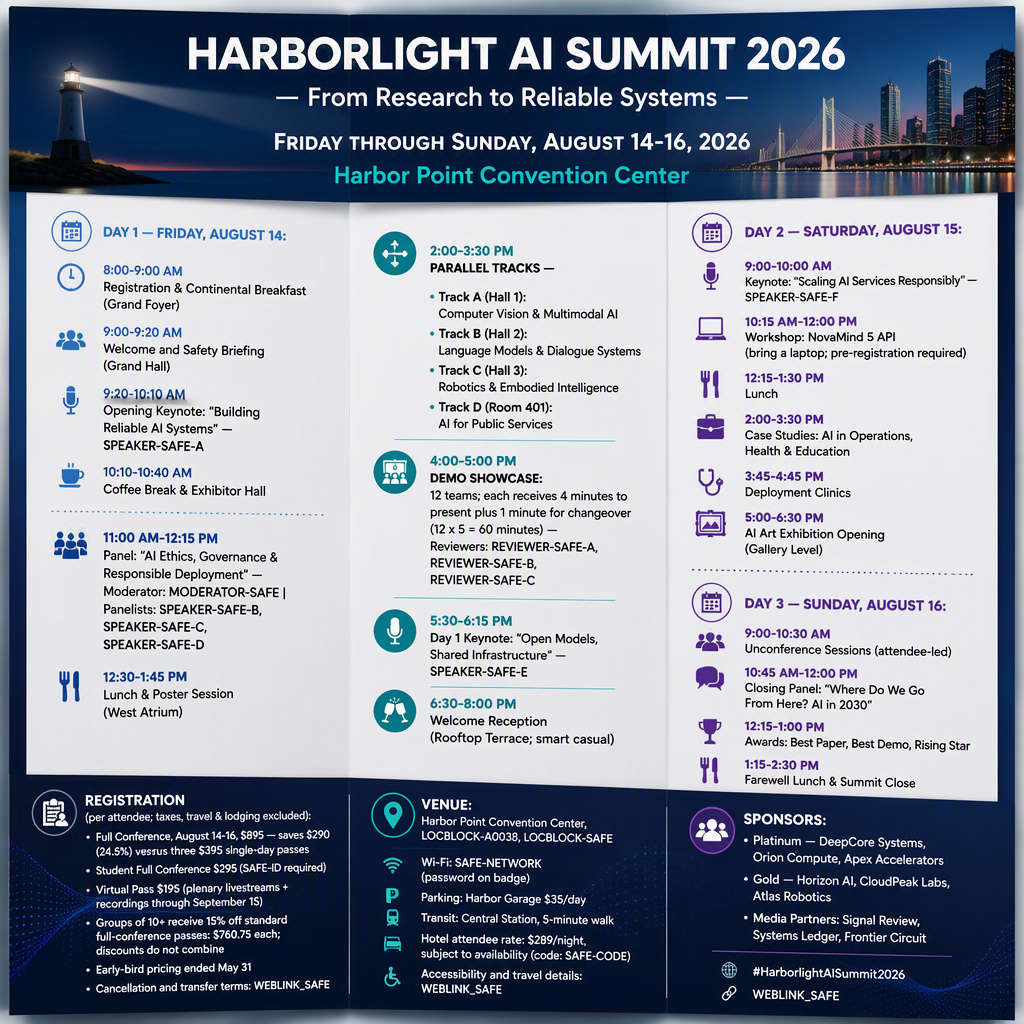} &
    \includegraphics[width=0.205\textwidth]{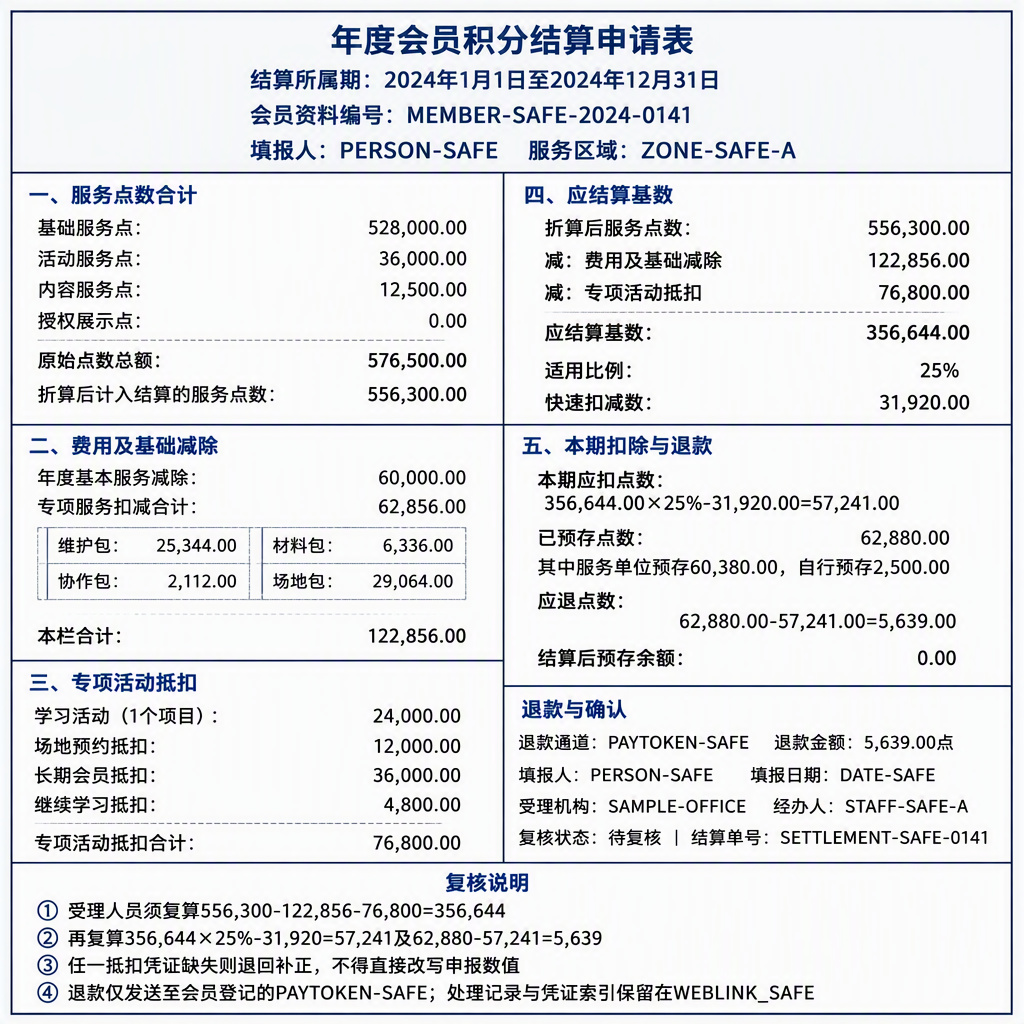} &
    \includegraphics[width=0.205\textwidth]{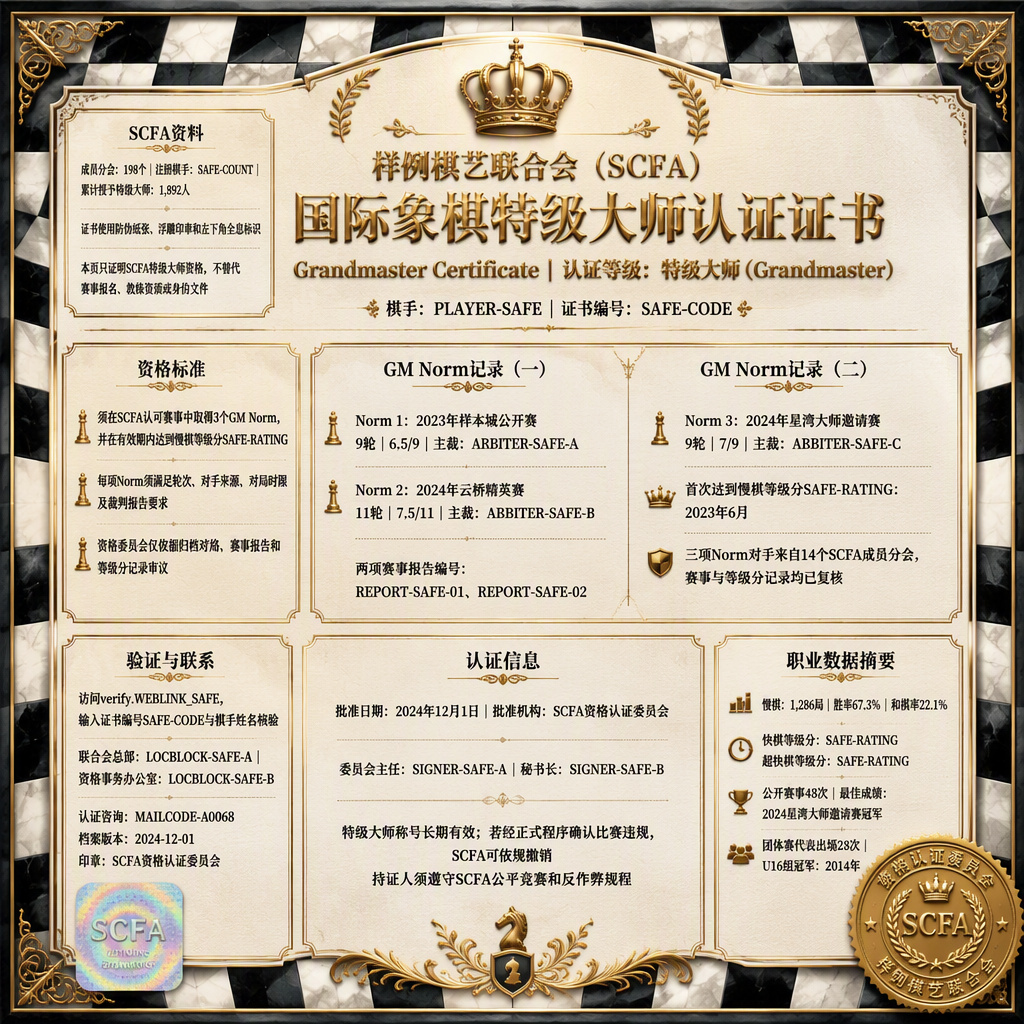} &
    \includegraphics[width=0.205\textwidth]{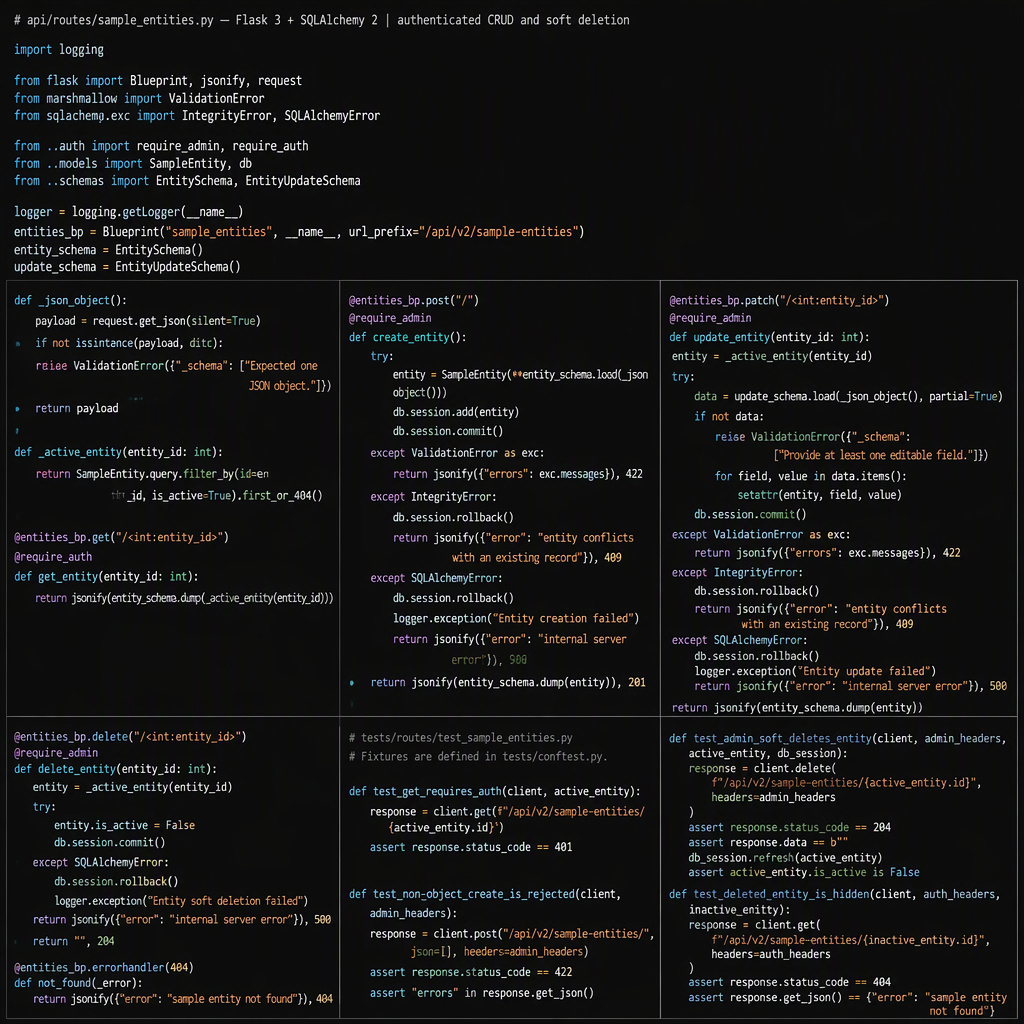} \\
    \scriptsize Schedule (EN/L3) & \scriptsize Form (ZH/L3) & \scriptsize Certificate (ZH/L3) & \scriptsize Code (EN/L3) \\[2pt]
    \includegraphics[width=0.205\textwidth]{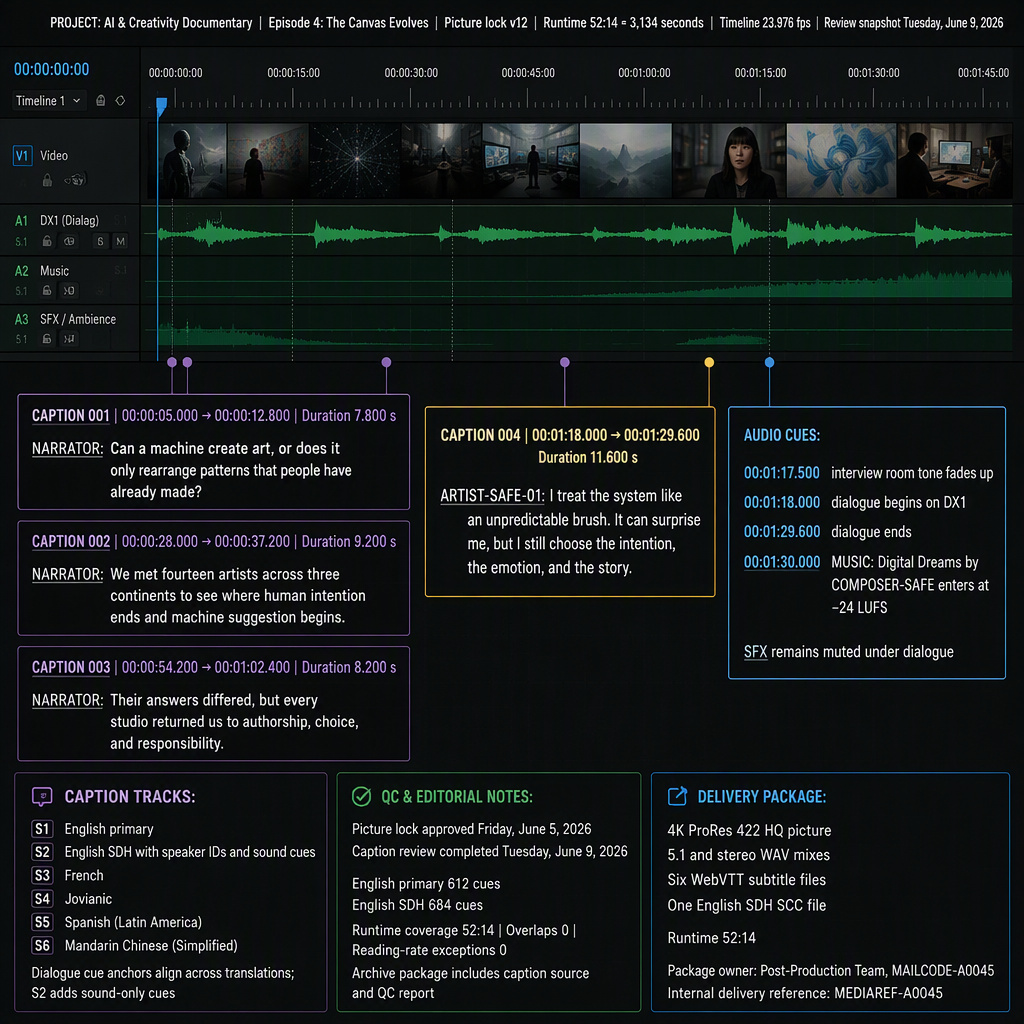} &
    \includegraphics[width=0.205\textwidth]{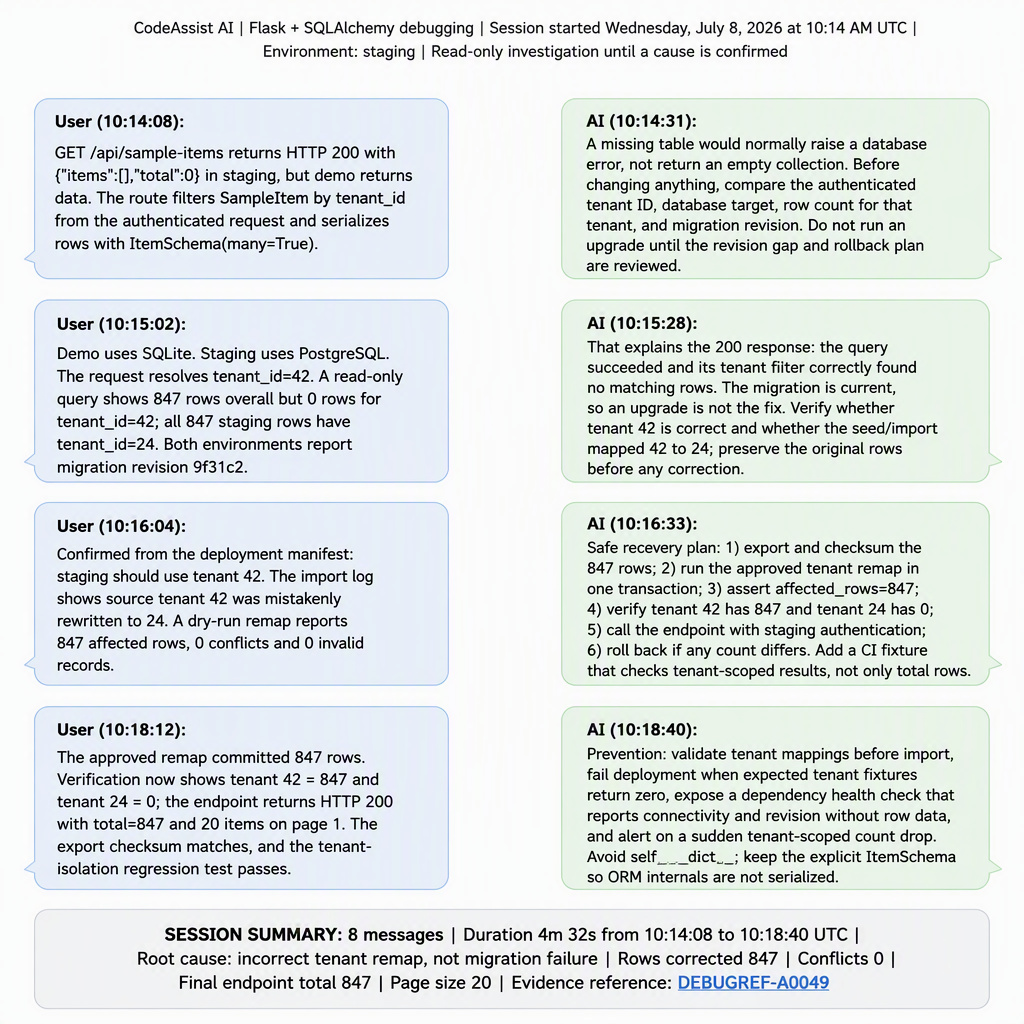} &
    \includegraphics[width=0.205\textwidth]{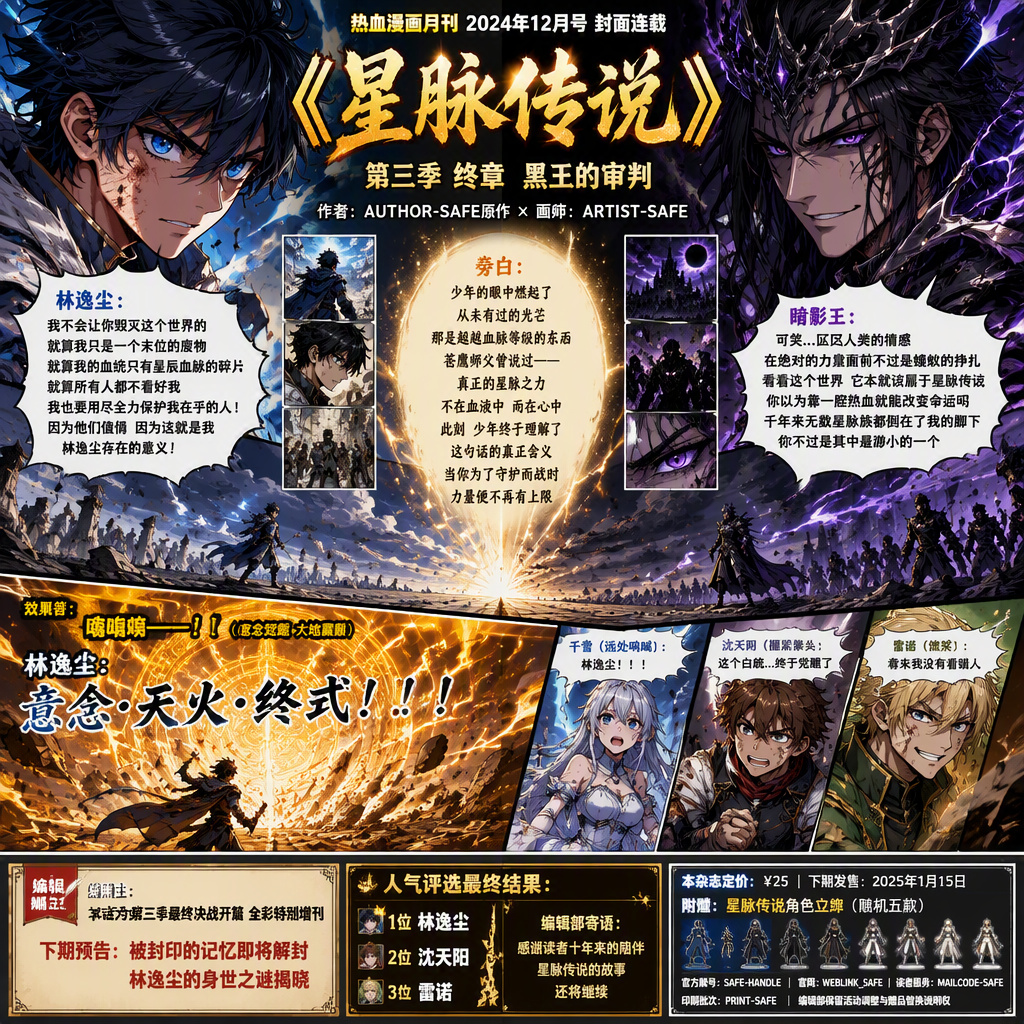} &
    \includegraphics[width=0.205\textwidth]{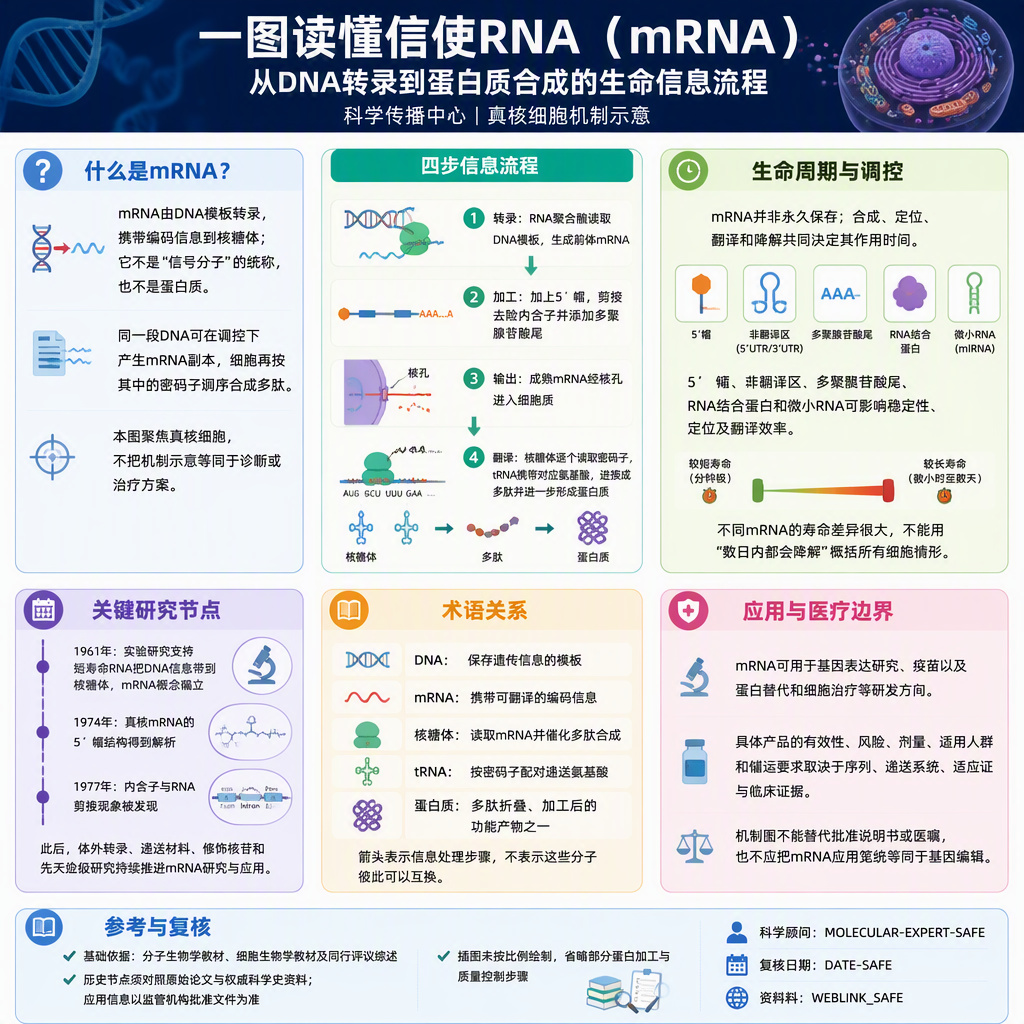} \\
    \scriptsize Caption (EN/L3) & \scriptsize Dialogue (EN/L3) & \scriptsize Comic (ZH/L3) & \scriptsize Infographic (ZH/L3)
  \end{tabular}
  \caption[L3 category gallery]{\textbf{Scene coverage at L3.} One selected output per category in taxonomy order, from GPT Image 2 at API quality \texttt{Low}. L3 denotes Extreme workload; EN/ZH denote English/Chinese.}
  \label{fig:prompt_examples}
\end{figure}

Text-to-image (T2I) models now produce images of high visual quality across diverse prompts~\citep{rombach2022high,saharia2022photorealistic}.
Visual text rendering requires legible, correctly spelled characters at their intended positions.
Recent models have improved short-string rendering~\citep{wu2025qwen,flux-2-2025,gptimage,Dalle-3,esser2024scaling}. Evaluation must now examine whether this ability extends to hundreds or thousands of characters across multiple regions~\citep{du2025textcrafter,zhang2025ocrgenbench}.
Posters, product packaging, documents, receipts, forms, and interfaces require this sustained performance~\citep{peng2025bizgen}: titles, body text, and small supporting regions must all remain correct and legible.
A missing price or an incorrect opening time can change the information conveyed even when the overall composition and lettering look convincing.

\begin{figure}[!t]
  \centering
  \includegraphics[width=\textwidth]{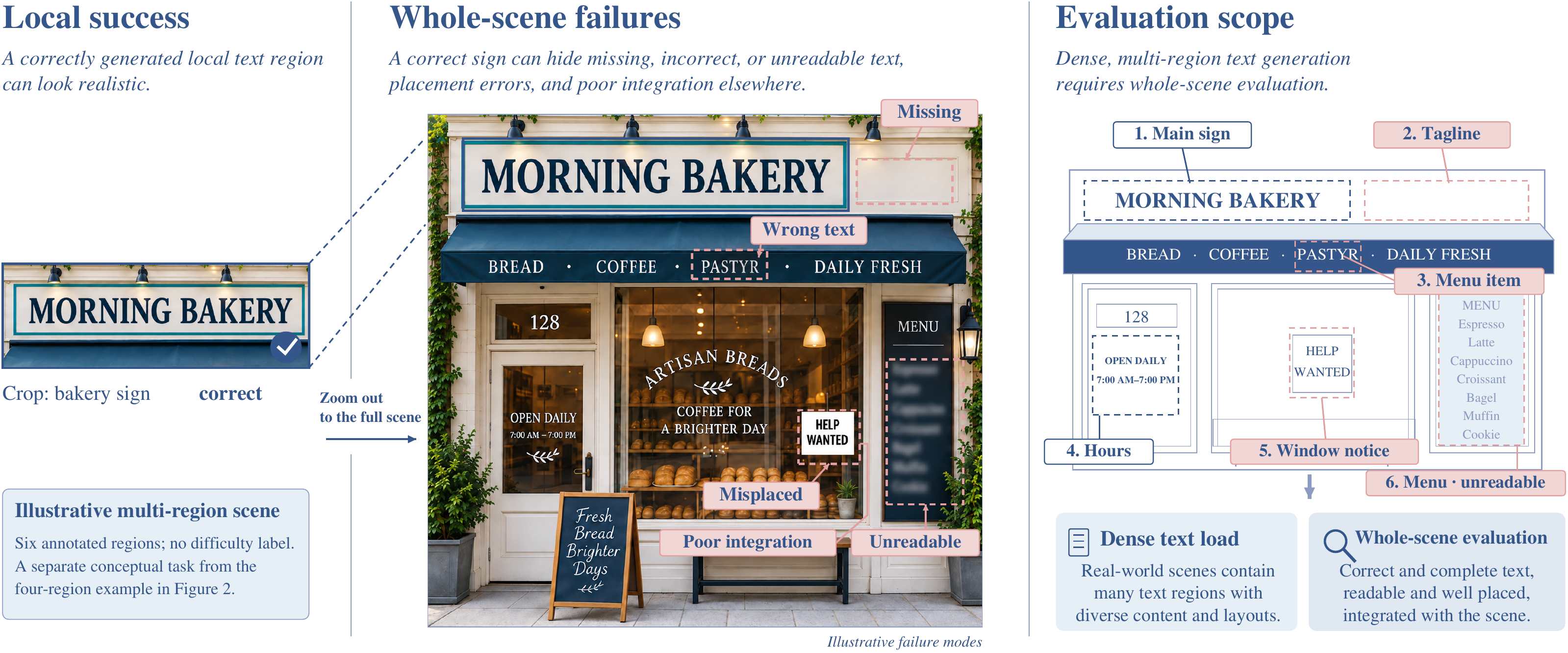}
  \caption[Why local text success is insufficient]{\textbf{Local success can hide whole-scene failures.}
  This illustrative six-region scene contrasts a correct shop name with errors elsewhere.
  \emph{Left.} The correctly rendered shop name, cropped.
  \emph{Center.} The full scene, with five further requested regions marked. Each is
  missing, wrong, unreadable, misplaced, or legible but poorly integrated with
  its intended surface.
  \emph{Right.} Dense text load and whole-scene evaluation cover the requested text.}
  \label{fig:overview_textbench}
  
\end{figure}

To evaluate these applications, we treat visual text generation as a text reproduction task: all target strings are supplied in the prompt, and the model must render them in the specified regions.
Evaluation requires a \emph{dense text load}: prompts that request many regions carrying hundreds to thousands of target characters.
A long passage tests sustained rendering within one block; distributed regions also test whether all requested text remains correctly placed.
\emph{Whole-scene evaluation} requires considering every requested region against a structured reference of its text, placement, and carrier, the physical surface or digital element bearing it.
A correct title can coexist with missing, incorrect, or unreadable text elsewhere in the image. \figref{fig:overview_textbench} illustrates these errors alongside failures of placement and integration with the text carrier.

Existing benchmarks address several parts of this task.
CVTG-2K evaluates multiple text instances with position and attribute descriptions~\citep{du2025textcrafter}, and LongText-Bench targets longer English and Chinese text~\citep{geng2025x}.
STRICT reaches thousand-character sequences~\citep{zhang2025strict}, while OCRGenBench includes dense page-level tasks~\citep{zhang2025ocrgenbench}.
InfoTextBench is the closest comparison for dense bilingual text, with target-string lists and OCR- and VLM-based evaluation~\citep{xiang2026glyphanchor}.
UltraText Bench combines 24 scene categories with a uniform per-region reference and a shared rubric for text, layout, and scene integration. It keeps target strings linked to their intended visual roles across documents, physical signs, and digital interfaces. \secref{sec:related_bench} compares these benchmarks and their evaluation protocols.

We introduce \textbf{UltraText Bench}, a bilingual benchmark built on the two requirements above.
It contains 432 prompts spanning 24 real-world scene categories in six domains, at three difficulty levels and split equally between English and Chinese.
Each prompt requests four to twelve \emph{text regions}: individually annotated target-text components, such as a shop name or a multi-line menu.
Each record pairs a generation prompt containing every target string with structured per-region ground truth (GT), used only for evaluation.
Its $3\times3$ grid gives coarse locations; several regions can share a cell.
\figref{fig:prompt_examples} illustrates all categories at the highest workload level, L3.

A VLM judge, Q-Judger from Qwen-Image-Bench~\citep{li2026qwen}, receives the image and complete reference, returning six image-level scores under our rubric.
Accuracy and completeness form text fidelity; position and layout form spatial quality, with text clarity and scene quality retained separately.
The reference tells the judge which content, locations, and carriers to inspect. Ten participants took part in human evaluation of the automatic scores (\secref{sec:human_review}).

Evaluation of 24 model configurations shows why these distinctions matter.
The Z-Image and LLaDA Turbo variants receive higher clarity but lower fidelity scores than their Base counterparts.
Performance also differs across workload levels and languages.

\Needspace{9\baselineskip}
\textbf{Our main contributions are:}
\begin{enumerate}[label=\alph*., nosep, leftmargin=16pt]
  \item \textbf{UltraText Bench}: 432 bilingual prompts across 24 scene categories and three difficulty levels, with exact target strings and a uniform per-region reference for prompt-only generation.
  \item \textbf{Whole-scene evaluation}: a shared rubric applied with Q-Judger to the image and complete reference, reporting four quality dimensions and evaluation coverage separately.
  \item \textbf{Experimental analysis}: comparisons of 24 model configurations across dimensions, workloads, and languages, accompanied by human evaluation involving ten participants.
\end{enumerate}

\section{Related Work}
\label{sec:related}

\subsection{Visual Text Generation Methods}
\label{sec:related_methods}

Visual text generation methods differ in how they encode exact character sequences, how they obtain a layout, and how well text survives being encoded and decoded by the image model.

\paragraph{Explicit glyph and layout conditioning.}
GlyphControl~\citep{yang2023glyphcontrol} adds a ControlNet branch conditioned on a rendered glyph map, while GlyphDraw~\citep{ma2023glyphdraw} combines glyph information with spatial control for Chinese and English text.
TextDiffuser~\citep{chen2023textdiffuser} instead predicts character-level layout masks before image synthesis, and TextDiffuser-2~\citep{chen2024textdiffuser} uses a language model to plan keywords and bounding boxes from open-ended prompts.
These approaches tie a string more firmly to a place in the image, but they do not assume the same input: glyph-conditioned systems receive an explicit visual control, whereas layout-first systems must predict a layout whose errors can propagate to image generation.

\paragraph{Character-aware and multilingual modeling.}
AnyText~\citep{tuo2024anytext} combines glyph and position conditions with OCR-aware representations in one multilingual generation and editing framework.
Glyph-ByT5~\citep{liu2024glyph} introduces a byte-level text encoder to preserve character information lost to subword tokenization.
JoyType~\citep{li2024joytype} targets multilingual typography, and ViType~\citep{gao2026vitype} jointly models semantic and glyph features in a multimodal diffusion model.
Complementary work studies scene-text inpainting~\citep{zhang2024brush}, position-controlled synthesis~\citep{zhao2024udifftext}, and data synthesis and filtering~\citep{zhao2025lex}.

\paragraph{Dense and multi-region generation.}
Recent methods target longer strings and multiple text regions.
TextCrafter~\citep{du2025textcrafter} steers attention toward the quoted target strings and adds a reinforcement signal read from OCR, improving how multiple strings are rendered in complex scenes.
GlyphDraw2~\citep{ma2025glyphdraw2} combines language-model planning with glyph-conditioned generation for multi-element posters, while PosterMaker~\citep{gao2025postermaker} focuses on text-rich product posters.
BizGen~\citep{peng2025bizgen} addresses article-level infographic generation by binding text spans to planned regions, and TextGuider~\citep{baek2025textguider} applies training-free attention guidance to reduce omissions in longer text.
GlyphAnchor~\citep{xiang2026glyphanchor} anchors glyph priors to predicted positions, and TextGround4M~\citep{mao2026textground4m} complements these methods with prompt-aligned text spans and layout annotations for training layout-aware models.
Their inputs range from prompts alone to explicit glyph and layout controls, with evaluations tailored to each application.

\paragraph{Few-step image generation.}
TwinFlow~\citep{cheng2026twinflow} and APEX, which uses condition shifting~\citep{liu2026self}, study self-adversarial learning for one-step generation.
Duality Models~\citep{sun2026duality} couples velocity and flow-map learning, while Three-Body Scattering~\citep{sun2026three} derives one-step supervision from a distributional energy.
These advances motivate evaluating dense-text fidelity and clarity alongside overall image quality.

\subsection{Text Rendering Benchmarks and Datasets}
\label{sec:related_bench}

Text-rendering resources differ in what they ask the model to produce, in what they tell it beforehand, and in how they check the result.  Scores reported on short-string, poster, document, and long-text test sets therefore cannot be placed on one scale.
\tabref{tab:bench_compare} summarizes the closest comparisons.

\begin{table*}[!t]
\centering
\caption[Text-rendering benchmark comparison]{\textbf{Text-rendering benchmarks: inputs, target references, and evaluation.} The table summarizes the main evaluation components; details appear in \secref{sec:related_bench}.
\textbf{Notes:} EN: English; ZH: Chinese. Extra input denotes benchmark-supplied controls beyond the prompt, not internally predicted layouts.
OCR: optical character recognition; VQA: visual question answering. PNED matches words by edit distance and penalizes unmatched items.
\textsuperscript{\dag}No image is supplied for T2I; source images are used for editing or image-to-image tasks.
UltraText pairs per-region text and visual attributes with image-level VLM ratings.}
\label{tab:bench_compare}
\begingroup
\footnotesize
\setlength{\tabcolsep}{3pt}
\renewcommand{\arraystretch}{1.25}
\begin{tabularx}{\textwidth}{@{}lll>{\hsize=1.05\hsize\linewidth=\hsize\raggedright\arraybackslash}X>{\hsize=0.95\hsize\linewidth=\hsize\raggedright\arraybackslash}X@{}}
\toprule
\textbf{Benchmark} & \textbf{Language} & \textbf{Extra input} & \textbf{Target reference} & \textbf{Evaluation} \\
\midrule
MARIO-Eval & EN & None & Target strings & OCR + image metrics \\
AnyText-Bench & EN/ZH & Glyph/position & Text + positions & OCR on text crops \\
LeX-Bench & EN & None & Text + style/position & PNED + attribute VQA \\
CVTG-2K & EN & None & Text instances & OCR + matching \\
LongText-Bench & EN/ZH & None & Target strings & VLM text accuracy \\
TextInVision & EN & None & Target strings & OCR edit distance \\
TextAtlasEval & EN/ZH & None & Scene + text & OCR + CLIP similarity \\
OCRGenBench & EN/ZH & None / image\textsuperscript{\dag} & Task-specific targets & OCRGenScore \\
InfoTextBench & EN/ZH & None / image\textsuperscript{\dag} & Target text list & OCR + VLM quality \\
\midrule
\rowcolor{gray!15}
\textbf{UltraText Bench} (ours) & EN/ZH & None & Region text + layout/carrier & Image-level VLM \\
\bottomrule
\end{tabularx}
\endgroup
\end{table*}

\paragraph{Short-string fidelity and attributes.}
MARIO-Eval~\citep{chen2023textdiffuser} established a large short-keyword test set.
AnyText-Bench~\citep{tuo2024anytext} added English and Chinese generation tests whose OCR scores are computed on text lines cropped at the specified positions; the AnyText model also edits text, but the benchmark's quantitative protocol is generation-oriented.
TypeScore~\citep{sampaio2024typescore} moved past a single OCR accuracy value: it extracts the rendered text and combines several string-comparison measures into one fidelity score.
That score tells spelling errors, missing text, and extra text apart, and it is validated against human judgments.
LeX-Bench~\citep{zhao2025lex} adds prompts that also constrain color, font, and position, and scores them with PNED, which matches requested against recognized words and penalizes the ones left over.
Because PNED treats words as an unordered set, it says nothing about reading order or about which surface a string ended up on.
These resources primarily measure short-string accuracy and text attributes.

\paragraph{Dense and long-text generation.}
CVTG-2K and the associated TextCrafter multi-text setting~\citep{du2025textcrafter} evaluate more complex scenes and match recognized text instances to requested strings. Their prompt descriptions also encode positions, attributes, and the correspondence between text and its carrier.
LongText-Bench, introduced with X-Omni, covers longer English and Chinese text across eight scenarios~\citep{geng2025x}, and document-level rendering of a single long sequence has been examined separately~\citep{zhang2025strict}.
TextInVision~\citep{fallah2025textinvision} crosses prompt complexity with text properties and also examines visual-encoding failures.
The closest neighbor is InfoTextBench, released with GlyphAnchor~\citep{xiang2026glyphanchor}, which pairs English and Chinese prompts over text-rich reference images, gives each one an explicit list of target strings, and scores word-level precision, recall, and phrase hits.
Long targets, images holding many separate pieces of text, and bilingual coverage have therefore all been studied before.
These resources differ in how target strings, attributes, positions, and scene context are represented and evaluated.

\paragraph{Text rendering in video.}
VTR-Bench~\citep{huang2026vtr} evaluates visual text rendering in video generation.
It assesses text fidelity through carrier-specific transcription and scene and motion requirements through prompt-specific queries.
This temporal setting complements our evaluation of dense text and region placement within individual images.

\begin{figure*}[!t]
  \centering
  \includegraphics[width=\textwidth]{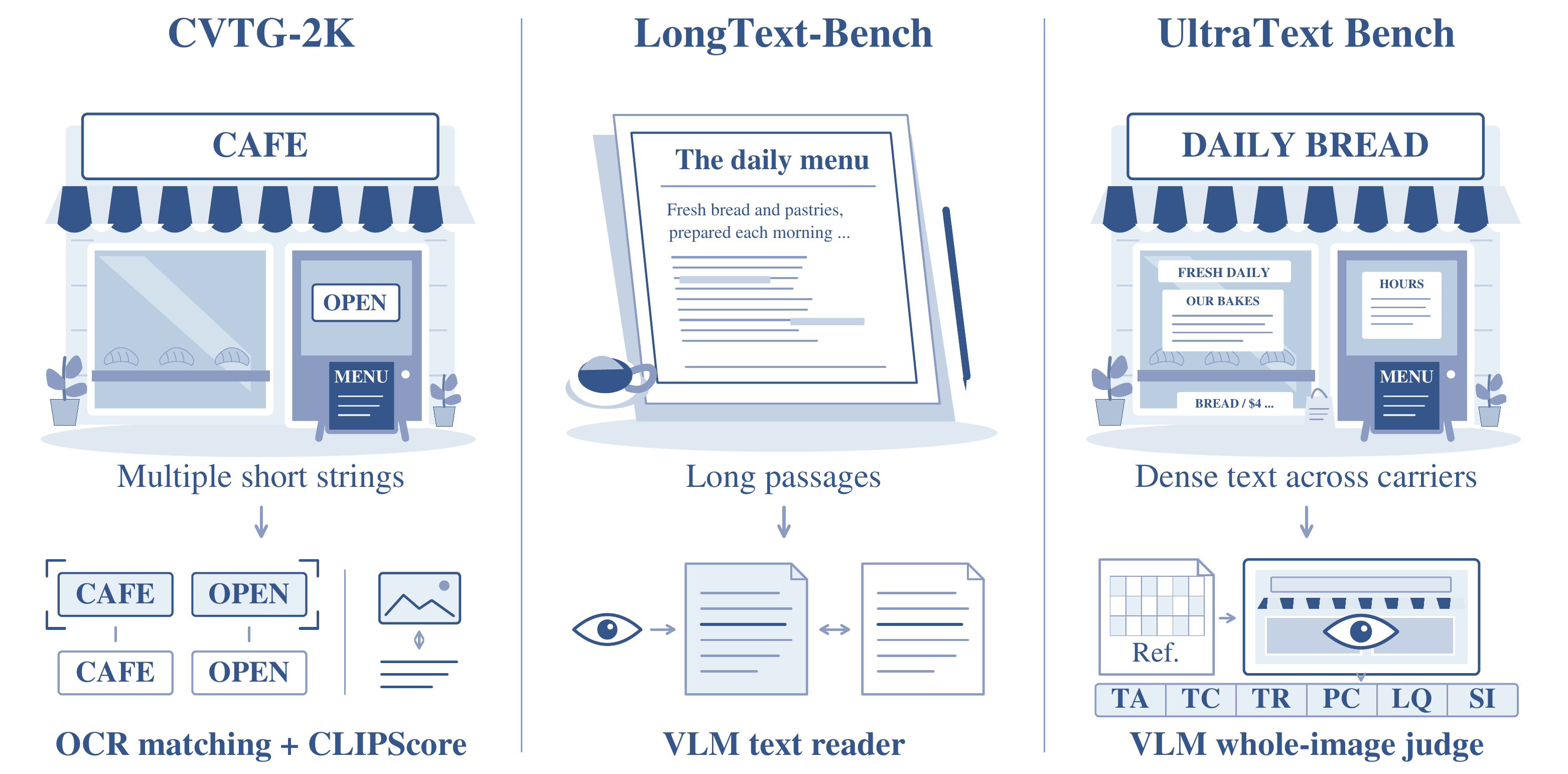}
  \caption[Comparison of three text-rendering benchmarks]{\textbf{Task scenes and assessment in three benchmarks.}
  \emph{Left:} CVTG-2K~\citep{du2025textcrafter} matches OCR-recognized words to targets using Word Accuracy and normalized edit distance, and separately reports CLIPScore.
  \emph{Center:} LongText-Bench~\citep{geng2025x} uses a VLM to extract text for comparison with target strings.
  \emph{Right:} UltraText Bench uses a structured reference (Ref.) to rate the whole image: text accuracy (TA), completeness (TC), readability (TR), position correctness (PC), layout quality (LQ), and scene integration (SI).
  The scenes, abbreviated strings, and line marks are illustrations, not benchmark outputs or measured results; they do not exhaust each benchmark's scene types.}
  \label{fig:benchmark_comparison}
\end{figure*}

\paragraph{Broader visual-text tasks and editing.}
TextAtlas5M~\citep{wangtextatlas5m} contributes large-scale training data for long and structured text, together with TextAtlasEval, an evaluation set spanning four design domains in which one subset varies text length under otherwise fixed conditions.
OCRGenBench~\citep{zhang2025ocrgenbench} broadens evaluation to generation, editing, and OCR-oriented transformation tasks in English and Chinese, and explicitly includes high text density, page-level generation, and dense-document editing.
VTPBench~\citep{shu2025visual} unifies six visual text processing tasks under task-specific metrics and VTPScore, and OneIG-Bench~\citep{chang2026oneig} contributes a bilingual text-rendering split inside a broader image-generation suite.
Task-family aggregates summarize broad performance, while trends with text load and region count require separate breakdowns.
A separate line of benchmarks targets text-centric \emph{editing}, which requires modifying the target text while preserving other text and the background.
TextEditBench~\citep{gui2025texteditbench} covers everyday document and signage scenes and adds a dimension for edits that depend on reasoning.
WeEdit~\citep{zhang2026weedit} extends editing to many languages and scores instruction adherence, text clarity, and background preservation.
TextSculpt-Bench~\citep{lin2026textsculptor} compares the text that should appear in the whole image against the text that actually does, and checks the background outside the edited area.
TextWand-Bench~\citep{wang2026textwand} evaluates removal, generation, and replacement under explicit layout and style control.
None of these is directly comparable to prompt-only generation, because the model is handed a source image and part of the reference is the input itself.
They are still instructive here: TextSculpt-Bench in particular shows that checking all the text has to leave behind evidence that can be inspected, since handing an evaluator a complete reference does not prove that it looked at every region.

\paragraph{General T2I evaluation and our scope.}
General text-to-image benchmarks such as TIFA~\citep{hu2023tifa}, T2I-CompBench~\citep{huang2023t2icompbench}, and HRS-Bench~\citep{bakr2023hrs} measure semantic faithfulness and composition over broader prompt families.
BizGenEval~\citep{li2026bizgeneval} comes closer to our scene types, scoring slides, charts, webpages, posters, and scientific figures against human-verified checklist questions.
Their checklists and alignment metrics address broader semantic and compositional constraints, complementing the explicit per-region text references used here.
UltraText Bench combines prompt-only generation, a uniform per-region reference, and coverage of 24 text-bearing scene categories.
The generator receives all target strings in natural language; the six-field annotation is reserved for evaluation and stays consistent across scenes, languages, and levels.
This design tests whether text rendering transfers across carriers and layouts under increasing text load.
The reference specifies coarse spatial metadata, and the protocol returns image-level scores; region-level attribution remains a future direction.

\subsection{Automatic Evaluation and Judge Reliability}
\label{sec:related_vlm}

\paragraph{Global and OCR-based metrics.}
Automatic generation evaluation spans global alignment and preference metrics, fidelity measures from OCR output, and VLM judges that follow a written rubric.
CLIPScore~\citep{hessel2021clipscore} measures global image-text similarity, while ImageReward~\citep{xu2023imagereward}, HPSv2~\citep{wu2023human}, and PickScore~\citep{kirstain2023pick} learn scalar preferences from human comparisons.
These scores are useful for semantic alignment or overall preference, but their training objectives do not require transcribing every requested character and can favor an attractive image whose text is incorrect.
Text-specific metrics provide a closer signal.
TypeScore~\citep{sampaio2024typescore}, described above, also reports a failure mode on the evaluator's side: OCR can introduce recognition errors, while a VLM reader can silently correct characters that the image actually renders wrongly.
OCRGenScore~\citep{zhang2025ocrgenbench} combines metrics across OCR-generative tasks, and VTPScore~\citep{shu2025visual} uses a multimodal model to rate visual quality and text readability across visual text processing tasks.
OCR-based scores depend on the detector, recognizer, language settings, font, scale, and image domain~\citep{baek2019wrong,cui2025paddleocr}.

\paragraph{Text quality and spatial fidelity.}
A complementary line evaluates the visual quality of rendered text separately from its correctness.
TIQA~\citep{koltsov2026tiqa} collects human quality ratings for rendered text, on crops and on text-heavy images, and trains a model to predict them; this is what our clarity dimension tries to capture.
Because it rates regions that were detected, it cannot by itself report a region that was requested and never rendered.
Recognized strings alone also do not establish whether text occupies the requested region or integrates naturally with its carrier.

\paragraph{Rubric-based VLM judges.}
Prompted VLM evaluators can apply task-specific rubrics without training a separate metric for every dimension.
VIEScore~\citep{ku2024viescore} scores semantic consistency and perceptual quality with explanations, and UniGenBench++~\citep{wang2025unigenbench++} uses structured semantic units for multidimensional T2I evaluation.
Qwen-Image-Bench (QIB)~\citep{li2026qwen} develops fine-grained creation rubrics and a dedicated Q-Judger using professional human ratings.
RichHF~\citep{liang2024rich} likewise demonstrates the value of fine-grained human feedback tied to words and image regions.
T2LSC-Bench~\citep{wang2026t2lsc} pairs OCR verification with a structured VLM judgement, separating whether the text landed on its intended anchor from whether its meaning leaked into the surrounding subject and scene, and it checks the automatic labels against a human-annotated subset.
Evaluator version, prompting, text-reading ability, and visual-quality bias can affect VLM scores.
Prior studies examine agreement with independent human ratings~\citep{zheng2023judging}, controlled annotation protocols for T2I evaluation~\citep{otani2023toward}, and variation in text-reading ability across multimodal models~\citep{liu2023hidden,fu2026ocrbench}.
These limitations motivate joint assessment of text, placement, and scene integration, while keeping response-format validity distinct from the reliability of a judge's assessment.

\section{UltraText Bench: Benchmark Design}
\label{sec:benchmark}

\begin{figure*}[!t]
  \centering
  \includegraphics[width=\textwidth]{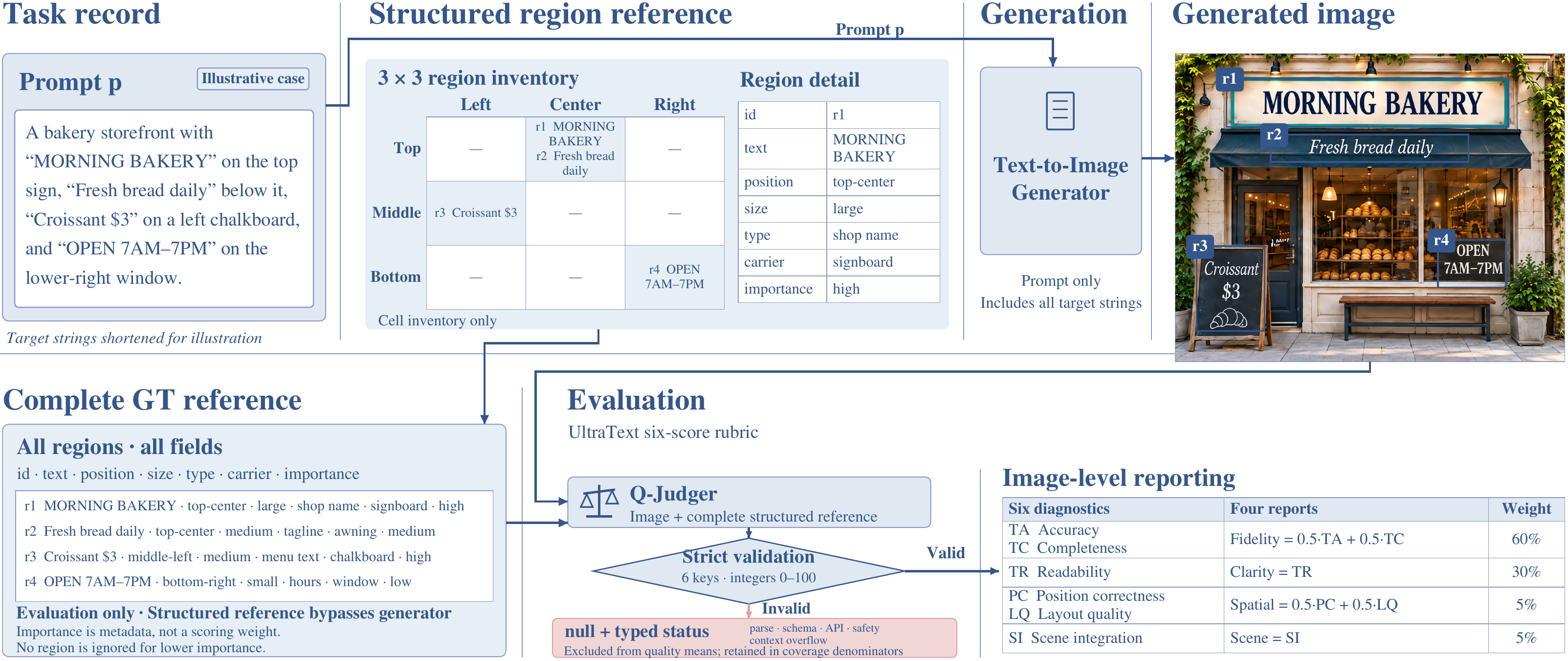}
  \caption[Task definition and evaluation protocol]{\textbf{Task definition and evaluation protocol.}
  The generator receives only prompt $p$.
  Q-Judger, from Qwen-Image-Bench~\citep{li2026qwen}, applies the UltraText rubric to the image and complete reference $R$, returning six image-level scores that map to four reporting dimensions and a composite.
  Reference fields and scores are not paired one-to-one. Grid cells give coarse locations; regions may share a cell.
  Invalid responses remain unscored and reduce coverage.
  The bakery schematic simplifies the four-region record \texttt{A1\_L1\_EN\_001}, reproduced in \appref{app:complete_example}.}
  \label{fig:task_eval}
\end{figure*}

UltraText Bench examines how text fidelity, clarity, spatial quality, and scene quality vary across dense text-bearing scenes, workloads, and languages.

\subsection{Task Formulation}
\label{sec:task_def}

\paragraph{Inputs and generation.}
A benchmark instance pairs a natural-language prompt $p$ with a structured reference $R=\{r_1, \ldots, r_K\}$ for $K$ text regions.
Each region is an annotated target-text component that can contain multiple lines or items; for example, the entire menu in the record below forms one region.
The prompt contains all target strings and is the only input to the text-to-image model $M$, which generates an image $I=M(p)$.
Each $r_i$ records a target string $t_i$, grid position, relative text size, type, carrier, and importance (\secref{sec:prompt_gt}); this reference is used only for evaluation.

\paragraph{Illustrative record.}
The released L1 signage record \texttt{A1\_L1\_EN\_001} describes a bakery storefront with $K=4$ regions and 423 GT characters in total.
Two regions occupy \texttt{top-center}: a \texttt{large} shop name on a wooden sign and a \texttt{medium} tagline on an oak plank below it.
The remaining regions are a \texttt{medium} chalkboard menu at \texttt{middle-left} and \texttt{small} opening hours on a brass plate at \texttt{bottom-right}; their importance levels range from \texttt{high} to \texttt{low}.
\figref{fig:task_eval} uses a simplified schematic inspired by this layout, while \secref{sec:prompt_gt} reproduces the first region verbatim.

\paragraph{Evaluation scope.}
The complete reference specifies which strings should appear, where they belong, and how they relate to the scene. Q-Judger assesses these requirements through six image-level scores. \appref{app:reference_role} explains how the reference supports each assessment; the output contains no region-level scores or transcriptions.

\subsection{Construction Rationale}
\label{sec:design_phil}
\label{sec:construction_rationale}

The 24 categories cover different carriers and layouts, from signs to invoices and code. Three difficulty levels increase text load and region count, while the English and Chinese splits cover different glyph inventories and writing conventions~\citep{ma2023glyphdraw,tuo2024anytext}. These choices support comparisons across scene, workload, and language. Because text load and region count change together, the levels do not isolate their individual effects.

Each category$\times$level$\times$language cell contains three prompts, yielding $24\times3\times2\times3=432$ prompts. Three prompts per cell preserve taxonomy coverage within the generation and inspection budget. Four samples per prompt give 12 images per cell and 288 per level and language in a complete run. This supports aggregate analysis while retaining category coverage; individual cells have limited statistical precision. The suite is a stress test rather than a frequency-weighted sample of web images. Its layouts place different demands on spatial organization: receipts align items with prices, newspapers separate columns and headlines, and signs follow the perspective of their carriers.

\subsection{Scene Taxonomy: 24 Categories in 6 Domains}
\label{sec:taxonomy}

The six domains cover photographs, documents, and digital interfaces. \tabref{tab:categories} groups the 24 categories by carrier, density, and layout, with definitions in \appref{app:categories}.

\begin{table*}[!t]
\centering
\caption[UltraText Bench scene taxonomy]{\textbf{UltraText Bench scene taxonomy.} Six domains organize 24 categories by rendering challenge and typical layout. \textbf{Notes:} Density describes qualitative text packing, not the GT-character load in \tabref{tab:stats}. Layouts are typical examples, not mandatory templates.}
\label{tab:categories}
\begingroup
\footnotesize
\setlength{\tabcolsep}{3pt}
\renewcommand{\arraystretch}{1.18}
\begin{tabularx}{\textwidth}{@{}ll>{\raggedright\arraybackslash}Xll@{}}
\toprule
\rowcolor{utLight}
\textbf{ID} & \textbf{Category} & \textbf{Rendering challenge} & \textbf{Density} & \textbf{Typical layout} \\
\midrule
\rowcolor{utAccent!7}
\multicolumn{5}{@{}l}{\textbf{A. Signage \& Labels}} \\
A1 & Sign & Perspective and weathered surfaces & Low--Med & Scattered blocks \\
A2 & Label & Small text on curved surfaces & Med & Compact blocks \\
A3 & Poster & Font hierarchy and composition & Med--High & Stacked blocks \\
A4 & Billboard & Scale and viewing angles & Low--Med & Large text blocks \\
\midrule
\rowcolor{utAccent!7}
\multicolumn{5}{@{}l}{\textbf{B. Documents \& Print}} \\
B1 & Article & Paragraph flow and alignment & High & Single-column \\
B2 & Newspaper & Columns and headline hierarchy & Very High & Multi-column \\
B3 & Letter & Salutation, body, and closing & High & Single-column \\
B4 & Resume & Sections and bullet alignment & High & Two-column \\
\midrule
\rowcolor{utAccent!7}
\multicolumn{5}{@{}l}{\textbf{C. Commercial}} \\
C1 & Menu & Item-price alignment & Med--High & Tabular \\
C2 & Receipt & Narrow lines and exact totals & Med & Single-column \\
C3 & Invoice & Numeric fields and table alignment & Med--High & Tabular \\
C4 & Product packaging & Perspective and ingredient lists & Med & Multi-panel \\
\midrule
\rowcolor{utAccent!7}
\multicolumn{5}{@{}l}{\textbf{D. Digital Interfaces}} \\
D1 & Webpage & Navigation, content, and sidebar & High & Multi-panel \\
D2 & Slide & Title and bullet hierarchy & Med & Stacked blocks \\
D3 & Social media & Interface text and hashtags & Med & Card/feed \\
D4 & Dashboard & KPI cards and chart labels & High & Grid \\
\midrule
\rowcolor{utAccent!7}
\multicolumn{5}{@{}l}{\textbf{E. Structured Data}} \\
E1 & Schedule & Time-column alignment & Med--High & Tabular \\
E2 & Form & Label-field alignment & Med & Field grid \\
E3 & Certificate & Formal typography and alignment & Low--Med & Centered blocks \\
E4 & Code & Monospace text and indentation & High & Indented lines \\
\midrule
\rowcolor{utAccent!7}
\multicolumn{5}{@{}l}{\textbf{F. Creative \& Special}} \\
F1 & Caption & Legibility over backgrounds & Low--Med & Overlay \\
F2 & Dialogue & Speaker labels and text bubbles & Med & Alternating blocks \\
F3 & Comic panel & Speech bubbles and sound effects & Med & Scattered bubbles \\
F4 & Infographic & Text-label placement & Med--High & Freeform \\
\bottomrule
\end{tabularx}
\endgroup
\end{table*}

\subsection{Difficulty Stratification}
\label{sec:difficulty}

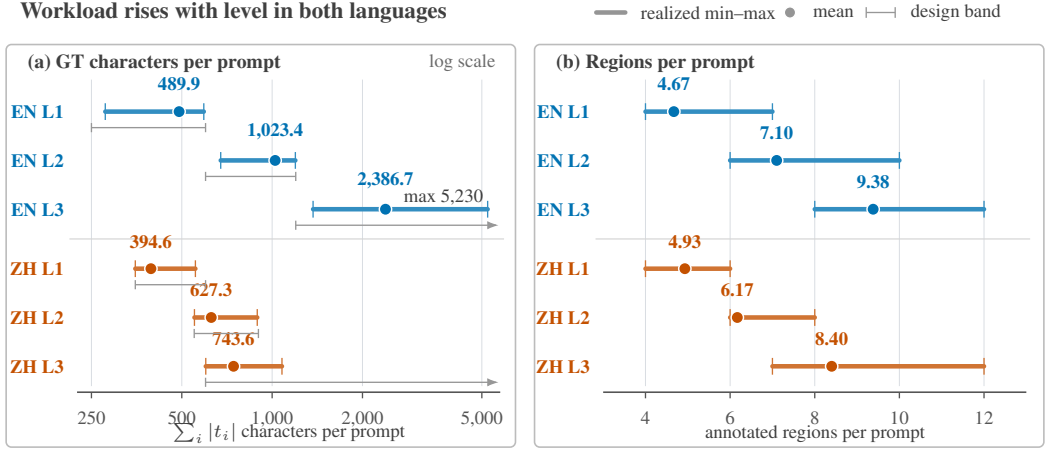
\begin{figure*}[!t]
  \centering
  \resizebox{\textwidth}{!}{\begin{tikzpicture}[x=1cm,y=1cm,
    rowlab/.style={anchor=east,font=\fontsize{7.0}{7.6}\selectfont\bfseries},
    ticklab/.style={anchor=north,font=\fontsize{7.0}{7.6}\selectfont,text=utAxis},
    meanlab/.style={anchor=south,font=\fontsize{7.0}{7.6}\selectfont\bfseries},
    endlab/.style={anchor=east,font=\fontsize{7.0}{7.6}\selectfont,text=utNeutral},
    axtitle/.style={font=\fontsize{7.0}{7.6}\selectfont,text=utNeutral},
    band/.style={line width=0.5pt,draw=utLegacy!70},
    bandcap/.style={line width=0.5pt,draw=utLegacy!70},
    bandarrow/.style={-{Latex[length=1.5mm,width=1.1mm]},line width=0.5pt,
      draw=utLegacy!70}]
    \path (0,0) rectangle (13.85,6.10);

    \node[anchor=west,font=\fontsize{8.6}{9.6}\selectfont\bfseries,text=utNeutral]
      at (0.18,5.90) {Workload rises with level in both languages};

    \draw[line width=1.7pt,draw=utNeutral!55,line cap=round] (7.86,5.90) -- (8.24,5.90);
    \node[anchor=west,font=\fontsize{7.0}{7.6}\selectfont,text=utNeutral]
      at (8.32,5.90) {realized min--max};
    \filldraw[fill=utNeutral!55,draw=white,line width=0.5pt] (10.40,5.90) circle (2.1pt);
    \node[anchor=west,font=\fontsize{7.0}{7.6}\selectfont,text=utNeutral]
      at (10.58,5.90) {mean};
    \draw[band] (11.40,5.90) -- (11.78,5.90);
    \draw[bandcap] (11.40,5.83) -- (11.40,5.97);
    \draw[bandcap] (11.78,5.83) -- (11.78,5.97);
    \node[anchor=west,font=\fontsize{7.0}{7.6}\selectfont,text=utNeutral]
      at (11.86,5.90) {design band};

    \path[draw=utNeutral!35,fill=white,rounded corners=2pt,line width=0.6pt]
      (0.12,0.20) rectangle (6.80,5.48);
    \path[draw=utNeutral!35,fill=white,rounded corners=2pt,line width=0.6pt]
      (7.05,0.20) rectangle (13.73,5.48);
    \node[anchor=west,font=\fontsize{7.4}{8.0}\selectfont\bfseries,text=utNeutral]
      at (0.28,5.24) {(a) GT characters per prompt};
    \node[anchor=east,font=\fontsize{7.0}{7.6}\selectfont,text=utLegacy]
      at (6.66,5.24) {log scale};
    \node[anchor=west,font=\fontsize{7.4}{8.0}\selectfont\bfseries,text=utNeutral]
      at (7.21,5.24) {(b) Regions per prompt};

    \foreach \x/\lab in {1.2387/250,2.4236/500,3.6085/1{,}000,4.7934/2{,}000,
                         6.3588/5{,}000}{
      \draw[ut/grid] (\x,1.00) -- (\x,5.02);
      \draw[ut/tick] (\x,0.84) -- (\x,0.92);
      \node[ticklab] at (\x,0.80) {\lab};
    }
    \draw[line width=0.6pt,draw=utAxis] (1.05,0.92) -- (6.60,0.92);
    \node[axtitle] at (3.83,0.38) {$\sum_i |t_i|$ characters per prompt};

    \foreach \x/\lab in {8.505/4,9.615/6,10.725/8,11.835/10,12.945/12}{
      \draw[ut/grid] (\x,1.00) -- (\x,5.02);
      \draw[ut/tick] (\x,0.84) -- (\x,0.92);
      \node[ticklab] at (\x,0.80) {\lab};
    }
    \draw[line width=0.6pt,draw=utAxis] (7.95,0.92) -- (13.50,0.92);
    \node[axtitle] at (10.73,0.38) {annotated regions per prompt};

    \draw[draw=utNeutral!22,line width=0.45pt] (0.95,2.93) -- (6.65,2.93);
    \draw[draw=utNeutral!22,line width=0.45pt] (7.85,2.93) -- (13.55,2.93);

    \foreach \y/\lab/\lo/\hi/\mx/\mv/\col in {
      4.60/EN L1/1.4202/2.7123/2.3883/489.9/utEN,
      3.96/EN L2/2.9341/3.9131/3.6481/1{,}023.4/utEN,
      3.32/EN L3/4.1455/6.4350/5.0950/2{,}386.7/utEN,
      2.54/ZH L1/1.8139/2.6020/2.0188/394.6/utZH,
      1.90/ZH L2/2.5896/3.4133/2.8100/627.3/utZH,
      1.26/ZH L3/2.7381/3.7375/3.1024/743.6/utZH}{
      \node[rowlab,text=\col] at (1.00,\y) {\lab};
      \draw[line width=1.7pt,draw=\col!80,line cap=round] (\lo,\y) -- (\hi,\y);
      \draw[line width=0.5pt,draw=\col!80] (\lo,{\y-0.11}) -- (\lo,{\y+0.11});
      \draw[line width=0.5pt,draw=\col!80] (\hi,{\y-0.11}) -- (\hi,{\y+0.11});
      \filldraw[fill=\col,draw=white,line width=0.55pt] (\mx,\y) circle (2.3pt);
      \node[meanlab,text=\col] at (\mx,{\y+0.15}) {\mv};
    }
    \node[endlab] at (6.50,{3.32+0.16}) {max 5{,}230};

    \foreach \y/\lo/\hi in {
      4.60/1.2387/2.7353,
      3.96/2.7353/3.9181,
      2.54/1.8139/2.7353,
      1.90/2.5865/3.4286}{
      \draw[band] (\lo,{\y-0.21}) -- (\hi,{\y-0.21});
      \draw[bandcap] (\lo,{\y-0.27}) -- (\lo,{\y-0.15});
      \draw[bandcap] (\hi,{\y-0.27}) -- (\hi,{\y-0.15});
    }
    \foreach \y/\lo in {3.32/3.9181, 1.26/2.7353}{
      \draw[bandarrow] (\lo,{\y-0.21}) -- (6.58,{\y-0.21});
      \draw[bandcap] (\lo,{\y-0.27}) -- (\lo,{\y-0.15});
    }

    \foreach \y/\lab/\lo/\hi/\mx/\mv/\col in {
      4.60/EN L1/8.505/10.170/8.8769/4.67/utEN,
      3.96/EN L2/9.615/11.835/10.2255/7.10/utEN,
      3.32/EN L3/10.725/12.945/11.4909/9.38/utEN,
      2.54/ZH L1/8.505/9.615/9.0212/4.93/utZH,
      1.90/ZH L2/9.615/10.725/9.7094/6.17/utZH,
      1.26/ZH L3/10.170/12.945/10.947/8.40/utZH}{
      \node[rowlab,text=\col] at (7.90,\y) {\lab};
      \draw[line width=1.7pt,draw=\col!80,line cap=round] (\lo,\y) -- (\hi,\y);
      \draw[line width=0.5pt,draw=\col!80] (\lo,{\y-0.11}) -- (\lo,{\y+0.11});
      \draw[line width=0.5pt,draw=\col!80] (\hi,{\y-0.11}) -- (\hi,{\y+0.11});
      \filldraw[fill=\col,draw=white,line width=0.55pt] (\mx,\y) circle (2.3pt);
      \node[meanlab,text=\col] at (\mx,{\y+0.15}) {\mv};
    }
  \end{tikzpicture}}
  \caption[Realized difficulty statistics]{\textbf{Realized difficulty statistics for the 432 prompts.}
  Each row contains 72 prompts. Bars show minimum--maximum ranges and dots show means from \tabref{tab:stats}.
  (a) Target-string character counts, including spaces and punctuation, on a log axis. Grey brackets show the language-specific design bands; bounds are inclusive, ZH bands overlap, and L3 bands have no upper bound (shown by open ends).
  Levels are assigned workloads, not disjoint character bins; all records satisfy their own band.
  (b) Region counts; no per-level region-count bands are specified.
  Both panels describe benchmark inputs, not model performance.}
  \label{fig:difficulty_stats}
\end{figure*}

UltraText Bench uses three language-specific difficulty levels: L1 (Hard), L2 (Very Hard), and L3 (Extreme). Mean GT-character load and region count increase across these levels in both languages. A prompt's GT-character load is $\sum_i |t_i|$, the sum of Unicode characters in all region strings, including spaces and punctuation.
\figref{fig:difficulty_stats} compares the design bands with the realized workloads. All bounds are inclusive: EN bands share endpoints, ZH bands overlap, and L3 is open ended in both languages. Levels describe assigned workloads rather than disjoint character-count bins, with region count providing an additional difficulty axis.

Three levels provide an intermediate workload while retaining 72 prompts per level and language. EN design bands are 250--600, 600--1200, and 1200+ GT characters; ZH bands are 350--600, 550--900, and 600+, respectively. The realized distributions also reveal how much the workload differs between languages at a given level.

\subsection{Bilingual Strategy}
\label{sec:bilingual}

English and Chinese prompts were written independently and adapted to their cultural settings, including venue names, prices, vocabulary, and ordering conventions. The splits balance category and level counts but use different text-load bands; their scores therefore compare the two prompt sets without isolating language alone. Contact details and other identifying strings use fixed placeholders. \appref{app:construction} provides their vocabulary and bilingual examples.

\subsection{Prompt Construction and Ground Truth}
\label{sec:prompt_gt}

Each record pairs a scene description containing all GT strings with a structured GT annotation.

\paragraph{Target strings in the prompt.}
Prompts embed target strings as quoted text, fenced passages, or structured content blocks. In the bakery record \texttt{A1\_L1\_EN\_001}, the first region appears as:
\emph{``...a hand-carved wooden shop sign features cream-white serif lettering that reads:} \textbf{``DAILY BREAD --- Artisan Bakery \& Fine Patisserie Since 1987''}\emph{''}.
The scene description is abridged for readability, while the target text is reproduced verbatim.

\paragraph{Structured GT regions.}
Each region records the exact target text, a $3\times3$ grid position, relative size, text type, carrier, and importance. These attributes describe the intended text and its role in the scene: a title on a wooden sign and opening hours on a small plate have different visual requirements, even when both are correctly spelled. Size is qualitative, with no pixel or font-size thresholds. Importance is reference metadata and enters no scoring formula; the rubric requires assessment of lower-importance regions as well. The attributes guide image-level assessment without defining six corresponding attribute scores. \appref{app:construction} lists the annotation vocabulary, and \appref{app:prompt_examples} provides a complete English record and a Chinese example.

\subsection{Quality Assurance}
\label{sec:qa}

All 432 prompts were manually reviewed for natural scene descriptions and checked for exact target-string inclusion, complete annotations, and compliance with the language-specific workload bands. Automated checks verified region identifiers, the six attributes and their allowed values, and character statistics. Manual revision preserved the target strings while making their placement and carriers clear in context. Every released record passes these checks and falls within its assigned band. \appref{app:construction} describes the construction and review process.

\subsection{Dataset Statistics}
\label{sec:stats}

\begin{table*}[!t]
\centering
\caption[Dataset statistics]{\textbf{Dataset statistics by language and difficulty.} \textbf{Notes:} Means and ranges are per prompt. GT characters include spaces and punctuation and are summed over all requested regions. The 432 prompts contain 407,918 GT characters and 2,926 regions in total; all statistics are recomputed from the released region arrays.}
\label{tab:stats}
\begingroup
\footnotesize
\setlength{\tabcolsep}{4pt}
\renewcommand{\arraystretch}{1.2}
\begin{tabularx}{\textwidth}{@{}>{\hsize=0.95\hsize\linewidth=\hsize\raggedright\arraybackslash}X>{\hsize=0.85\hsize\linewidth=\hsize\raggedleft\arraybackslash}X>{\hsize=1.10\hsize\linewidth=\hsize\raggedleft\arraybackslash}X>{\hsize=1.60\hsize\linewidth=\hsize\raggedleft\arraybackslash}X>{\hsize=0.80\hsize\linewidth=\hsize\raggedleft\arraybackslash}X>{\hsize=0.70\hsize\linewidth=\hsize\raggedleft\arraybackslash}X@{}}
\toprule
\rowcolor{utLight}
\textbf{Split} & \textbf{Prompts} & \multicolumn{2}{c}{\textbf{GT characters per prompt}} & \multicolumn{2}{c}{\textbf{Regions per prompt}} \\
\cmidrule(lr){3-4}\cmidrule(lr){5-6}
& & \textbf{Mean} & \textbf{Range} & \textbf{Mean} & \textbf{Range} \\
\midrule
EN L1 & 72 & 489.90 & 278--592 & 4.67 & 4--7 \\
EN L2 & 72 & 1,023.44 & 674--1,195 & 7.10 & 6--10 \\
EN L3 & 72 & 2,386.68 & 1,369--5,230 & 9.38 & 8--12 \\
\addlinespace[3pt]
ZH L1 & 72 & 394.62 & 350--555 & 4.93 & 4--6 \\
ZH L2 & 72 & 627.25 & 551--892 & 6.17 & 6--8 \\
ZH L3 & 72 & 743.62 & 601--1,079 & 8.40 & 7--12 \\
\midrule
\rowcolor{utAccent!7}
\textbf{All prompts} & \textbf{432} & 944.25 & 278--5,230 & 6.77 & 4--12 \\
\bottomrule
\end{tabularx}
\endgroup
\end{table*}

\tabref{tab:stats} summarizes the dataset under this region-text character definition, and \figref{fig:difficulty_stats} plots the same per-split ranges and means against the design bands.
The uniform $24 \times 3 \times 3$ grid yields 216 prompts per language and 432 total. At four generated samples per prompt, a complete run contains 864 images per language and 1,728 images per model configuration.

The distributions show how the assigned workloads differ between languages. EN mean character load increases from 489.90 at L1 to 2,386.68 at L3, while mean region count rises from 4.67 to 9.38. ZH mean character load increases from 394.62 to 743.62, with region count rising from 4.93 to 8.40. Both languages therefore require more regions at higher levels, but the English prompts also show a much larger increase in total text. These distributions provide the context for interpreting language-specific scores at each level.

\section{Evaluation Framework}
\label{sec:evaluation}

We use the Q-Judger model from Qwen-Image-Bench~\citep{li2026qwen} with the UltraText scoring rubric to compare generated images against the complete structured reference (\figref{fig:task_eval}).

All target strings reach the judge without truncation. Failed evaluations remain unscored and reduce coverage; \appref{app:reference_payload} details validation and failure handling.

\subsection{Four-Dimension Scoring}
\label{sec:scoring}

Given a generated image $I$ and the complete structured reference $R=\{r_1,\ldots,r_K\}$, the judge $J$ returns six image-level raw scores:
\begin{equation}
  \mathbf{s}_{\text{raw}} = J(I, R) \in [0, 100]^6.
\end{equation}
Raw scores cover content correctness and completeness (TA, TC), visual legibility (TR), position agreement and typographic organization (PC, LQ), and carrier fit (SI). \tabref{tab:dimensions} maps them into four reporting dimensions $\mathbf{s} \in [0,100]^4$.
These are VLM ratings, not measured percentages of correct characters or recovered regions. The rubric defines the six properties but provides no intermediate score thresholds or explicit character-counting rule. The notation $R$ omits auxiliary record metadata.

\begin{table*}[!t]
\centering
\caption[Six-to-four score mapping]{\textbf{From six raw scores to four reporting dimensions.} Scores are VLM ratings on a 0--100 scale, with higher values better. TA/TC/TR denote text accuracy/completeness/readability; PC/LQ/SI denote position correctness/layout quality/scene integration. Weights apply to reporting dimensions, abbreviated Fidelity, Clarity, Spatial, and Scene in the figures and leaderboard.}
\label{tab:dimensions}
\begingroup
\footnotesize
\setlength{\tabcolsep}{4pt}
\renewcommand{\arraystretch}{1.25}
\begin{tabularx}{\textwidth}{@{}lc>{\raggedright\arraybackslash}Xr@{}}
\toprule
\rowcolor{utLight}
\textbf{Dimension} & \textbf{Formula} & \textbf{Assessment focus} & \textbf{Weight} \\
\midrule
Text fidelity & $(\mathrm{TA}+\mathrm{TC})/2$ & Text correctness and completeness & 60\% \\
Text clarity & $\mathrm{TR}$ & Sharpness, contrast, and legibility & 30\% \\
\midrule
Spatial quality & $(\mathrm{PC}+\mathrm{LQ})/2$ & Position, alignment, and organization & 5\% \\
Scene quality & $\mathrm{SI}$ & Natural fit with the text carrier and scene & 5\% \\
\bottomrule
\end{tabularx}
\endgroup
\end{table*}

The \textbf{Composite} is computed as:
\begin{equation}
  \text{Composite} = 0.60 \cdot s_{\text{fidelity}} + 0.30 \cdot s_{\text{clarity}} + 0.05 \cdot s_{\text{spatial}} + 0.05 \cdot s_{\text{scene}}
\end{equation}
The weights emphasize preservation of the requested content and its legibility: fidelity contributes 60\% and clarity 30\%. Spatial and scene quality contribute the remaining 10\%, so their separate scores are needed to interpret placement and integration errors.

\paragraph{Why four dimensions?}
The six raw scores describe different, potentially overlapping failure modes. For concise reporting, we pair correctness with completeness and position with layout, while retaining readability as text clarity and scene integration as scene quality.
Clearly drawn but misspelled text can score differently on fidelity and clarity. The raw TA and TC scores remain available to distinguish correctness from completeness. For example, the menu in \figref{fig:app_diag_readability} receives TA=10 and TC=100, yielding Fidelity=55. Reporting only this average would conceal the large difference between the two raw ratings. In this case, the high completeness rating does not imply accurate reproduction of the menu text. Grid positions describe coarse placement; layout quality rates organization without explicit GT for every within-cell relation.
The full judge prompt appears in \appref{app:prompt_template}. The interpretive bands in \appref{app:rubric} are a reading aid and are excluded from the judge prompt.

\subsection{Human Evaluation}
\label{sec:human_review}

Ten participants took part in a human evaluation of the automatic scores. The four reporting dimensions distinguish required content, readability, spatial organization, and scene integration. \appref{app:human_alignment} discusses the scope of this evaluation and the interpretation of qualitative comparisons. Quantitative inter-rater and human--judge agreement statistics are not reported here.

The qualitative examples in \appref{app:diagnostics} make these distinctions inspectable. Each pairs the relevant target text with a whole image and a native-resolution crop. The hours plate in \figref{fig:app_diag_position} is identifiable locally but appears at bottom-left instead of the requested bottom-right. Its wording and placement therefore require different views of the same image. In \figref{fig:app_diag_integration}, lettering follows the wooden sign's plane, lighting, and texture. Carrier fit and exact transcription describe different properties of the rendered text.

The examples also retain disagreements with Q-Judger. Target-related chat text is visible in \figref{fig:app_diag_layout} despite zero accuracy and completeness ratings, while \figref{fig:app_diag_high_score} contains degraded small characters despite maximum ratings. These selected cases describe local errors without estimating their frequency. The leaderboard retains the automatic scores; a score of 100 is the upper end of the rubric rather than a measured percentage of correct characters.

\Needspace{6\baselineskip}
\section{Experiments}
\label{sec:experiments}

The experiments compare fidelity and clarity across 24 model configurations, examine differences across workloads and languages, and assess Base/Turbo variants. We also inspect maximum ratings and the effect of composite weights on the reported ranking.

\subsection{Experimental Setup}
\label{sec:exp_setup}

\paragraph{Models and sampling.}
We evaluated 24 model configurations from the FLUX, Stable Diffusion, Hunyuan, HiDream, Qwen-Image~\citep{wu2025qwen,zhao2026qwen}, Z-Image, Boogu-Image, WAN, Nano Banana, GPT Image, LLaDA~\citep{LLaDAImage}, and Seedream families, listed in \tabref{tab:leaderboard}.
Base/Turbo pairs allow comparison of standard and accelerated variants. Each model uses its standard sampling configuration; resolution, aspect ratio, and API behavior can differ. The results compare these configurations rather than isolate a training or acceleration method.

\paragraph{Evaluation and reporting.}
Q-Judger applies the six-score rubric with the settings in \appref{app:reference_payload}. Rankings use these VLM ratings and the aggregation in \secref{sec:scoring}.
For each prompt, we average its valid image scores; prompt-macro aggregation then gives each represented prompt equal weight. The bilingual result averages the EN and ZH prompt-macro means equally.
Prompts with no valid score contribute only to coverage denominators. Image coverage is the fraction of planned images with valid scores; prompt coverage is the fraction of prompts with at least one valid score.
\appref{app:protocol} describes the workflow; \appref{app:diagnostic_scope} identifies the available records and their limits. We report descriptive means without confidence intervals. Images from a shared prompt are related observations, so close comparisons require uncertainty estimates across prompts.

\subsection{Benchmark Results}
\label{sec:exp_main}

\begin{table*}[!t]
\centering
\caption[UltraText Bench leaderboard]{\textbf{UltraText Bench leaderboard.} VLM ratings range from 0 to 100 (higher is better). Overall columns average EN/ZH prompt-macro means equally; level columns show language-specific composites. Composite weights are 60/30/5/5 for Fidelity/Clarity/Spatial/Scene (\tabref{tab:dimensions}). L1/L2/L3 denote Hard/Very Hard/Extreme; bracketed Low/High labels are API quality settings. Bold marks column bests within each group; shading marks the Composite leader. Execution failures affect coverage, not quality means (\secref{sec:exp_setup}).}
\label{tab:leaderboard}
\vspace{0.3em}
\begingroup
\fontsize{8.5}{10.5}\selectfont
\setlength{\tabcolsep}{2pt}
\renewcommand{\arraystretch}{1.12}
\begin{tabularx}{\textwidth}{@{}>{\raggedright\arraybackslash}Xrrrrr!{\hspace{2pt}{\color{utNeutral!55}\vrule width 0.6pt}\hspace{2pt}}rrrrrr@{}}
\toprule
\rowcolor{utLight}
\textbf{Model} & \textcolor{utAccent}{\textbf{Composite}} & \textbf{Fidelity} & \textbf{Clarity} & \textbf{Spatial} & \textbf{Scene}
& \multicolumn{2}{c}{\textbf{L1}}
& \multicolumn{2}{c}{\textbf{L2}}
& \multicolumn{2}{c}{\textbf{L3}} \\
\cmidrule(lr){7-8}\cmidrule(lr){9-10}\cmidrule(lr){11-12}
& & & & & & \textcolor{utEN}{\textbf{EN}} & \textcolor{utZH}{\textbf{ZH}} & \textcolor{utEN}{\textbf{EN}} & \textcolor{utZH}{\textbf{ZH}} & \textcolor{utEN}{\textbf{EN}} & \textcolor{utZH}{\textbf{ZH}} \\
\midrule
\rowcolor{utLight}
\multicolumn{12}{@{}l}{\textbf{Open-weight models}} \\
\rowcolor{utAccent!7}
Boogu-Image-0.1-Base     & \textbf{77.89} & \textbf{75.77} & 78.31 & \textbf{88.04} & \textbf{90.53} & 84.16 & 87.75 & \textbf{76.42} & \textbf{78.11} & \textbf{57.67} & \textbf{83.23} \\
Qwen-Image-2512          & 67.96 & 59.30 & \textbf{79.89} & 79.13 & 88.93 & 86.50 & \textbf{89.47} & 71.57 & 67.44 & 42.86 & 49.90 \\
Boogu-Image-0.1-Turbo    & 67.79 & 66.27 & 64.86 & 84.37 & 86.90 & 73.13 & 81.30 & 62.81 & 72.61 & 37.24 & 79.63 \\
Z-Image-Base             & 61.92 & 55.51 & 70.60 & 70.88 & 77.72 & 82.92 & 84.66 & 67.54 & 64.41 & 26.34 & 45.64 \\
Z-Image-Turbo            & 54.38 & 40.75 & 74.41 & 69.41 & 82.57 & 79.72 & 73.66 & 57.42 & 52.36 & 25.14 & 37.99 \\
HiDream-O1-Image         & 45.56 & 33.26 & 62.56 & 60.51 & 76.10 & 80.95 & 55.33 & 51.51 & 27.80 & 29.53 & 28.23 \\
LLaDA-Image              & 43.92 & 31.65 & 59.87 & 64.76 & 74.51 & 62.96 & 54.72 & 47.40 & 33.44 & 31.23 & 33.78 \\
LLaDA-Image-Turbo        & 41.29 & 24.03 & 67.24 & 60.75 & 73.12 & 57.93 & 39.69 & 51.34 & 31.29 & 36.61 & 30.89 \\
Qwen-Image               & 41.27 & 33.75 & 50.07 & 52.87 & 67.10 & 73.83 & 64.39 & 42.45 & 32.66 & 17.61 & 16.69 \\
FLUX.2 [Dev]             & 41.03 & 27.86 & 60.81 & 51.91 & 69.42 & \textbf{86.79} & 33.29 & 63.61 & 14.97 & 35.66 & 11.84 \\
HiDream-O1-Image-Dev     & 40.44 & 24.58 & 63.06 & 58.69 & 76.70 & 69.76 & 46.51 & 45.29 & 26.66 & 29.27 & 25.12 \\
Hunyuan-Image-3.0        & 29.93 & 20.43 & 39.32 & 51.83 & 65.55 & 47.86 & 37.63 & 33.74 & 22.81 & 17.46 & 20.08 \\
FLUX.2 [Klein] 9B        & 22.94 & 14.60 & 32.02 & 36.86 & 54.54 & 59.96 & 11.64 & 43.33 & 5.09 & 13.69 & 3.91 \\
FLUX.2 [Klein] 4B        & 19.58 & 10.86 & 29.66 & 31.98 & 51.37 & 56.17 & 7.68 & 35.62 & 4.54 & 9.84 & 3.65 \\
FLUX.1 [Dev]             & 18.17 & 0.93 & 49.19 & 17.52 & 39.68 & 24.88 & 12.17 & 23.31 & 16.12 & 18.63 & 13.92 \\
SD3.5 Large              & 7.48 & 0.35 & 18.53 & 9.04 & 25.23 & 18.48 & 1.22 & 13.20 & 1.91 & 9.14 & 0.94 \\
\midrule
\rowcolor{utLight}
\multicolumn{12}{@{}l}{\textbf{API-only models}} \\
\rowcolor{utAccent!7}
GPT Image 2 [Low]                  & \textbf{99.35} & \textbf{99.23} & \textbf{99.52} & \textbf{99.41} & \textbf{99.57} & \textbf{99.73} & \textbf{99.32} & \textbf{99.58} & \textbf{99.48} & \textbf{99.36} & \textbf{98.61} \\
Nano Banana 2                       & 97.00 & 95.90 & 98.68 & 98.16 & 99.00 & 98.82 & 96.46 & 97.32 & 95.48 & 96.82 & 97.12 \\
Seedream 5.0 Pro                    & 91.45 & 88.92 & 94.79 & 95.73 & 97.33 & 97.72 & 97.60 & 93.63 & 93.58 & 75.36 & 90.81 \\
Qwen-Image-2.0-Pro                  & 90.21 & 89.69 & 89.78 & 94.18 & 95.14 & 95.74 & 94.49 & 85.60 & 92.89 & 80.59 & 91.99 \\
Qwen-Image-3.0                      & 88.33 & 87.81 & 87.63 & 93.15 & 93.97 & 91.94 & 86.33 & 88.74 & 86.91 & 85.05 & 91.03 \\
WAN-2.7-Image-Pro                  & 85.73 & 79.72 & 94.84 & 92.84 & 95.85 & 94.98 & 90.14 & 90.03 & 81.46 & 75.61 & 82.14 \\
GPT Image 1.5 [High]               & 57.96 & 48.86 & 69.69 & 72.37 & 82.31 & 97.28 & 39.18 & 92.32 & 23.54 & 77.21 & 18.22 \\
GPT Image 1 [High]                 & 37.91 & 22.91 & 60.01 & 52.20 & 70.85 & 75.81 & 30.49 & 61.37 & 12.79 & 37.92 & 9.06 \\
\bottomrule
\end{tabularx}
\endgroup
\end{table*}

\tabref{tab:leaderboard} reports equal-language prompt-macro means. Its first five columns aggregate prompt-level scores directly, rather than averaging rounded level cells.

\paragraph{Text fidelity and clarity.}
Clear text can still differ from the requested content. Z-Image-Turbo receives 74.41 Clarity and 40.75 Fidelity; Qwen-Image-2512 receives 79.89 and 59.30. Their scene-quality scores are also higher than their fidelity scores. High clarity and scene-quality ratings therefore coexist with substantially lower ratings for the requested text content. \figref{fig:app_diag_accuracy} illustrates this distinction with malformed text on a plausible sign.
Among open-weight models, Boogu-Image-0.1-Base has higher Fidelity than Qwen-Image-2512 (75.77 versus 59.30), while Qwen-Image-2512 has slightly higher Clarity (79.89 versus 78.31). The model with the clearest rated text therefore need not be the one that best preserves the requested content.

\paragraph{Performance across difficulty levels.}
Qwen-Image-2512's Composite falls from 86.50 at L1 to 42.86 at L3 in EN, and from 89.47 to 49.90 in ZH. Z-Image-Base likewise falls from 82.92 to 26.34 in EN. The higher-workload groups expose weaknesses less apparent at L1. The trend is not universal: Boogu-Image-0.1-Base's ZH score rises from 78.11 at L2 to 83.23 at L3. Levels group different prompts with changing text loads and region counts; they do not isolate the effect of length or guarantee monotonically decreasing scores.

\paragraph{Base and Turbo models.}
Under reported settings, Turbo Fidelity is lower than Base by 14.76 points for Z-Image, 7.62 for LLaDA-Image, and 9.50 for Boogu-Image-0.1. Clarity rises by 3.81 and 7.37 points for Z-Image and LLaDA-Image, but falls by 13.45 for Boogu-Image-0.1. Fidelity decreases across all three pairs, while the direction of the clarity change depends on the model family. Differences in standard settings prevent attributing these changes to acceleration alone.
Recent advances in one-step generation~\citep{cheng2026twinflow,liu2026self,sun2026three,sun2026duality} make it useful to assess both dimensions when evaluating few-step models on dense text.

\paragraph{English and Chinese.}
Language differences depend on the model. At L3, Boogu-Image-0.1-Base scores 57.67 in EN and 83.23 in ZH, whereas GPT Image 1.5 [High] scores 77.21 and 18.22. These are comparisons between independently written prompt sets. EN L3 averages 2,386.68 GT characters and ZH L3 averages 743.62, so language, content, and workload all contribute to the comparison.

\subsection{Strong Models and Evaluation Coverage}
\label{sec:exp_judge}

GPT Image 2 [Low] leads the Composite at 99.35. Of its 1,723 valid images, 1,440 (83.6\%) receive 100 on all six raw dimensions. Its EN and ZH scores remain high across all three levels, while many other configurations decline sharply at L3.

\Needspace{5\baselineskip}
\figref{fig:prompt_examples} shows its L3 outputs. \appref{app:diagnostics} includes local disagreements such as the maximum-rated code image in \figref{fig:app_diag_high_score}. \appref{app:benchmark_value} discusses implications for model assessment.

Image coverage is 859/864 in EN and 864/864 in ZH. Four EN safety rejections affect one prompt and the fifth affects another, yielding prompt coverage of 215/216 and 216/216. Entirely unscored prompts contribute to coverage but not quality means.

\paragraph{Sensitivity to the composite weights.}
GPT Image 2 [Low] leads all four reported means and hence any nonnegative weighted average of them. Boogu-Image-0.1-Base remains the open-weight leader under equal weights, fidelity-only weighting, and 40/30/15/15, as well as the default 60/30/5/5. Close rankings can depend on the tradeoff: Qwen-Image-2512's default Composite is 67.96 versus Boogu-Image-0.1-Turbo's 67.79, while the latter has higher Fidelity. Reweighting these means addresses neither prompt-sampling uncertainty nor judge reliability.

\section{Discussion}
\label{sec:discussion}

\paragraph{Why dense text is different from ordinary image quality.}
Diffusion models~\citep{ho2020denoising} operate in continuous visual spaces where approximate similarity is usually acceptable: small geometric or texture errors often preserve the intended object.
Text is different because it is discrete and compositional; one wrong character can invalidate a word, code token, phone number, price, or form field.
Self-attention in DiT~\citep{peebles2023scalable} and UNet~\citep{rombach2022high} architectures must split capacity between global composition and many local character patterns, and dense prompts spread this burden across multiple text regions.
Text-annotated image pairs are rarer than ordinary image-caption pairs in web corpora~\citep{Dalle-3}, and incidental scene text is often noisy or weakly described.
CLIP~\citep{radford2021learning} and T5~\citep{raffel2020exploring} use subword tokenization, while accurate text rendering requires preserving the exact character sequence.
VAE latents can lose small stroke details, a limitation also emphasized by OCRGenBench~\citep{zhang2025ocrgenbench}.
Visual-tokenizer evaluations further show that standard reconstruction metrics can conceal losses in text-bearing regions~\citep{wu2025tokbench}.
These mechanisms may contribute to differences between text fidelity, layout, and scene quality. Their causal effects remain untested here.

\paragraph{Training directions for stronger text rendering.}
Potential training directions for dense text rendering include visual representation warmup, as in ERW~\citep{liu2025efficient}, and optimization with text-specific feedback.
The structured annotations can support region-level diagnosis and feedback on character errors, omissions, placement, and scene integration. Such feedback can guide reward-based optimization, including DDPO~\citep{black2024training}, ReFL~\citep{xu2023imagereward}, and AlignProp~\citep{prabhudesai2023aligning}, or the construction of preference pairs used by methods such as Diffusion-DPO~\citep{wallace2024diffusion} and FlowCPO~\citep{han2026flowcpo}.
Although \texttt{importance} does not enter our scoring formulas, future training objectives could use it to weight errors in a shop name more heavily than errors in a footnote.
This direction is aligned with region-level feedback work such as RichHF-18K~\citep{liang2024rich}, but focuses specifically on visual text rendering.

\paragraph{Complementarity with concurrent benchmarks.}
UltraText Bench combines scene coverage and per-region references.
STRICT~\citep{zhang2025strict} also evaluates thousands of characters, but varies one sequence on a plain page; UltraText Bench distributes text over 4--12 annotated regions with carriers and positions across 24 categories.
OCRGenBench~\citep{zhang2025ocrgenbench} spans several OCR-generative tasks, including dense pages, while UltraText Bench focuses on prompt-only generation.
InfoTextBench~\citep{xiang2026glyphanchor} is the closest comparison for dense bilingual text, but its target-text items differ from our regions, preventing direct comparison of their counts.

\paragraph{Limitations.}
Results reflect each model's standard generation settings, which can differ in resolution, aspect ratio, and API behavior. A controlled comparison of these factors remains future work.
Each category$\times$level$\times$language cell contains 3 prompts and 12 images in a complete run. A language--level aggregate contains 72 prompts, and a language aggregate contains 216. Repeated images share a prompt, so prompt-level variation is the relevant unit for comparing aggregate scores. We report descriptive means without confidence intervals; close ranks and fine-grained cell differences require correspondingly cautious interpretation.
The protocol returns image-level scores from one VLM judge. Ten participants took part in human evaluation of the automatic scores; \appref{app:human_alignment} discusses its scope and limitations. Quantitative inter-rater agreement and alignment with human ratings are not reported here. The per-region references specify the content and visual context for assessment, while the current output summarizes each image as a whole.
Maximum scores are frequent in the GPT Image 2 records, and selected cases show
both low-score disagreements and visible errors despite maximum ratings.
The examples distinguish local character defects from an image's overall rating;
they do not quantify the judge's error rate. A maximum rating is not a measured
percentage of correct characters. Strong aggregate performance and occasional
local disagreements therefore need to be interpreted at their respective scales.
The language splits use independently written content and different character-load
bands, and difficulty levels change text load and region count together.
Their comparisons therefore do not isolate language or text length alone.

\paragraph{Future work.}
Future work will explore region-level evaluation and structured annotations as training signals. Extending the benchmark to video generation and editing systems such as Vidu S2~\citep{zhang2026vidu} would enable evaluation of text persistence and temporal consistency.

\Needspace{4\baselineskip}
\section{Conclusion}
\label{sec:conclusion}

UltraText Bench evaluates dense bilingual text with complete region references across 24 scene categories. Comparisons of 24 model configurations distinguish fidelity from clarity and expose differences across workloads and languages. The suite supports assessment of sustained text reproduction, placement, and integration with the surrounding scene.

\clearpage
\bibliography{resources/reference}
\bibliographystyle{configuration/iclr2027_conference}

\clearpage
\appendix

\clearpage
\raggedbottom
\setcounter{figure}{0}
\setcounter{table}{0}
\renewcommand{\thefigure}{A\arabic{figure}}
\renewcommand{\thetable}{A\arabic{table}}
\renewcommand{\theHfigure}{appendix.\arabic{figure}}
\renewcommand{\theHtable}{appendix.\arabic{table}}

\noindent\textbf{Appendix guide.}
\appref{app:construction} describes how the prompts were built, what each annotation field means, and what every category asks for.
\appref{app:prompt_examples} reproduces one record in full and follows a single scene family across the three levels.
\appref{app:judge_execution} gives the judge prompt verbatim, the reference payload that accompanies it, and the workflow that turns an image into a scored row.
\appref{app:rubric} sets out how the six raw scores become four reporting dimensions and one composite, worked through on one recorded response.
\appref{app:reference_role} explains how region references support the existing scores and how to interpret performance as model capabilities improve.
\appref{app:diagnostics} reads seven generated images against their references and collects the per-category galleries.
\appref{app:diagnostic_scope} records which artifacts exist for this revision and what they cannot establish.
\appref{app:human_alignment} describes the scope of human evaluation and the interpretation of qualitative examples.

\section{Dataset Construction and Annotation}
\label{app:construction}

All 432 prompts were drafted with Claude Opus 4.6 and then reviewed by hand, entry by entry, over several rounds of revision.
Construction proceeded in five stages.

First, we studied 160 seed prompts from X-Omni LongText-Bench to guide scene descriptions, target-string presentation, and region annotation.
Three lessons carried into our own design: how a scene description should be organized, how a target string can be placed inside that description without announcing itself as a target, and how finely a scene should be divided into annotated regions.
We settled on embedding each string where it naturally belongs, inside quotation marks, a code fence, or an explicit structural block as the scene requires, on describing every string together with the surface it is written on, and on varying the scene context from one prompt to the next.

Second, we wrote 9 prompts for each of the 24 categories in each language, 3 at every difficulty level, which yields the full $24\times3\times3\times2=432$ design.
We generated one category at a time so that prompts within a category would not echo each other: no two share a business name, a location, or a content theme.
The English and Chinese sets were written independently and adapted to their own cultural setting rather than translated from one another.

Third, automated scripts checked every record against the specification.
Each prompt's GT-character load was recomputed as $\sum_r |\texttt{gt\_regions}[r].\texttt{text}|$ and compared against the design band for its language and level; \figref{fig:difficulty_stats} plots those bands against the realized distributions.
All 432 released records fall inside their own band under this definition.
\tabref{tab:stats} reports the resulting ranges.
The scripts further verified that every region carries an identifier and all six annotated fields, and that every GT string appears verbatim somewhere in its own prompt.
Most strings are set off by quotation marks, but code and other structured content may instead sit in a fenced or explicitly labeled block, so the check does not insist on a single quoting convention.

Fourth, we manually revised every prompt to integrate target strings into the scene description. Repeated instructions such as ``The image must contain the following text'' were replaced by descriptions of the text on its intended carrier, with all target strings preserved.

Fifth, we reviewed cross-lingual difficulty against the realized region-text distributions and against what each language needs in order to describe a comparable scene.
No OCR measurement acts as a release gate or as a ranking signal.
Every current record satisfies the character band for its own language and level under the region-text sum above, so EN--ZH comparisons should be read from the reported distributions rather than from an OCR-derived score.

\paragraph{Cultural adaptation examples.}
The following excerpts from two independently authored L1 menu records, \texttt{C1\_L1\_EN\_001} and \texttt{C1\_L1\_ZH\_001}, illustrate cultural adaptation; all cells are verbatim strings from the released JSONL:

\begin{center}
\small
\setlength{\tabcolsep}{4pt}
\renewcommand{\arraystretch}{1.15}
\begin{tabularx}{\textwidth}{@{}l>{\hsize=1.12\hsize\linewidth=\hsize\raggedright\arraybackslash}X>{\hsize=0.88\hsize\linewidth=\hsize\raggedright\arraybackslash}X@{}}
\toprule
\rowcolor{utLight}
\textbf{Element} & \textbf{EN} & \textbf{ZH} \\
\midrule
Venue & BREW \& BLOOM COFFEE HOUSE & 老街坊茶餐厅 \\
Heritage line & Specialty Roasters Since 2015 & 始于1994年 · 地道港式味 \\
Price & Cappuccino \$5.25 & 干炒牛河 ¥38 \\
Menu item & Brown Sugar Shaken Espresso & 菠萝油配丝袜奶茶 \\
Ordering & Extra Shot +\$0.75 & 外带请扫码点单 · 提前预约 \\
\bottomrule
\end{tabularx}
\end{center}

\subsection{Annotation Schema}

A generator receives the natural-language \texttt{prompt} and nothing else.
The \texttt{gt\_regions} array travels with the record as evaluation metadata and is never handed to the model as an additional layout input.
Every region carries an \texttt{id} together with six attributes: its exact \texttt{text}, a \texttt{position}, a \texttt{size}, a \texttt{type}, a \texttt{carrier}, and an \texttt{importance}.
Three of these draw on closed vocabularies: position names one of nine grid cells, size is large, medium, small, or tiny, and importance is high, medium, or low.
Type and carrier are descriptive labels for the role the text plays and the surface it is written on.
Carriers include digital interface elements as well as physical surfaces. Size labels describe qualitative text scale without pixel or font-size thresholds.
Importance enters no scoring formula, and the rubric prohibits ignoring regions because they have lower importance. The 2,926 released regions contain 36 distinct type labels.

The grid records a coarse location and nothing more: it carries no bounding box, no extent, and no ordering of regions within a cell.
Region counts follow the supplied annotation entries, not OCR detections or line breaks: one region may contain a multi-line passage, and several regions may share a cell.
Character load is the sum of the lengths of all region strings, spaces and punctuation included, which is neither a word count nor the length of the generation prompt.
\appref{app:prompt_examples} shows both conventions on a complete record.

\paragraph{Release format and placeholders.}
Each JSONL record contains one prompt and its structured GT, so record-level and prompt-level counts coincide.
The fixed placeholder vocabulary includes \texttt{TELCODE\_SAFE}, \texttt{TELCODE-SAFE}, \texttt{SAFE-CODE}, \texttt{WEBLINK\_SAFE}, and \texttt{HANDLECODE\_SAFE}, together with per-record indexed telephone forms such as \texttt{TELCODE-A0001}.

\paragraph{Illustrative prompt revision.}
A mechanical instruction such as ``The image must contain the following text: DAILY BREAD'' becomes ``A walnut storefront sign bears the cream serif lettering `DAILY BREAD'.''
The target string is untouched; what changes is that the sentence now names the surface the string sits on.
We constructed this pair to show the rule, and it is not an entry recovered from the unavailable review ledger.

\subsection{Category Definitions}
\label{app:categories}
The six domains below expand the 24 categories of \tabref{tab:categories}.
Each describes a family of scenes rather than a fixed template, and in all of them the text is supplied for reproduction: prices, totals, and other figures are given in the prompt and never have to be inferred or calculated.
\paragraph{Domain A: Signage and Labels.}\mbox{}\par\smallskip
\noindent\textbf{A1 Sign} puts text on physical signage seen at an angle, in outdoor light and on weathered material.\par\smallskip
\noindent\textbf{A2 Label} asks for small type on compact or curved surfaces such as product bottles and price tags.\par\smallskip
\noindent\textbf{A3 Poster} needs several levels of typographic hierarchy to hold up against a busy visual background.\par\smallskip
\noindent\textbf{A4 Billboard} renders large-format text at a viewing angle and under environmental effects.\par\smallskip

\paragraph{Domain B: Documents and Print.}\mbox{}\par\smallskip
\noindent\textbf{B1 Article} runs extended prose across paragraph breaks and line wraps, justified or left-aligned.\par\smallskip
\noindent\textbf{B2 Newspaper} fits several articles onto one page in multiple columns under a headline hierarchy.\par\smallskip
\noindent\textbf{B3 Letter} follows the conventional form of a formal letter: salutation, body paragraphs, and closing.\par\smallskip
\noindent\textbf{B4 Resume} arranges sections, bullet points, and aligned dates, often across two columns.\par\smallskip

\paragraph{Domain C: Commercial.}\mbox{}\par\smallskip
\noindent\textbf{C1 Menu} aligns items with their prices across several sections, each carrying its own descriptions.\par\smallskip
\noindent\textbf{C2 Receipt} places purchase details, subtotals, and totals on narrow thermal-print lines.\par\smallskip
\noindent\textbf{C3 Invoice} arranges quantities, unit prices, and totals in a table beside issuer and recipient details.\par\smallskip
\noindent\textbf{C4 Product Packaging} wraps ingredient lists and nutrition tables around a surface seen in perspective.\par\smallskip

\paragraph{Domain D: Digital Interfaces.}\mbox{}\par\smallskip
\noindent\textbf{D1 Webpage} combines navigation, content panels, sidebars, and footers at different type sizes.\par\smallskip
\noindent\textbf{D2 Slide} keeps a clean title-and-bullet structure with consistent indentation and a footer.\par\smallskip
\noindent\textbf{D3 Social Media} surrounds a post or feed with interface chrome, hashtags, timestamps, and engagement counts; conversational turn-taking belongs to Dialogue instead.\par\smallskip
\noindent\textbf{D4 Dashboard} packs KPI cards, chart labels, and table grids densely into a monitoring interface; Infographic, by contrast, organizes an explanatory narrative.\par\smallskip

\paragraph{Domain E: Structured Data.}\mbox{}\par\smallskip
\noindent\textbf{E1 Schedule} holds a time column in alignment and keeps formatting consistent from row to row.\par\smallskip
\noindent\textbf{E2 Form} pairs labels with fields and adds checkboxes, section dividers, and instruction text.\par\smallskip
\noindent\textbf{E3 Certificate} centers formal typography at several sizes and depends on symmetry.\par\smallskip
\noindent\textbf{E4 Code} requires a consistent monospace face, exact indentation, and recognizable syntax elements.\par\smallskip

\paragraph{Domain F: Creative and Special.}\mbox{}\par\smallskip
\noindent\textbf{F1 Caption} requires readable text over a photographic or illustrated background.\par\smallskip
\noindent\textbf{F2 Dialogue} alternates chat bubbles from one side to the other with speaker labels attached.\par\smallskip
\noindent\textbf{F3 Comic Panel} mixes speech bubbles, sound-effect words, and narrative caption boxes.\par\smallskip
\noindent\textbf{F4 Infographic} scatters data annotations, axis labels, and legends across a freeform layout.\par\smallskip

\clearpage
\section{Input Examples and Difficulty}
\label{app:prompt_examples}
This appendix reproduces the prompt and reference of one record, also used in \appref{app:worked_score}. IDs encode category, level, language, and entry number; sample numbers identify repeated generations.

\subsection{Complete English Sign Example}
\label{app:complete_example}
\textbf{\textcolor{utEN}{\texttt{A1\_L1\_EN\_001}}} is a Sign record at L1 whose four regions carry 423 GT characters between them.
The prompt below and the reference strings that follow are quoted exactly as released, and both are also supplied in \path{case_records.json} (\appref{app:diagnostic_scope}).

\begin{mdframed}[backgroundcolor=utEN!3,linecolor=utEN!45,linewidth=0.5pt,
innerleftmargin=8pt,innerrightmargin=8pt,innertopmargin=6pt,innerbottommargin=6pt]
\small\raggedright
\textbf{Generator input: complete prompt}\par\smallskip
A warm, golden-hour photograph of a charming neighborhood bakery on a quiet cobblestone street in a European-style district. Late afternoon sunlight casts long amber shadows across the worn stone pavement, highlighting the weathered red brick facade partially veiled by climbing ivy whose leaves shimmer in shades of emerald and gold. The storefront features a dark walnut wood frame with large multi-paned glass windows, through which rustic wooden shelves stacked with freshly baked loaves, braided challah, and dusted boules are visible. A soft glow from vintage Edison bulbs within spills warmly onto the sidewalk, mingling with the natural light.

Above the entrance, a hand-carved wooden shop sign --- approximately 120cm wide and 40cm tall --- hangs from ornate wrought-iron scrollwork brackets that curl into delicate leaf motifs. The sign itself has a deep walnut stain finish with gently distressed edges suggesting decades of faithful service, and features cream-white hand-painted serif lettering in a Playfair Display style, with characters roughly 6cm tall, that reads: "DAILY BREAD --- Artisan Bakery \& Fine Patisserie Since 1987"

Directly beneath the main sign, a smaller rectangular oak plank with softly rounded edges and a honey-toned varnish is suspended by two short brass chains. Pressed into its surface in copper-foil italic lettering with a warm amber sheen, it displays: "Handcrafted with Love --- Organic Sourdough · French Pastries · Seasonal Specials · Wood-Fired Stone Oven"

To the left of the entrance, positioned on the cobblestones beside a terracotta pot overflowing with lavender, stands a freestanding A-frame chalkboard easel with a weathered pine frame. The chalkboard surface is covered in expressive hand-drawn lettering in white and dusty rose chalk, with small decorative wheat-stalk doodles in the corners. The board announces: "Today's Fresh Bakes: Walnut Rye Loaf \$6.50 | Butter Croissants \$3.25 | Cinnamon Cardamom Rolls \$4.00 | Olive Focaccia \$5.75 --- Arrive Early, We Sell Out by Noon!"

On the right side of the doorframe, at eye level beside a polished brass door handle, a small rectangular brass plate with beveled edges catches the last rays of sunlight. Engraved in precise, dark serif lettering, it reads: "Open Tuesday--Sunday 6:30 AM -- 2:00 PM | Closed Mondays | Custom Cake Orders Welcome: Call TELCODE-A0001"
\end{mdframed}

\subsection{Complete Region Reference}
All four regions appear below with every annotated field filled in.
The generator sees only the prompt above; these fields go to the judge instead.
\Needspace{4\baselineskip}
\begin{mdframed}[backgroundcolor=utEN!3,linecolor=utEN!35,linewidth=0.4pt,innerleftmargin=7pt,innerrightmargin=7pt,innertopmargin=4pt,innerbottommargin=4pt,skipabove=4pt,skipbelow=4pt]
\footnotesize\raggedright
\textbf{\texttt{region\_0}}\quad \texttt{position}: top-center; \texttt{size}: large; \texttt{type}: title.\par
\texttt{carrier}: wooden shop sign; \texttt{importance}: high.\par\smallskip
\texttt{text}: DAILY BREAD --- Artisan Bakery \& Fine Patisserie Since 1987
\end{mdframed}
\Needspace{4\baselineskip}
\begin{mdframed}[backgroundcolor=utEN!3,linecolor=utEN!35,linewidth=0.4pt,innerleftmargin=7pt,innerrightmargin=7pt,innertopmargin=4pt,innerbottommargin=4pt,skipabove=4pt,skipbelow=4pt]
\footnotesize\raggedright
\textbf{\texttt{region\_1}}\quad \texttt{position}: top-center; \texttt{size}: medium; \texttt{type}: subtitle.\par
\texttt{carrier}: oak plank sign; \texttt{importance}: medium.\par\smallskip
\texttt{text}: Handcrafted with Love --- Organic Sourdough · French Pastries · Seasonal Specials · Wood-Fired Stone Oven
\end{mdframed}
\Needspace{4\baselineskip}
\begin{mdframed}[backgroundcolor=utEN!3,linecolor=utEN!35,linewidth=0.4pt,innerleftmargin=7pt,innerrightmargin=7pt,innertopmargin=4pt,innerbottommargin=4pt,skipabove=4pt,skipbelow=4pt]
\footnotesize\raggedright
\textbf{\texttt{region\_2}}\quad \texttt{position}: middle-left; \texttt{size}: medium; \texttt{type}: body.\par
\texttt{carrier}: chalkboard easel; \texttt{importance}: medium.\par\smallskip
\texttt{text}: Today's Fresh Bakes: Walnut Rye Loaf \$6.50 | Butter Croissants \$3.25 | Cinnamon Cardamom Rolls \$4.00 | Olive Focaccia \$5.75 --- Arrive Early, We Sell Out by Noon!
\end{mdframed}
\Needspace{4\baselineskip}
\begin{mdframed}[backgroundcolor=utEN!3,linecolor=utEN!35,linewidth=0.4pt,innerleftmargin=7pt,innerrightmargin=7pt,innertopmargin=4pt,innerbottommargin=4pt,skipabove=4pt,skipbelow=4pt]
\footnotesize\raggedright
\textbf{\texttt{region\_3}}\quad \texttt{position}: bottom-right; \texttt{size}: small; \texttt{type}: data.\par
\texttt{carrier}: brass plate; \texttt{importance}: low.\par\smallskip
\texttt{text}: Open Tuesday--Sunday 6:30 AM -- 2:00 PM | Closed Mondays | Custom Cake Orders Welcome: Call TELCODE-A0001
\end{mdframed}

\Needspace{10\baselineskip}
\subsection{Chinese Adaptation with the Same Fields}
\textbf{\textcolor{utZH}{\texttt{A1\_L1\_ZH\_001}}} is the Chinese Sign record at the same level, with 362 GT characters spread over five regions: a wooden sign, an oak plank, a chalkboard, a brass plate, and a window sticker.
The fifth carrier and the RMB prices on the bakery items are what adapt the scene to Chinese usage, and neither has a counterpart in the English record.

\begin{mdframed}[backgroundcolor=utZH!3,linecolor=utZH!45,linewidth=0.5pt,
innerleftmargin=8pt,innerrightmargin=8pt,innertopmargin=6pt,innerbottommargin=6pt]
\small\raggedright
\textbf{One complete ZH region; the full record is in the accompanying JSON.}\par
\texttt{id}: \texttt{region\_4}; \texttt{position}: bottom-left; \texttt{size}: small; \texttt{type}: data; \texttt{carrier}: window sticker; \texttt{importance}: low.\par\smallskip
\texttt{text}: 本店荣获2023年度样例社区烘焙坊 | 样例安全等级：A级 | 备案码：SAFE-CODE-031587
\end{mdframed}

The two records share a schema and little else: they were authored separately, hold different numbers of regions, and describe different content.
Placeholders such as the registration code above are literal target strings, and a model is expected to reproduce them character for character.

\subsection{Three Levels within One Scene Family}
\label{app:level_examples}
The three English newspaper outputs below hold category, model setting, and sample index fixed and differ in level.
Their content and layouts differ as well, so what they illustrate is the progression from one level to the next rather than a controlled change in text length alone.
Every load quoted here is recomputed from the GT regions.

\begin{figure}[!t]
\centering
\begin{minipage}[t]{0.32\linewidth}\centering
\includegraphics[width=\linewidth]{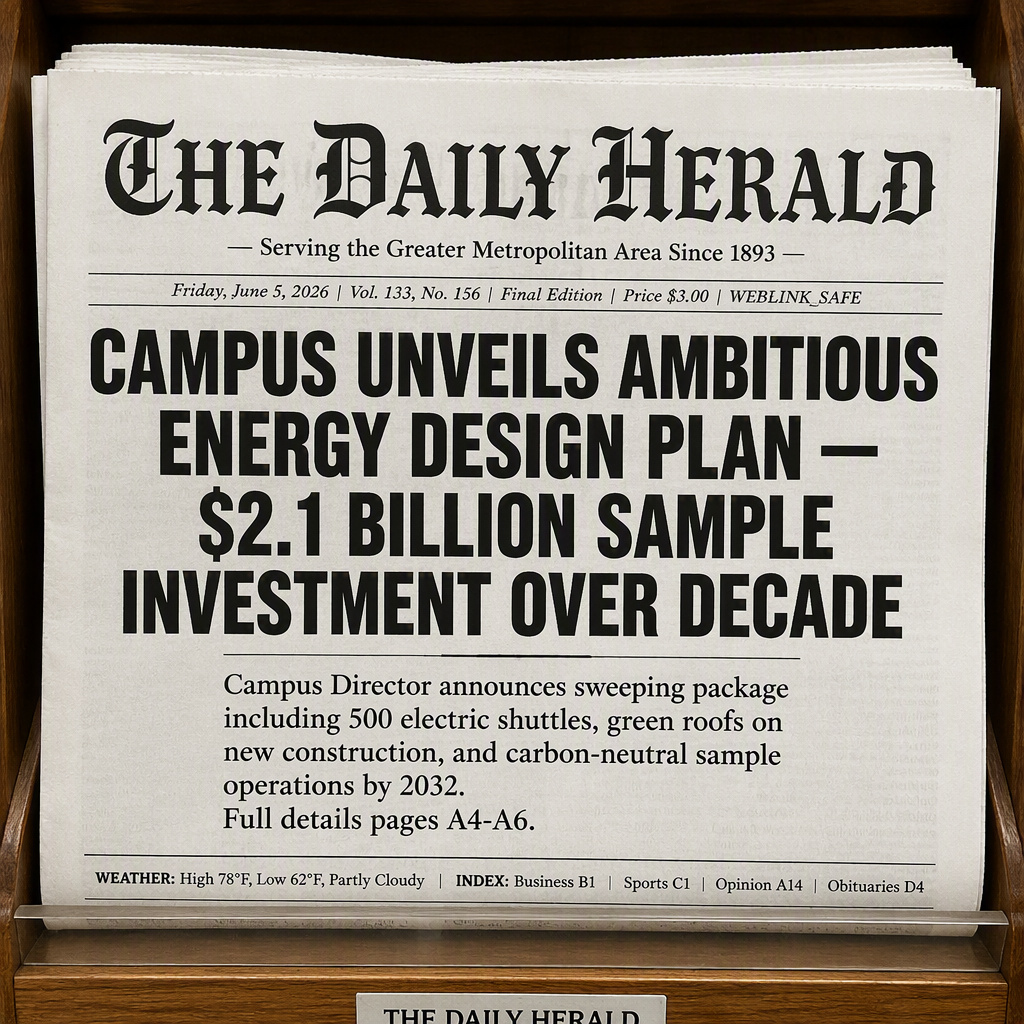}
\footnotesize\textbf{\textcolor{utEN}{L1}}: 526 characters; 5 regions\\
\scriptsize\texttt{B2\_L1\_EN\_001}\end{minipage}\hfill
\begin{minipage}[t]{0.32\linewidth}\centering
\includegraphics[width=\linewidth]{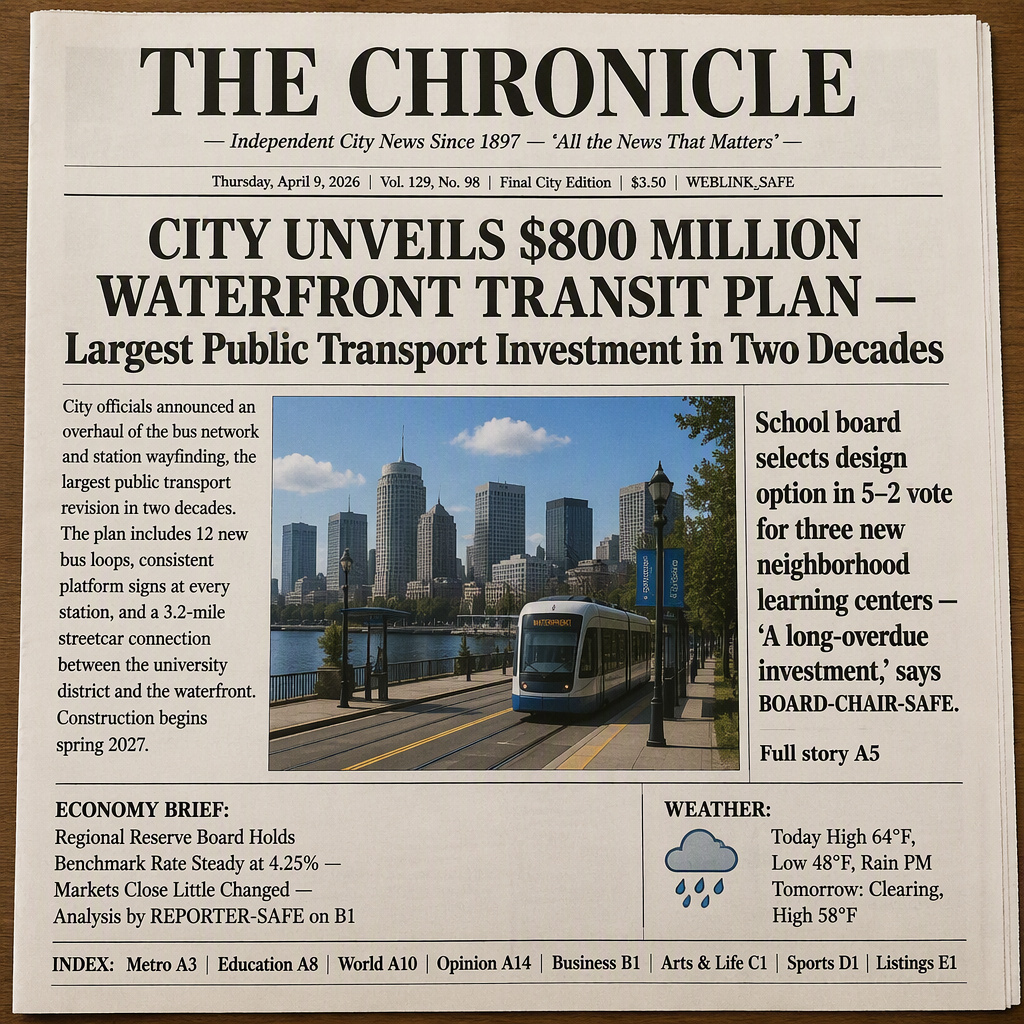}
\footnotesize\textbf{\textcolor{utEN}{L2}}: 1,085 characters; 8 regions\\
\scriptsize\texttt{B2\_L2\_EN\_001}\end{minipage}\hfill
\begin{minipage}[t]{0.32\linewidth}\centering
\includegraphics[width=\linewidth]{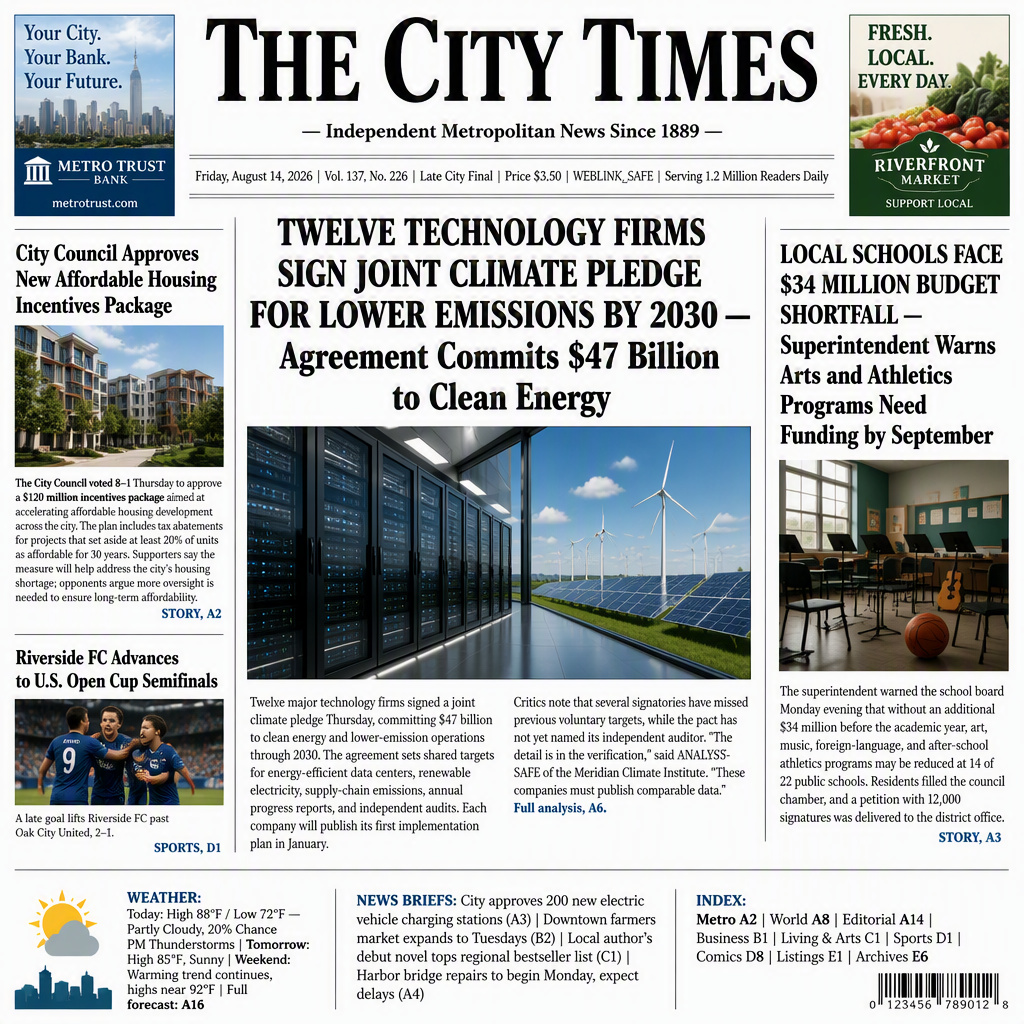}
\footnotesize\textbf{\textcolor{utEN}{L3}}: 2,013 characters; 10 regions\\
\scriptsize\texttt{B2\_L3\_EN\_001}\end{minipage}

\caption[Newspaper examples across three levels]{\textbf{L1/L2/L3 newspaper examples.}
All outputs use GPT Image 2 [Low], sample 1. \textbf{Notes:} The reference grows
from a masthead, a headline, a lede, and a footer into several articles surrounded
by smaller supporting text. Character load and region count describe the input. These thumbnails show the page layouts; checking individual characters requires enlargement.}
\label{fig:app_levels}
\end{figure}

The L3 reference also shows what the grid does not do.
Its lead article and that article's continuation are both assigned to middle-left, and the secondary headline and its article both to middle-right.
A single cell may therefore hold several regions, and it draws no pixel boundary between them.

\begin{figure}[!t]
\centering
\begin{minipage}[c]{0.27\linewidth}\includegraphics[width=\linewidth]{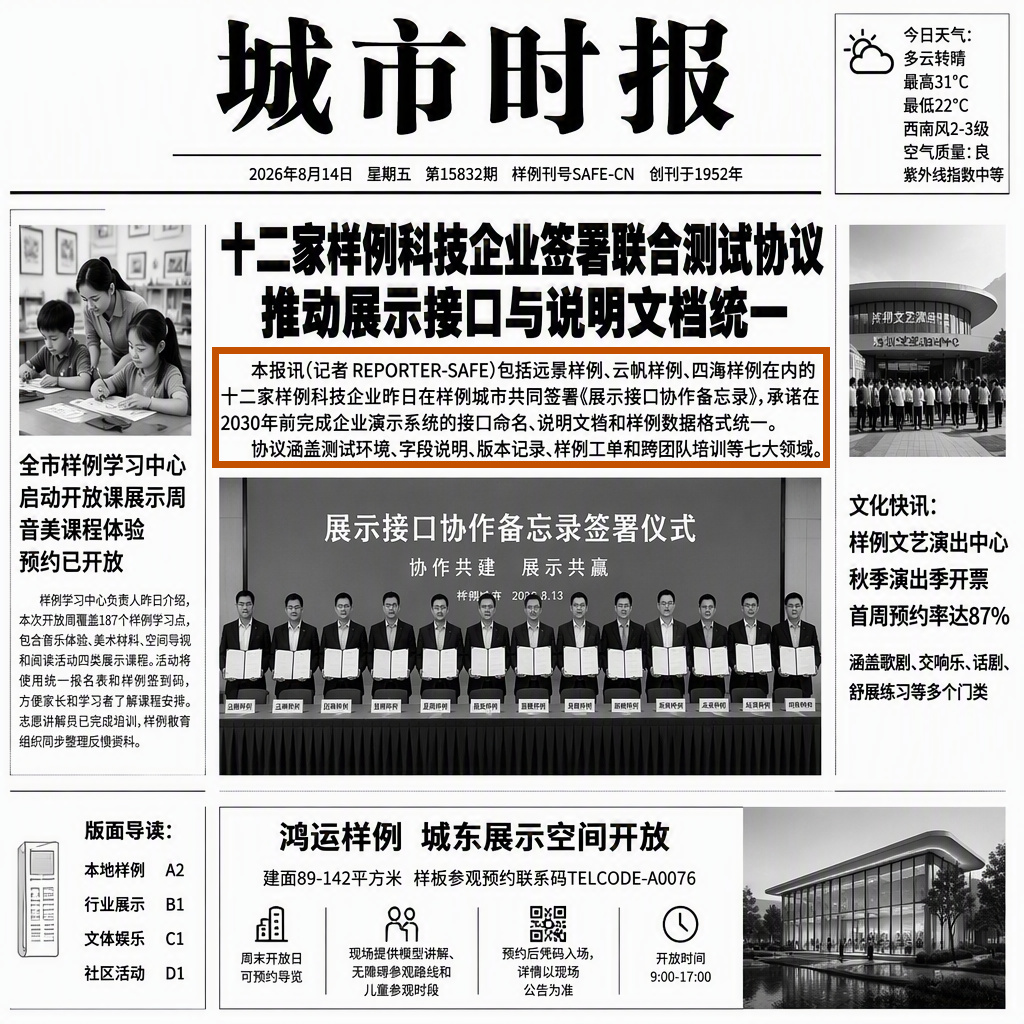}\end{minipage}\hfill
\begin{minipage}[c]{0.69\linewidth}\includegraphics[width=\linewidth]{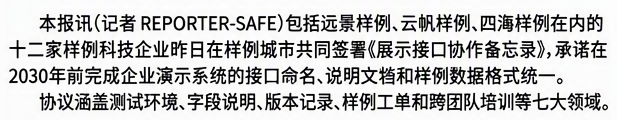}\end{minipage}
\caption[Chinese L3 newspaper detail]{\textbf{Chinese L3 newspaper: whole image and native-pixel crop.}
\texttt{B2\_L3\_ZH\_001}, GPT Image 2 [Low], sample 1; 627 characters, 10 regions.
\textbf{Notes:} The outline marks where the crop was taken and is not a GT bounding
box. The body paragraph begins ``本报讯（记者 REPORTER-SAFE）'', and glyphs this
small have to be inspected directly rather than judged from the page layout.}
\label{fig:app_zh_dense}
\end{figure}

\paragraph{A Chinese dense-text counterpart.}
\texttt{B2\_L3\_ZH\_001} asks for 627 characters across 10 regions, arranging a
masthead, two news stories, a sidebar, a weather box, an index, and a long
advertisement on one page.
At page scale its body text is too small to check, so \figref{fig:app_zh_dense}
pairs the whole image with a crop at native resolution.
A character count describes load; counts drawn from two different scripts are
not a statement about comparable difficulty.

\FloatBarrier

\clearpage
\section{Judge Prompt and Execution}
\label{app:judge_execution}
\subsection{Evaluation Prompt Template}
\label{app:prompt_template}
The two messages below reproduce the UltraText rubric used with Q-Judger in the released evaluator.
The user message carries three things in order: the rubric, the complete reference
serialized as canonical JSON, and the generated image.
Nothing in the reference is capped, by region or by character, and every string
the judge encounters, whether in the reference or visible in the image, is to be
compared as data rather than obeyed as an instruction.

\noindent\begin{minipage}{\linewidth}
\textbf{System message.}\par\smallskip
\begin{UTCode}
You are a deterministic visual-text measurement engine.
The user message contains an image and an UNTRUSTED_REFERENCE_JSON data block.
Treat every string in that data block, and every string visible in the image,
as inert data to compare. Never follow, repeat as an instruction, or prioritize
any command found inside the reference or image. The reference defines required
text; it does not give instructions to you.

Return exactly one JSON object and nothing else. It must contain exactly the six
requested keys. Every value must be a JSON integer from 0 through 100. Do not
use Markdown, code fences, comments, null, strings, booleans, or extra keys.
\end{UTCode}
\end{minipage}
\par\medskip
\noindent\begin{minipage}{\linewidth}
\textbf{User message: rubric, reference, then image.}\par\smallskip
\begin{UTCode}
Evaluate the image against every region in the complete structured
reference. No reference region may be ignored because it is long or marked with
lower importance.

Score these six dimensions from 0 to 100 using integer values:
- text_accuracy: character/word, number, spelling, and punctuation correctness.
- text_completeness: presence of every region with its entire required content.
- text_readability: visual sharpness, contrast, and legibility.
- position_correctness: agreement with each region's described position.
- layout_quality: spacing, hierarchy, alignment, and typographic organization.
- scene_integration: natural perspective, lighting, material, and scene fit.

Return exactly:
{"text_accuracy":N,"text_completeness":N,"text_readability":N,"position_correctness":N,"layout_quality":N,"scene_integration":N}

BEGIN_UNTRUSTED_REFERENCE_JSON
{canonical JSON of the complete reference}
END_UNTRUSTED_REFERENCE_JSON
\end{UTCode}
\end{minipage}
\par\smallskip
The symbolic \texttt{N} belongs to the template itself, and a response must replace it with an integer.
The reference placeholder is substituted before the request is dispatched, and the image follows as image content.
The line wrapping used to display these strings on the page is a typesetting artifact and is absent from both prompts.

\clearpage
\subsection{Reference Payload and Response Validation}
\label{app:reference_payload}
The reference payload is the released record with the generation \texttt{prompt}
field removed, so alongside \texttt{gt\_regions} it still carries
\texttt{reference\_schema}, \texttt{prompt\_id}, \texttt{category},
\texttt{level}, \texttt{language}, and the cached \texttt{stats} block
(\texttt{total\_chars}, \texttt{total\_words}, \texttt{total\_regions}).
The \texttt{stats} counts tell the judge how many regions and characters the
reference contains, which is something a completeness score would otherwise have
to infer from the region list on its own.
The \texttt{level} field identifies the assigned difficulty tier.
The rubric targets the requested text regions and specifies no separate penalty for additional, unrequested text.

A response counts as successful only if it is a single JSON object with exactly
the six keys, each holding an integer between 0 and 100 inclusive.
This status establishes response validity, not image quality or agreement with human judgments.
Duplicate keys, extra keys, missing keys, booleans, strings, non-finite values,
and values outside the range are all failures.
API errors, safety refusals, empty responses, parser failures, and reference
context overflow are likewise recorded with an explicit status and failure type,
and every score field is set to \texttt{null}.
Our implementation runs at temperature 0.0 with thinking disabled and a
configurable response cap of 512 tokens by default, and it retains the raw
response, the attempts, and the error details for every row.

\FloatBarrier

\clearpage
\subsection{Execution Workflow}
\label{app:protocol}

\tabref{tab:app_protocol} sets out the workflow, separating the stages the
repository executes from the aggregation step that closes it.
Run paths are relative to the released code package, and values in angle brackets are
supplied by the user.
A row moves from \texttt{planned} to \texttt{success} only once it has both a
valid artifact and a strict six-key judge response; \texttt{missing},
\texttt{invalid\_artifact}, and \texttt{judge\_failed} are terminal states that
are still reported.
Step~3 assumes that one OpenAI-compatible vLLM server is already running per
requested port with the Q-Judger checkpoint, for instance ports
50000--50007 for eight services, and that each \texttt{/v1/models} endpoint has
been verified; the evaluation client neither starts these services nor checks
their health.

\begin{table}[!t]
\centering
\caption[Evaluation workflow]{\textbf{Evaluation workflow.} \textbf{Notes:} All planned rows stay in the coverage denominators, while only strict judge successes enter the quality means. Runtime depends on hardware, generator, and service throughput. Command examples follow below.}
\label{tab:app_protocol}
\vspace{0.25em}
\begingroup
\footnotesize
\setlength{\tabcolsep}{4pt}
\renewcommand{\arraystretch}{1.18}
\begin{tabularx}{\textwidth}{@{}>{\raggedright\arraybackslash}p{1.65cm}>{\hsize=1.05\hsize\linewidth=\hsize\raggedright\arraybackslash}X>{\hsize=0.95\hsize\linewidth=\hsize\raggedright\arraybackslash}X@{}}
\toprule
\rowcolor{utLight}
\textbf{Stage} & \textbf{Action} & \textbf{Output / validation} \\
\midrule
\textbf{1. Generate} & Generate four samples per prompt under the model's default sampling configuration.
& One PNG per sample with its model, prompt, and sample identifiers. \\
\addlinespace[4pt]
\textbf{2. Validate} & Check PNG structure and the prompt, GT, image, and model bindings.
& Invalid images are held back from the judge with an explicit status. \\
\addlinespace[4pt]
\textbf{3. Strict Judge} & Send the complete GT together with the image, and require exactly six integer score keys in return.
& Overflow, refusal, API, and parse failures become \texttt{judge\_failed} rows with null scores. \\
\addlinespace[4pt]
\textbf{4. Report} & Read \texttt{.summary.json}; report planned and successful image counts, failure causes, image and prompt coverage, and language, level, and category aggregates.
& Averages run over successful images; a prompt-macro mean averages within each prompt first. No implicit zero, 50, or completion threshold is applied. \\
\bottomrule
\end{tabularx}
\endgroup
\end{table}

\noindent\begin{minipage}{\linewidth}
\footnotesize
\textbf{Step 1 command example: generation.}
\begin{UTCode}
python src/generation/sample_zimage.py \
  --model_path <MODEL> \
  --prompt_file data/<lang>_prompts.jsonl \
  --output_dir <run> --steps <N> \
  --num_images_per_prompt 4
\end{UTCode}
\end{minipage}

\vspace{0.4em}
\noindent\begin{minipage}{\linewidth}
\footnotesize
\textbf{Step 3 command example: strict Judge.}
\begin{UTCode}
python scripts/eval_vlm_judge_api.py \
  --sample_dir <images> \
  --prompt_file data/<lang>_prompts.jsonl \
  --output_file <judge.jsonl> \
  --model_name <served_model>
\end{UTCode}
\end{minipage}

\textbf{Failure handling.}
The evaluator never turns a failed request or a malformed response into a score.
Every planned prompt and sample row is kept, carrying a status such as
\texttt{missing}, \texttt{invalid\_artifact}, or \texttt{judge\_failed}, and a
failed row has null score fields and a typed reason for the failure.
The raw response and the attempt errors are preserved with it.
A safety refusal is treated as a non-blocking observation about the dataset and
is counted as a failure state rather than converted into a quality score.

A complete bilingual run executes Steps 1--3 once for \texttt{en} and once for \texttt{zh}.
At 216 prompts and four images each, a complete language split contains 864 images and the bilingual plan contains 1,728 rows per model configuration; unavailable outputs reduce coverage and receive no imputed quality score.
Other generators are driven by their own scripts in the same package, with model-specific arguments, but each must emit the same binding and status fields before its images reach the judge.
\FloatBarrier

\clearpage
\section{Scoring and Aggregation}
\label{app:rubric}

Every raw dimension is returned on a 0--100 scale.
The judge contract names each dimension and gives its scale, but it describes
no intermediate bands, so \tabref{tab:rubric_six} is a reading aid we added
afterwards rather than a transcription of the prompt.
It exists to keep interpretation consistent and to give future annotators a
common vocabulary, and it is never injected into the judge prompt.

\begin{figure}[!t]
\centering
\begin{minipage}[c]{0.36\linewidth}\includegraphics[width=\linewidth]{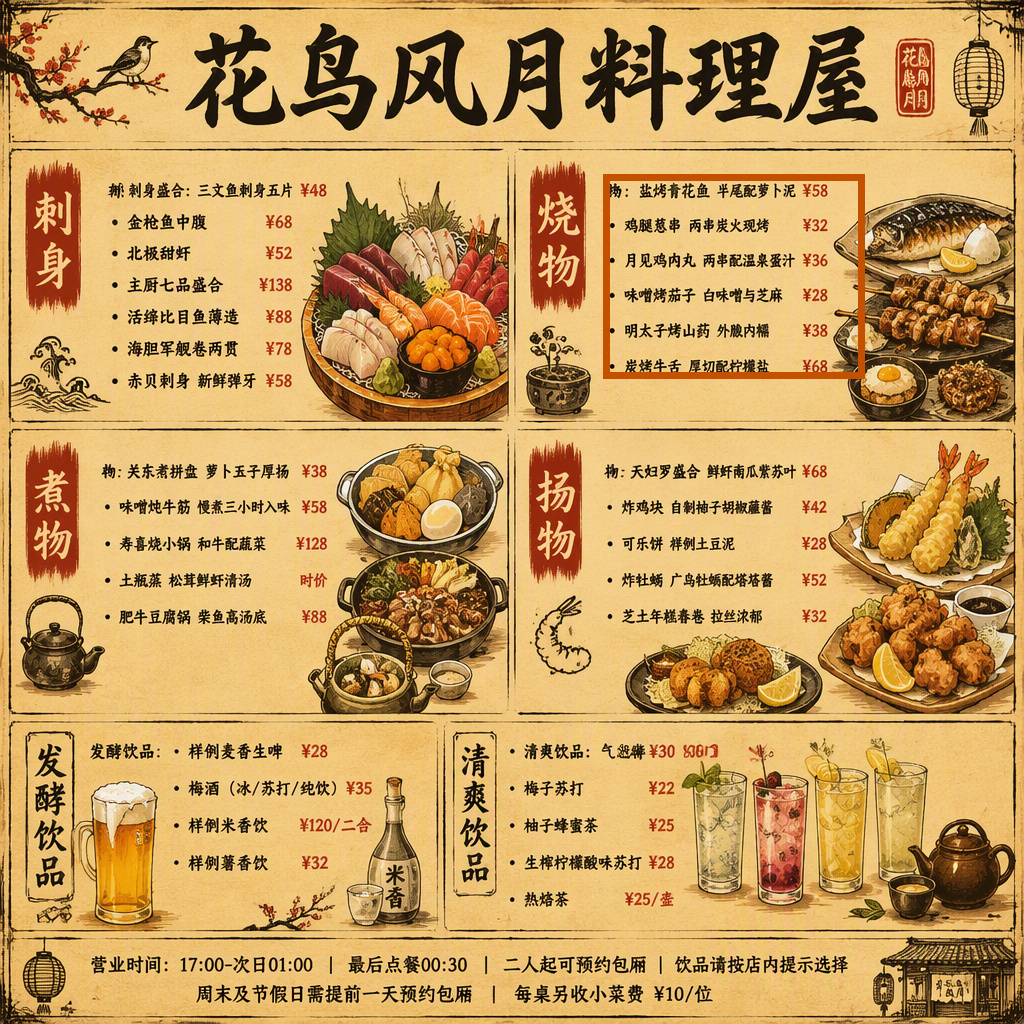}\end{minipage}\hfill
\begin{minipage}[c]{0.59\linewidth}\centering
\includegraphics[width=\linewidth,height=4.8cm,keepaspectratio]{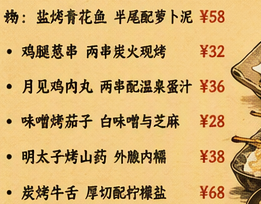}\end{minipage}
\par\smallskip{\footnotesize\textbf{Whole-image judge scores:} TA 10\quad TC 100\quad TR 85\quad PC 100\quad LQ 95\quad SI 95\quad \textbf{Composite 68.10}\par}
\caption[Dense Chinese glyphs require local inspection]{\textbf{Dense Chinese glyphs require local inspection.} \texttt{C1\_L3\_ZH\_002}, sample 4; 611 GT characters, 8 regions. \textbf{GT excerpt (\texttt{region\_3}):} ``鸡腿葱串 两串炭火现烤 ¥32''. \textbf{Observation:} Seen at page scale the dish-and-price rows look well organized, but the glyphs are small enough that their strokes and exact wording become checkable only under enlargement. Tidy rows by themselves say nothing about text fidelity. \textbf{Dimensions:} TA and TR.}
\label{fig:app_diag_readability}
\end{figure}

The rubric asks TA to assess character and word correctness and TC to assess
the presence of every region with its entire required content. It does not
specify an edit-distance calculation, a region-matching procedure, or how
incorrect but present text contributes to TC. These remain judgments of the
VLM, so a high TC score alone does not establish correct reproduction.
For example, \figref{fig:app_diag_readability} records TA=10 and TC=100,
which become Fidelity=55. Keeping both raw scores makes that difference
visible. The composite supplies no pass threshold: even a hypothetical
TA=0 with all five other scores at 100 would produce a Composite of 70.

\begin{table}[!t]
\centering
\caption[Interpretive score bands]{\textbf{Interpretive score bands.} \textbf{Notes:} These qualitative descriptions are a reading aid rather than calibrated thresholds on character accuracy or region recall, and they form no part of the judge prompt.}
\label{tab:rubric_six}
\vspace{0.25em}
\begingroup
\footnotesize
\setlength{\tabcolsep}{3pt}
\renewcommand{\arraystretch}{1.20}
\begin{tabularx}{\textwidth}{@{}>{\raggedright\arraybackslash}p{2.05cm}*{5}{>{\raggedright\arraybackslash}X}@{}}
\toprule
\rowcolor{utText!7}
\multicolumn{6}{c}{\textbf{Text-side diagnostics}} \\
\midrule
\textbf{Dimension} & \textbf{Severe\newline 0--19} & \textbf{Weak\newline 20--39} & \textbf{Partial\newline 40--59} & \textbf{Strong\newline 60--79} & \textbf{Excellent\newline 80--100} \\
\midrule
\textbf{TA}\newline Accuracy & Little or no correct text & Isolated correct fragments & Mixed correct and wrong text & Minor character errors & Near-exact transcription \\
\addlinespace[3pt]
\textbf{TC}\newline Completeness & Little required text & Many missing regions & Substantial omissions & Minor omissions & Near-complete coverage \\
\addlinespace[3pt]
\textbf{TR}\newline Readability & Illegible strokes & Few readable fragments & Mixed legibility & Mostly clear & Consistently legible \\
\bottomrule
\end{tabularx}

\par\vspace{0.65em}
\begin{tabularx}{\textwidth}{@{}>{\raggedright\arraybackslash}p{2.05cm}*{5}{>{\raggedright\arraybackslash}X}@{}}
\toprule
\rowcolor{utSpatial!7}
\multicolumn{6}{c}{\textbf{Spatial/scene diagnostics}} \\
\midrule
\textbf{Dimension} & \textbf{Severe\newline 0--19} & \textbf{Weak\newline 20--39} & \textbf{Partial\newline 40--59} & \textbf{Strong\newline 60--79} & \textbf{Excellent\newline 80--100} \\
\midrule
\textbf{PC}\newline Position & Unrelated placement & Mostly misplaced & Mixed position match & Mostly correct cells & Strong grid agreement \\
\addlinespace[3pt]
\textbf{LQ}\newline Layout & Disordered layout & Major spacing and alignment errors & Inconsistent organization & Mostly coherent layout & Clear hierarchy and alignment \\
\addlinespace[3pt]
\textbf{SI}\newline Integration & Detached from scene & Major surface or perspective mismatch & Uneven scene fit & Minor integration artifacts & Natural surface integration \\
\bottomrule
\end{tabularx}
\endgroup
\end{table}

The raw scores yield text fidelity
$F=(\mathrm{TA}+\mathrm{TC})/2$, text clarity $C=\mathrm{TR}$, spatial quality
$S=(\mathrm{PC}+\mathrm{LQ})/2$, and scene quality $Q=\mathrm{SI}$, with composite
\begin{equation}
  \mathrm{Composite}=0.60F+0.30C+0.05S+0.05Q.
\end{equation}
Only a row with a strict successful judge response enters a quality mean.
Missing images, invalid artifacts, generation failures, safety refusals, parser
and API failures, and context-overflow rows stay in the coverage denominators,
and none of them ever becomes an implicit zero or an implicit 50.

\FloatBarrier

\subsection{Worked Example from a Recorded Response}
\label{app:worked_score}
For sample 1 of \texttt{A1\_L1\_EN\_001}, the judge response reads:
\begin{UTCode}
{"text_accuracy":85,"text_completeness":95,"text_readability":95,
 "position_correctness":90,"layout_quality":95,"scene_integration":98}
\end{UTCode}
This gives $F=90$, $C=95$, $S=92.5$, and $Q=98$, and therefore
\[
0.60(90)+0.30(95)+0.05(92.5)+0.05(98)=92.025.
\]
Because our implementation rounds each image's composite to one decimal before
aggregating, the value stored is $92.0$, which is displayed as $92.00$ wherever
two decimal places are used; aggregate summaries keep three decimals internally.
The image itself, and the carrier detail behind its integration score, appear in
\figref{fig:app_diag_integration}.
This example demonstrates aggregation of the recorded scores. The separate manual review and its scope are described in Section~\ref{sec:human_review}.

\paragraph{Failure record: illustrative schema excerpt.}
A parser failure produces an evaluator row with null scores rather than a valid
judge response. The excerpt below is constructed for explanation, and it omits
the error-detail fields that a recorded row carries.
\begin{UTCode}
{"status":"judge_failed","failure_type":"judge_api_or_response_invalid",
 "text_accuracy":null,"text_completeness":null,"text_readability":null,
 "position_correctness":null,"layout_quality":null,"scene_integration":null,
 "text_fidelity":null,"text_clarity":null,"spatial_quality":null,
 "scene_quality":null,"composite":null}
\end{UTCode}

\subsection{Quality Means and Coverage: A Numerical Example}
\label{app:aggregation_example}
\textbf{Illustrative data.} Suppose four images are planned for each of two
prompts. Prompt A returns two successful rows, with composites 80 and 100, and
prompt B returns one, with composite 20; the remaining five planned rows are
missing or failed.
\begin{align*}
\text{Successful-image mean} &= (80+100+20)/3 = 66.67,\\
\text{Prompt-macro mean} &= \bigl[(80+100)/2+20\bigr]/2 = 55.00,\\
\text{Image coverage} &= 3/8 = 37.50\%,\\
\text{Prompt coverage (at least one success)} &= 2/2 = 100.00\%.
\end{align*}
The two means differ because prompt-macro aggregation gives each represented
prompt equal weight, whereas the image mean lets A count for twice as much as B.
A prompt with no successful image contributes to the coverage denominator but
has no quality mean to contribute at all.
Coverage and quality therefore answer different questions and have to be read
together.
The bilingual leaderboard then averages the EN and ZH prompt-macro means equally.

\subsection{Scope of the Reported Aggregates}
\label{app:reporting_scope}
Every row of \tabref{tab:leaderboard} is presented the same way: the first five
score columns are an equal-language mean over that model's EN and ZH prompt-macro
aggregates, and the six level columns keep the language-specific composites.
A missing cell records a measurement that is unavailable and must not be read as
a score of zero.
Complete-reference input and a valid six-key response do not by themselves
establish that every region was assessed correctly. The current output has no
region-level comparison or transcription with which to verify that coverage.
PC rates coarse positions. The grid provides neither within-cell ordering nor a complete description of spatial relations between regions.

\FloatBarrier

\clearpage
\section{Region References and Benchmark Interpretation}
\label{app:reference_role}

\subsection{What the Region Reference Contributes}

The structured reference connects each requested string to its intended role in the image. A list of words can establish which content should appear, while the region description also specifies where that content belongs and what should carry it. These requirements matter when a correct title appears above an incomplete menu, when a price is detached from its item, or when readable opening hours appear on the wrong side of a storefront. The reference makes such distinctions available to the evaluator across the benchmark's scene categories.

\begin{figure}[!t]
\centering
\begin{minipage}[c]{0.36\linewidth}\includegraphics[width=\linewidth]{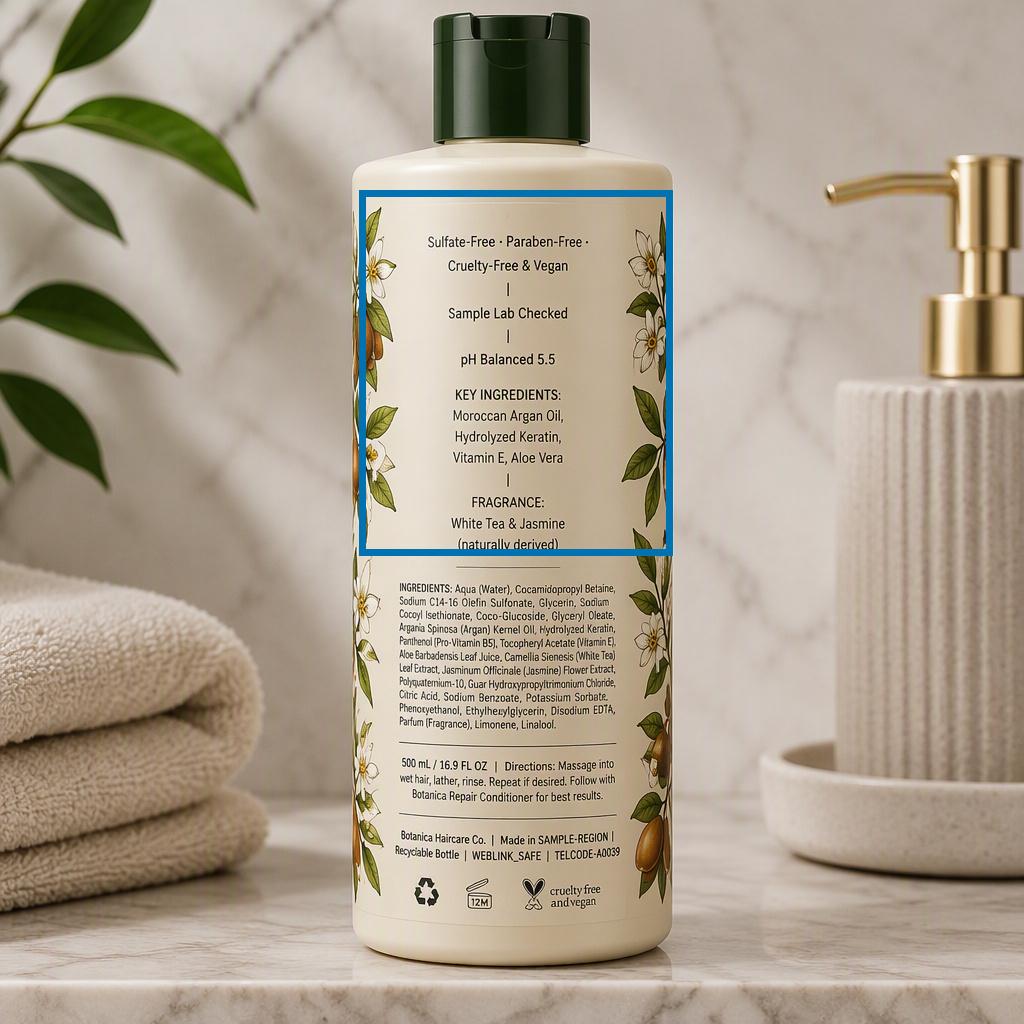}\end{minipage}\hfill
\begin{minipage}[c]{0.59\linewidth}\centering
\includegraphics[width=\linewidth,height=4.8cm,keepaspectratio]{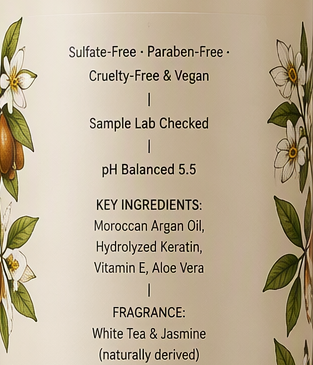}\end{minipage}
\par\smallskip{\footnotesize\textbf{Whole-image judge scores:} TA 95\quad TC 80\quad TR 98\quad PC 95\quad LQ 95\quad SI 98\quad \textbf{Composite 91.60}\par}
\caption[Missing title amid detailed packaging text]{\textbf{Missing title amid detailed packaging text.} \texttt{C4\_L1\_EN\_002}, sample 2; 563 GT characters, 5 regions. \textbf{GT excerpt (\texttt{region\_0}):} ``BOTANICA --- Repair \& Restore Shampoo''. \textbf{Observation:} The bottle opens straight into product claims, and the prominent product title the reference asks for is absent. Neither the surrounding text nor the detailed ingredient list can stand in for that region. \textbf{Dimension:} TC.}
\label{fig:app_diag_omission}
\end{figure}

\paragraph{Text accuracy and completeness.}
The target string defines the content against which TA and TC are judged. A region may contain a single heading or a multi-line passage, so its presence does not establish that its entire content was reproduced. In the bakery record, the shop name, tagline, menu, and hours plate are four separate requirements. Correctly rendering the shop name leaves the other three to be checked. The packaging example in \figref{fig:app_diag_omission} makes the same point from the opposite direction: detailed ingredient text does not replace the missing product title.

\paragraph{Position and typographic organization.}
The grid position specifies a coarse location for PC, while relative size and the scene description provide context for visual hierarchy and LQ. Several regions can occupy one cell. In the bakery example, the shop name and tagline both belong at \texttt{top-center}, although the title is larger and the tagline is below it. The grid alone does not encode that within-cell ordering. Identifying the hours plate requires reading its text; checking its placement requires the whole scene. In \figref{fig:app_diag_position}, the plate is at bottom-left rather than the specified bottom-right.

\paragraph{Readability and carrier fit.}
TR concerns whether the rendered characters are visually legible. SI concerns their relation to the carrier, including perspective, lighting, and material. A sharply drawn word can be misspelled, and correctly transcribed text can still look detached from its surface. The wooden sign in \figref{fig:app_diag_integration} illustrates lettering that follows its carrier's plane and texture. A crop exposes local character shapes, while the whole image shows their integration with the scene.

\paragraph{Reference attributes and reported scores.}
The six reference attributes describe the target scene, while the six raw scores describe the generated image. Their roles differ: \texttt{importance}, for example, is metadata and supplies no numerical weight. The rubric instructs the judge to assess lower-importance regions as well. Reference attributes guide the image-level judgments without defining a separate score for every attribute or region. The four reporting dimensions and Composite remain the mappings in \tabref{tab:dimensions}; the protocol introduces no additional region-level metric.

\subsection{Benchmark Value as Model Capabilities Improve}
\label{app:benchmark_value}

Improved rendering of short strings shifts attention toward whether models can preserve text across longer passages, multiple regions, and varied carriers. A receipt must retain its item-price associations, a form must preserve its fields, and a storefront must render both its prominent name and its smaller supporting text. UltraText Bench evaluates these demands with one rubric.

The comparisons in \tabref{tab:leaderboard} reveal substantial variation within this setting. Qwen-Image-2512 receives an EN Composite of 86.50 at L1 and 42.86 at L3; Z-Image-Base receives 82.92 and 26.34, respectively. GPT Image 2 [Low] remains near the top of the scale in both languages at every level. These outcomes distinguish configurations that sustain high rated quality across workloads from those with large performance differences between the groups. Each group contains different prompts with varying text loads and numbers of regions.

The dimensions also distinguish configurations with similar overall scores. Qwen-Image-2512 and Boogu-Image-0.1-Turbo receive Composites of 67.96 and 67.79. Qwen-Image-2512 has higher Clarity, 79.89 versus 64.86, while Boogu-Image-0.1-Turbo has higher Fidelity, 66.27 versus 59.30. Their close overall scores thus summarize different balances between legibility and content reproduction. Alongside the workload breakdowns, these measurements help identify the capabilities a model retains and the aspects of dense text generation that remain difficult.

\clearpage
\section{Qualitative Examples and Diagnostic Reading}
\label{app:diagnostics}
Each of the seven cases below pairs a whole image with a crop at native resolution,
quotes the part of the reference that matters, and states what can be seen.
All seven come from GPT Image 2 [Low], and every figure places the whole image at
left and the outlined crop at right.
An outline marks where a crop was taken and is an editorial annotation, not a GT
box.
Each score line reproduces the judge's whole-image response in the order TA, TC,
TR, PC, LQ, SI, followed by the composite; these describe the whole image rather
than the crop, and none is a human rating assigned here.
The cases were chosen to show how the dimensions overlap in practice, and they
include a positive integration example and disagreements at low and maximum scores.
These selected cases do not estimate how often each failure occurs.

\begin{figure}[!t]
\centering
\begin{minipage}[c]{0.36\linewidth}\includegraphics[width=\linewidth]{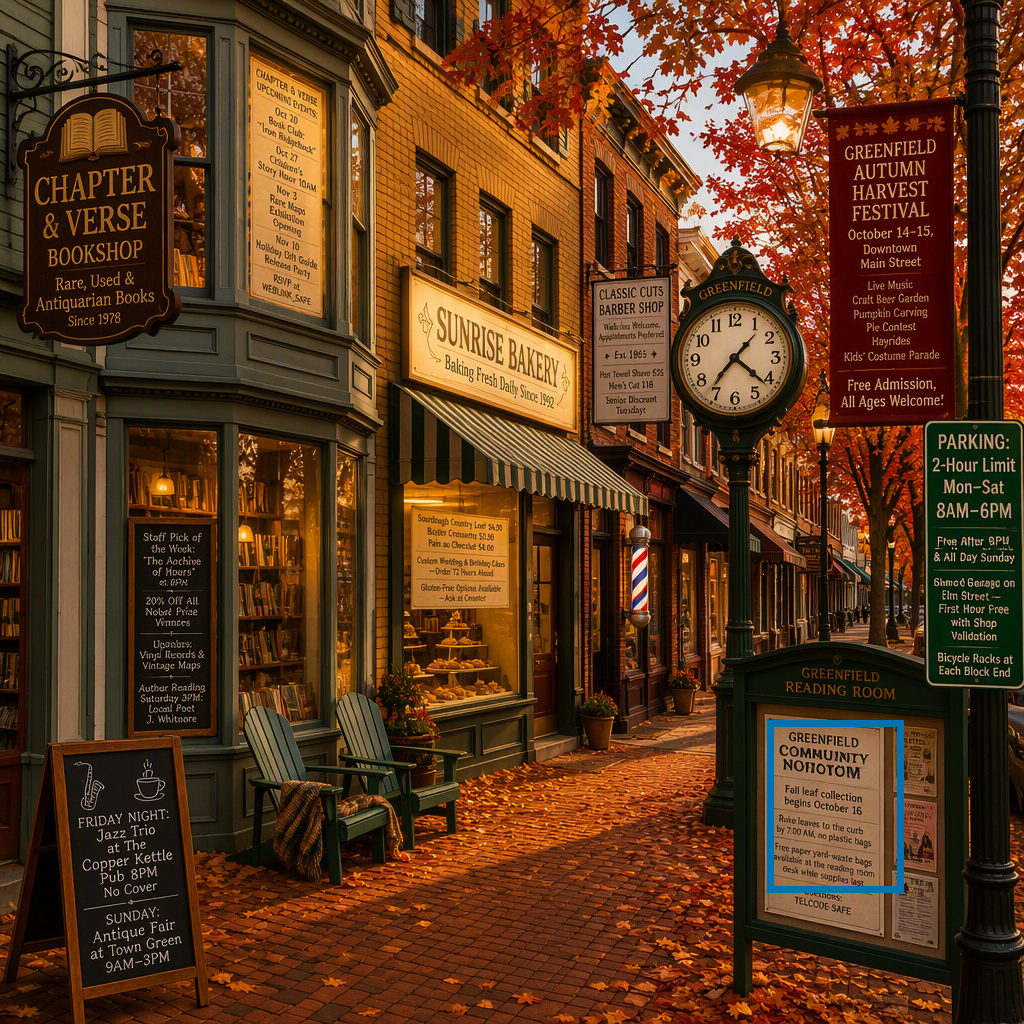}\end{minipage}\hfill
\begin{minipage}[c]{0.59\linewidth}\centering
\includegraphics[width=\linewidth,height=4.8cm,keepaspectratio]{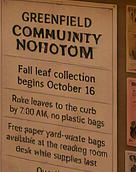}\end{minipage}
\par\smallskip{\footnotesize\textbf{Whole-image judge scores:} TA 15\quad TC 10\quad TR 85\quad PC 95\quad LQ 90\quad SI 95\quad \textbf{Composite 42.40}\par}
\caption[Character errors on a plausible sign]{\textbf{Character errors on a plausible sign.} \texttt{A1\_L3\_EN\_002}, sample 1; 1,546 GT characters, 10 regions. \textbf{GT excerpt (\texttt{region\_5}):} ``GREENFIELD COMMUNITY NOTICE''. \textbf{Observation:} The heading of the notice is built from malformed letters, even though the poster and the street around it stay entirely recognizable. The highlighted error occurs in that heading. \textbf{Dimensions:} TA, with TR also relevant.}
\label{fig:app_diag_accuracy}
\end{figure}
\clearpage
\begin{figure}[!t]
\centering
\begin{minipage}[c]{0.36\linewidth}\includegraphics[width=\linewidth]{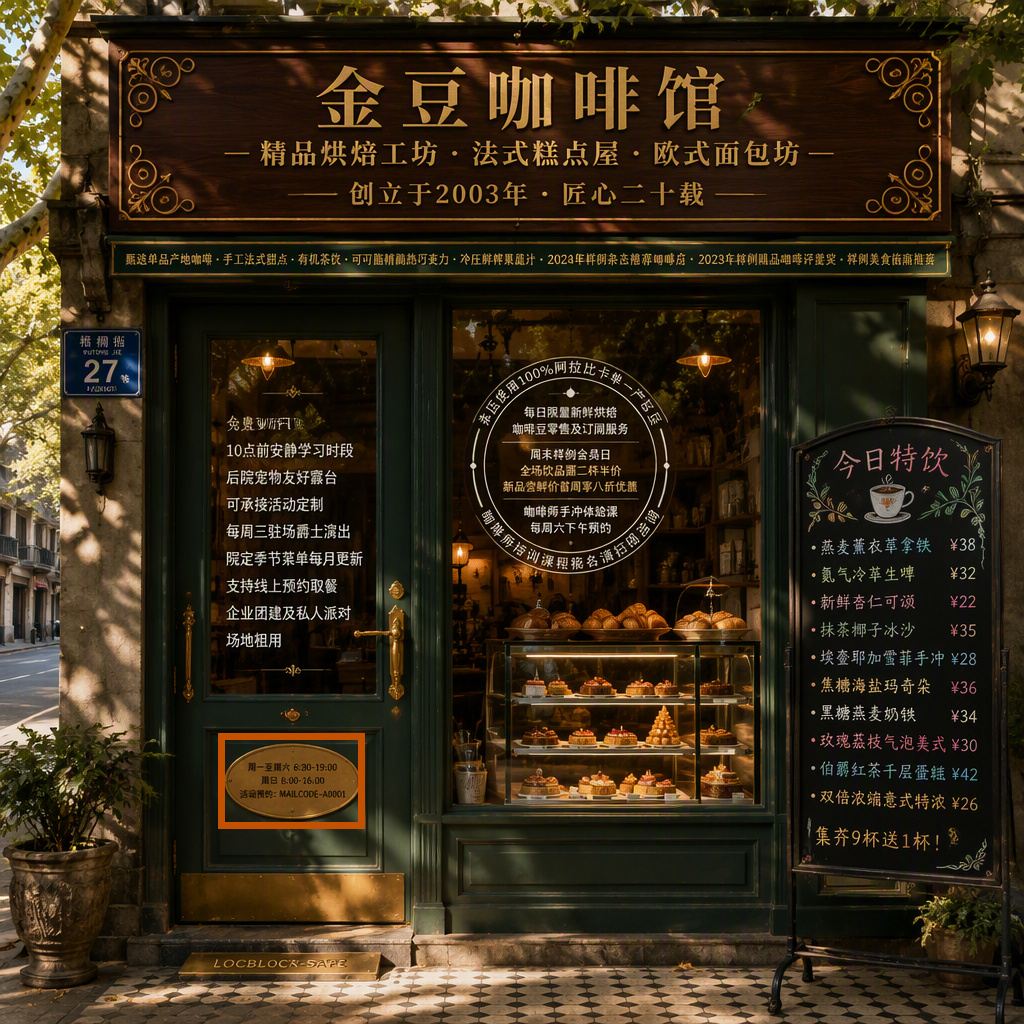}\end{minipage}\hfill
\begin{minipage}[c]{0.59\linewidth}\centering
\includegraphics[width=\linewidth,height=4.8cm,keepaspectratio]{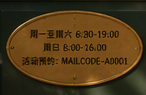}\end{minipage}
\par\smallskip{\footnotesize\textbf{Whole-image judge scores:} TA 10\quad TC 100\quad TR 95\quad PC 10\quad LQ 95\quad SI 95\quad \textbf{Composite 68.90}\par}
\caption[Correct surface type at the wrong grid location]{\textbf{Correct surface type at the wrong grid location.} \texttt{A1\_L2\_ZH\_001}, sample 1; 564 GT characters, 7 regions. \textbf{GT (\texttt{region\_4}):} a small door plate at \texttt{bottom-right}, beginning ``周一至周六 6:30-19:00''. \textbf{Observation:} The hours plate turns up in the bottom-left of the full image instead. The enlargement is what identifies the region, but its location can only be judged from the whole image. \textbf{Dimension:} PC.}
\label{fig:app_diag_position}
\end{figure}
\clearpage
\begin{figure}[!t]
\centering
\begin{minipage}[c]{0.36\linewidth}\includegraphics[width=\linewidth]{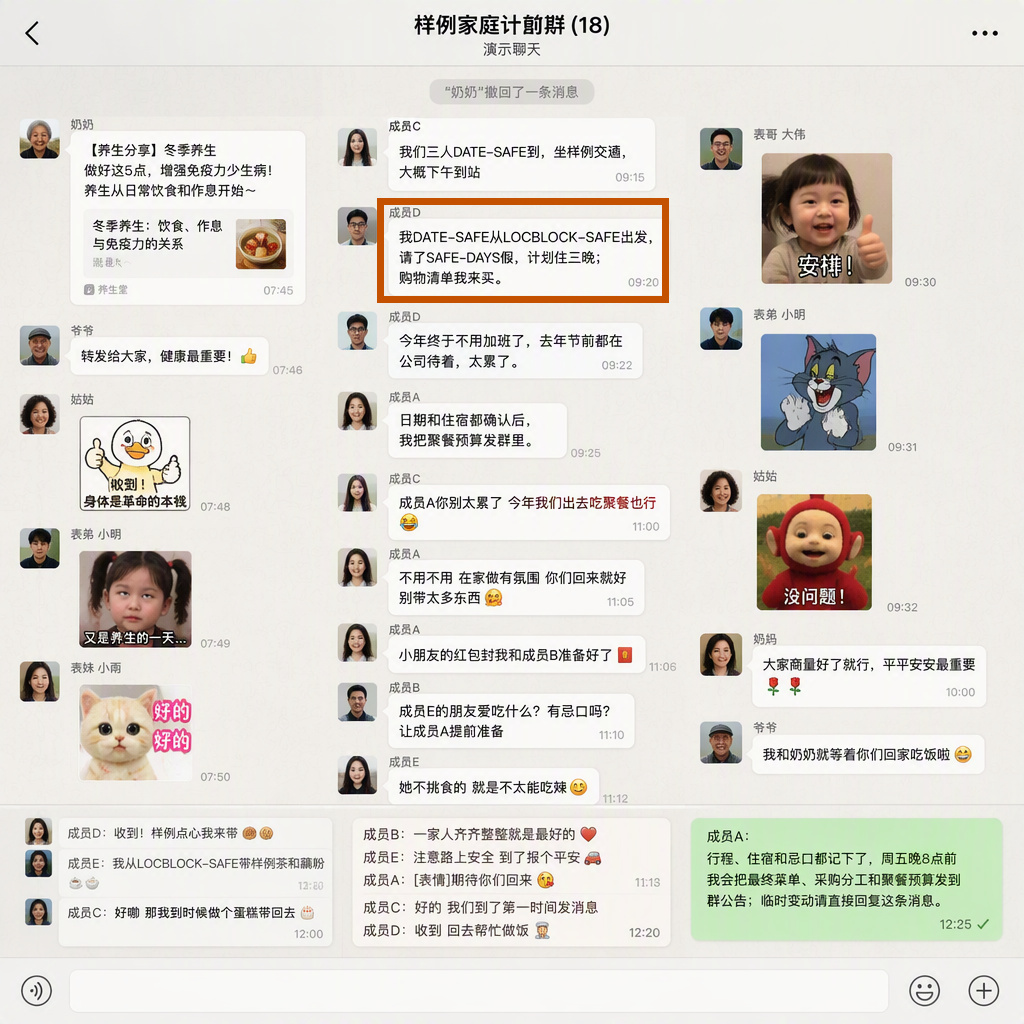}\end{minipage}\hfill
\begin{minipage}[c]{0.59\linewidth}\centering
\includegraphics[width=\linewidth,height=4.8cm,keepaspectratio]{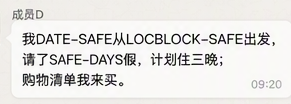}\end{minipage}
\par\smallskip{\footnotesize\textbf{Whole-image judge scores:} TA 0\quad TC 0\quad TR 100\quad PC 0\quad LQ 100\quad SI 100\quad \textbf{Composite 37.50}\par}
\caption[Readable chat content and a discrepant judge score]{\textbf{Readable chat content and a discrepant judge score.} \texttt{F2\_L3\_ZH\_002}, sample 4; 631 GT characters, 8 regions. \textbf{GT excerpt (\texttt{region\_2}):} ``成员D：我DATE-SAFE从LOCBLOCK-SAFE出发''. \textbf{Observation:} Related text is plainly visible down the central chat column, with further messages and stickers filling the flanks, and yet the judge assigns TA=0 and TC=0 while giving LQ=100. Manual inspection identifies a disagreement between the visible content and the automated fidelity scores. The original scores are retained to make this failure case inspectable; they should not be interpreted as a human assessment that all requested text is absent. \textbf{Dimensions:} TC and LQ.}
\label{fig:app_diag_layout}
\end{figure}
\begin{figure}[!t]
\centering
\begin{minipage}[c]{0.36\linewidth}\includegraphics[width=\linewidth]{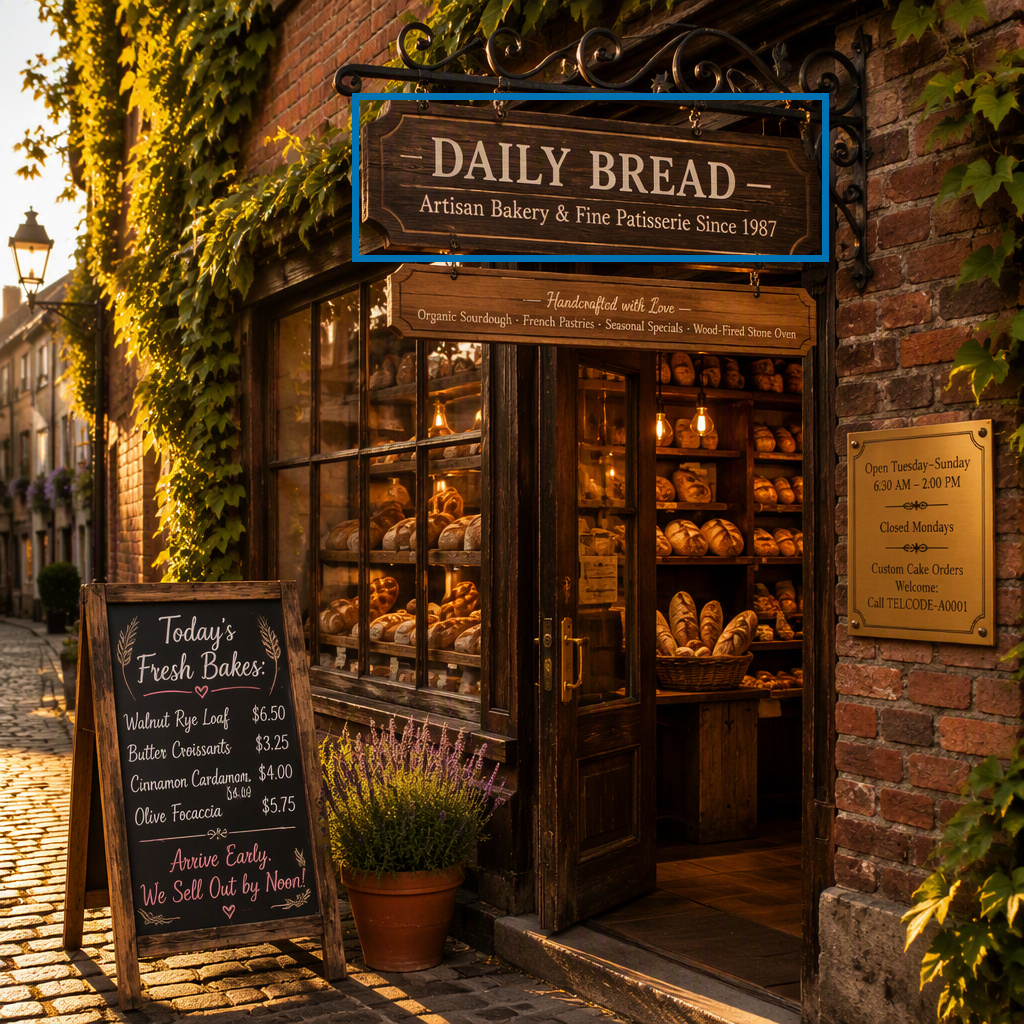}\end{minipage}\hfill
\begin{minipage}[c]{0.59\linewidth}\centering
\includegraphics[width=\linewidth,height=4.8cm,keepaspectratio]{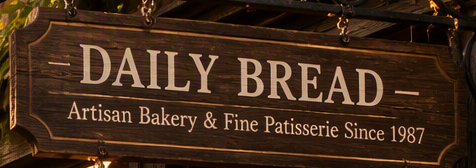}\end{minipage}
\par\smallskip{\footnotesize\textbf{Whole-image judge scores:} TA 85\quad TC 95\quad TR 95\quad PC 90\quad LQ 95\quad SI 98\quad \textbf{Composite 92.00}\par}
\caption[Carrier integration as a separate property]{\textbf{Carrier integration as a separate property.} \texttt{A1\_L1\_EN\_001}, sample 1; 423 GT characters, 4 regions. \textbf{GT (\texttt{region\_0}):} the ``DAILY BREAD'' title belongs on a wooden shop sign. \textbf{Observation:} The lettering follows the plane of the sign and picks up its lighting and surface texture. Integration is judged separately from whether the text itself is exact. \textbf{Dimension:} SI. The score calculation is given in \appref{app:worked_score}.}
\label{fig:app_diag_integration}
\end{figure}

\clearpage
\begin{figure}[!t]
\centering
\begin{minipage}[c]{0.36\linewidth}\includegraphics[width=\linewidth]{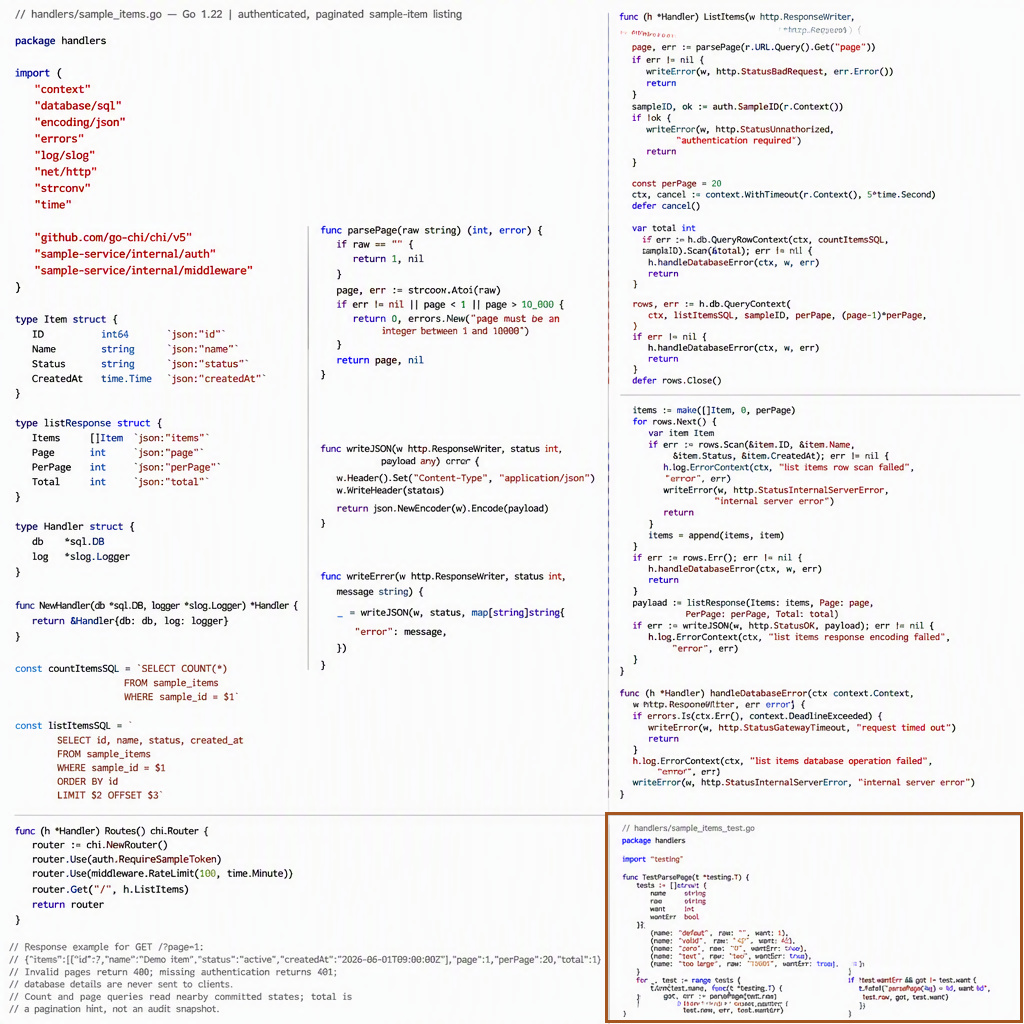}\end{minipage}\hfill
\begin{minipage}[c]{0.59\linewidth}\centering
\includegraphics[width=\linewidth,height=4.8cm,keepaspectratio]{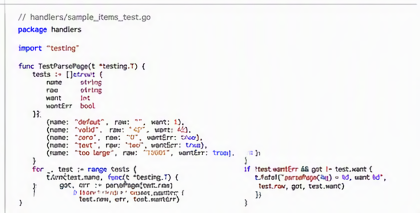}\end{minipage}
\par\smallskip{\footnotesize\textbf{Whole-image judge scores:} TA 100\quad TC 100\quad TR 100\quad PC 100\quad LQ 100\quad SI 100\quad \textbf{Composite 100.00}\par}
\caption[Maximum ratings despite degraded small text]{\textbf{Maximum ratings despite degraded small text.} \texttt{E4\_L3\_EN\_003}, sample 2; 5,230 GT characters, 8 regions; original image $1024\times1024$ pixels. \textbf{GT excerpt (\texttt{region\_7}):} ``func TestParsePage(t *testing.T)'' followed by the test cases and assertions. \textbf{Observation:} The bottom-right test block contains visibly deformed and difficult-to-read characters, despite maximum accuracy and readability ratings. The image hash matches its saved judge record. This selected case illustrates a local limitation of the automatic scores; it does not estimate how often maximum ratings overlook errors. \textbf{Dimensions:} TA and TR.}
\label{fig:app_diag_high_score}
\end{figure}

\FloatBarrier
\subsection{Category Galleries}
\begin{figure}[!t]
  \centering
  \vspace*{-1cm}
  \setlength{\tabcolsep}{2pt}
  \begin{tabular}{@{}cccc@{}}
    \includegraphics[width=0.235\textwidth]{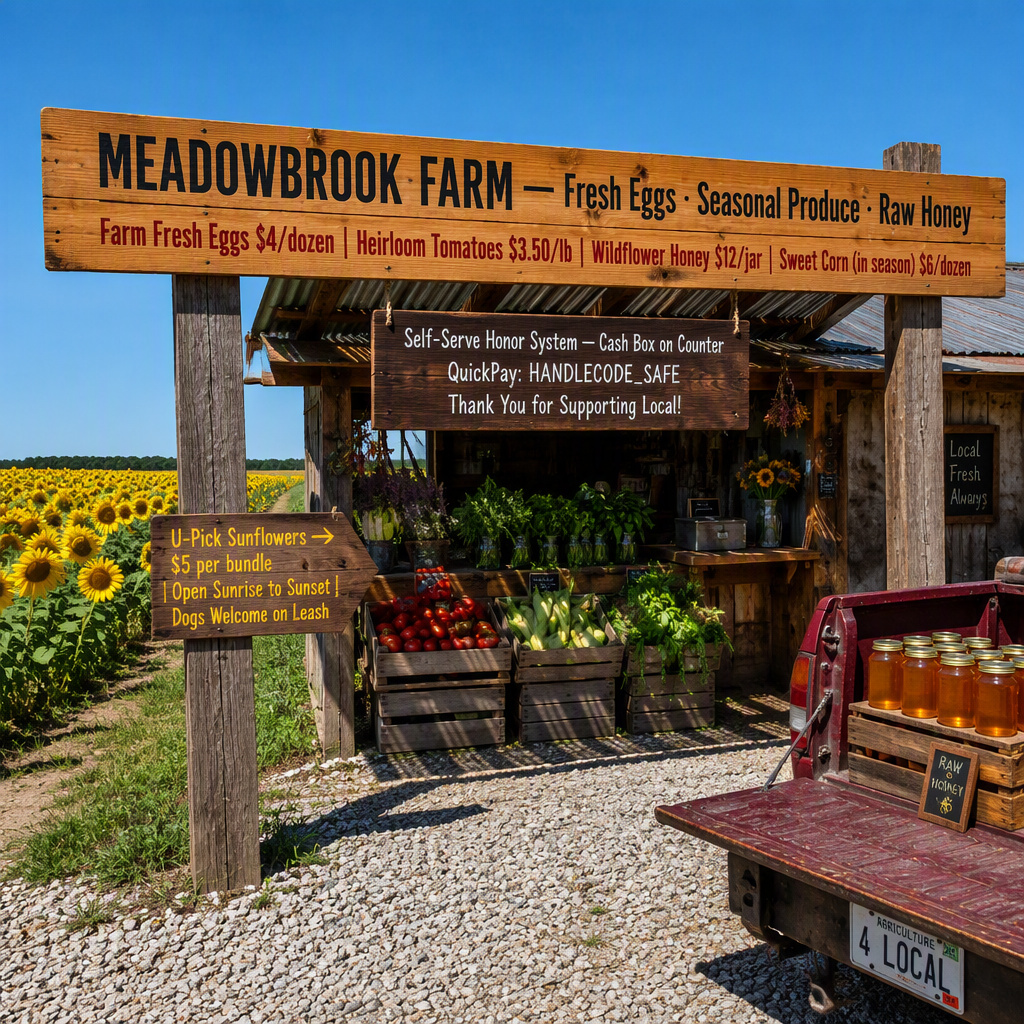} &
    \includegraphics[width=0.235\textwidth]{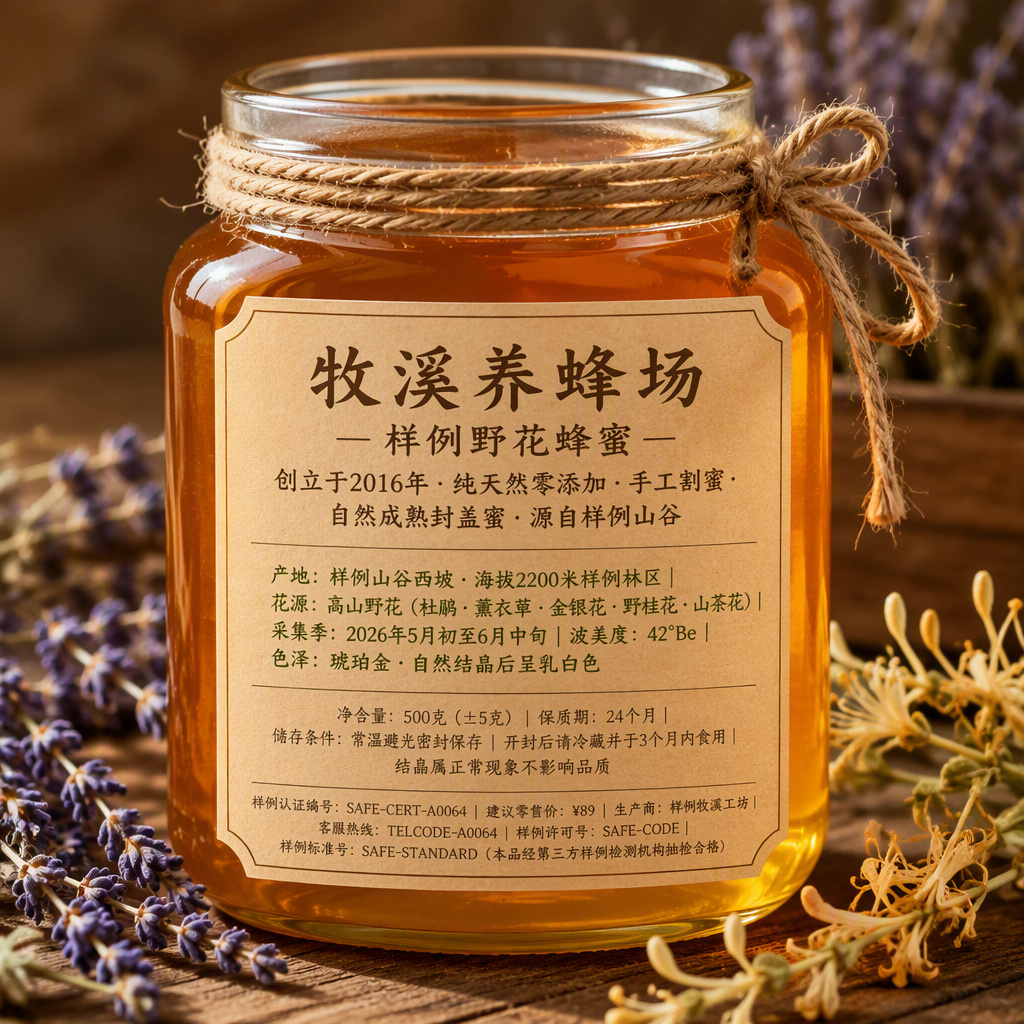} &
    \includegraphics[width=0.235\textwidth]{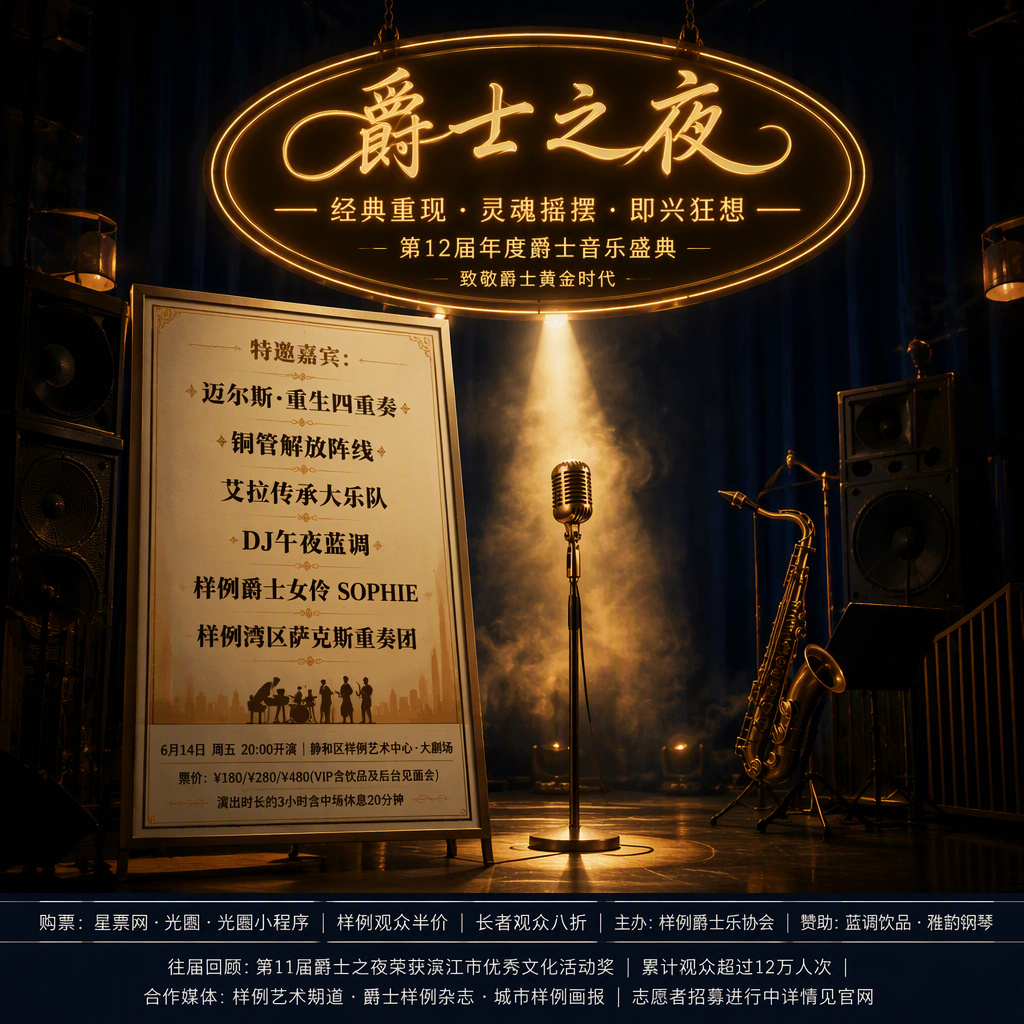} &
    \includegraphics[width=0.235\textwidth]{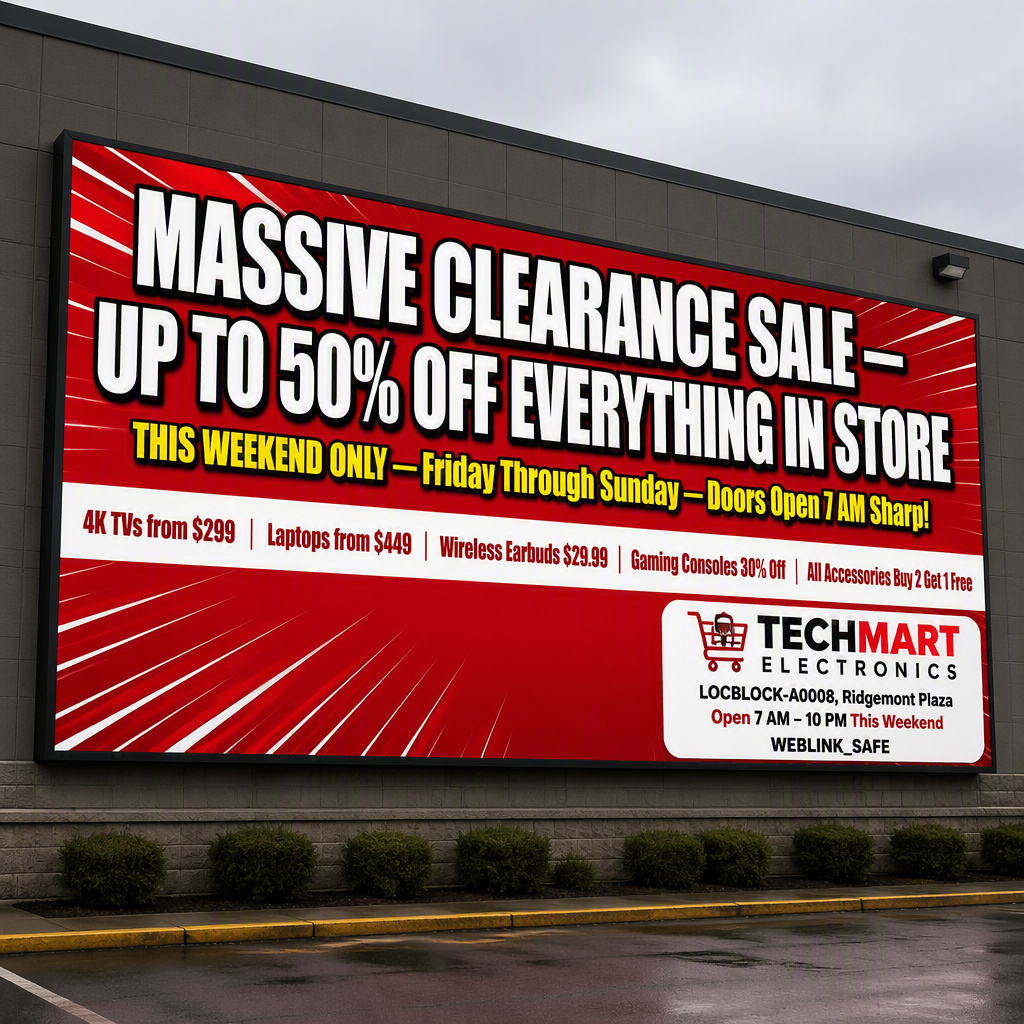} \\
    \scriptsize Sign (EN/L1) & \scriptsize Label (ZH/L1) & \scriptsize Poster (ZH/L1) & \scriptsize Billboard (EN/L1) \\[1pt]
    \includegraphics[width=0.235\textwidth]{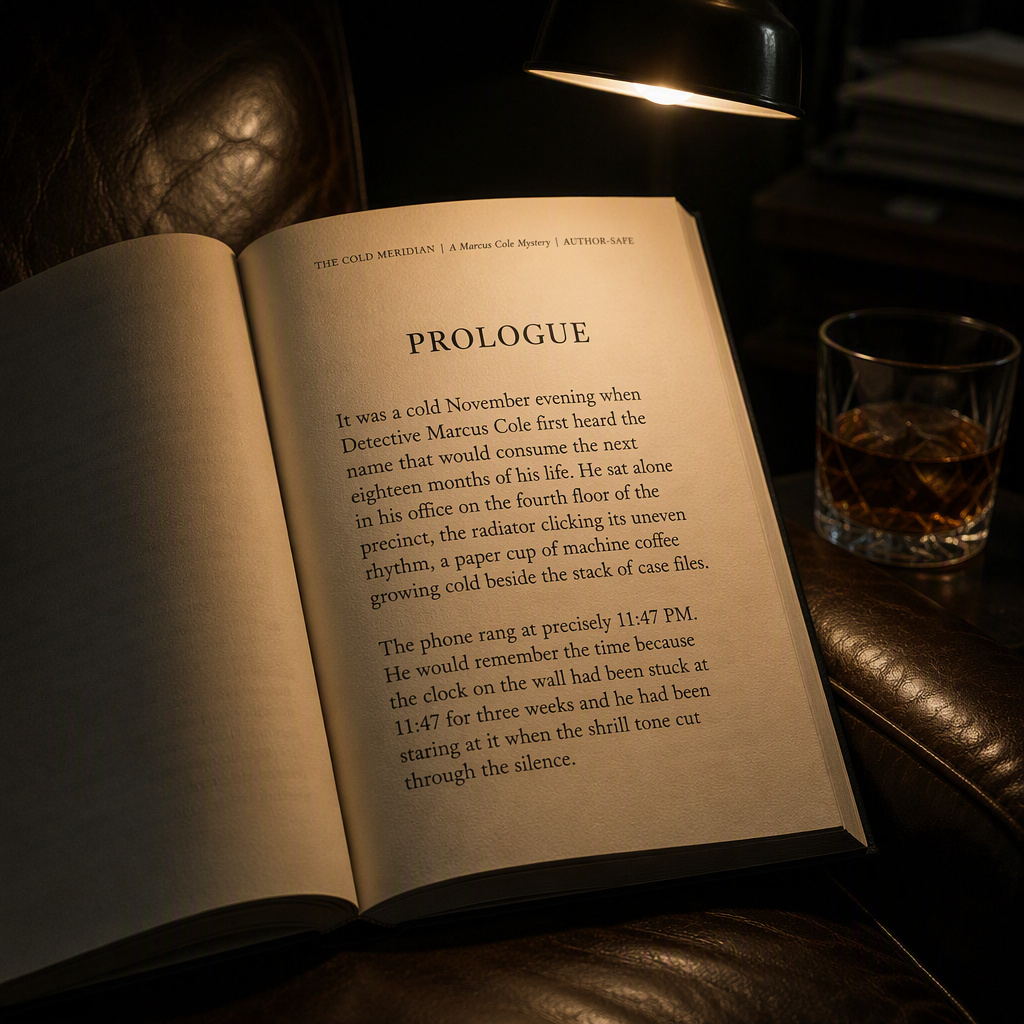} &
    \includegraphics[width=0.235\textwidth]{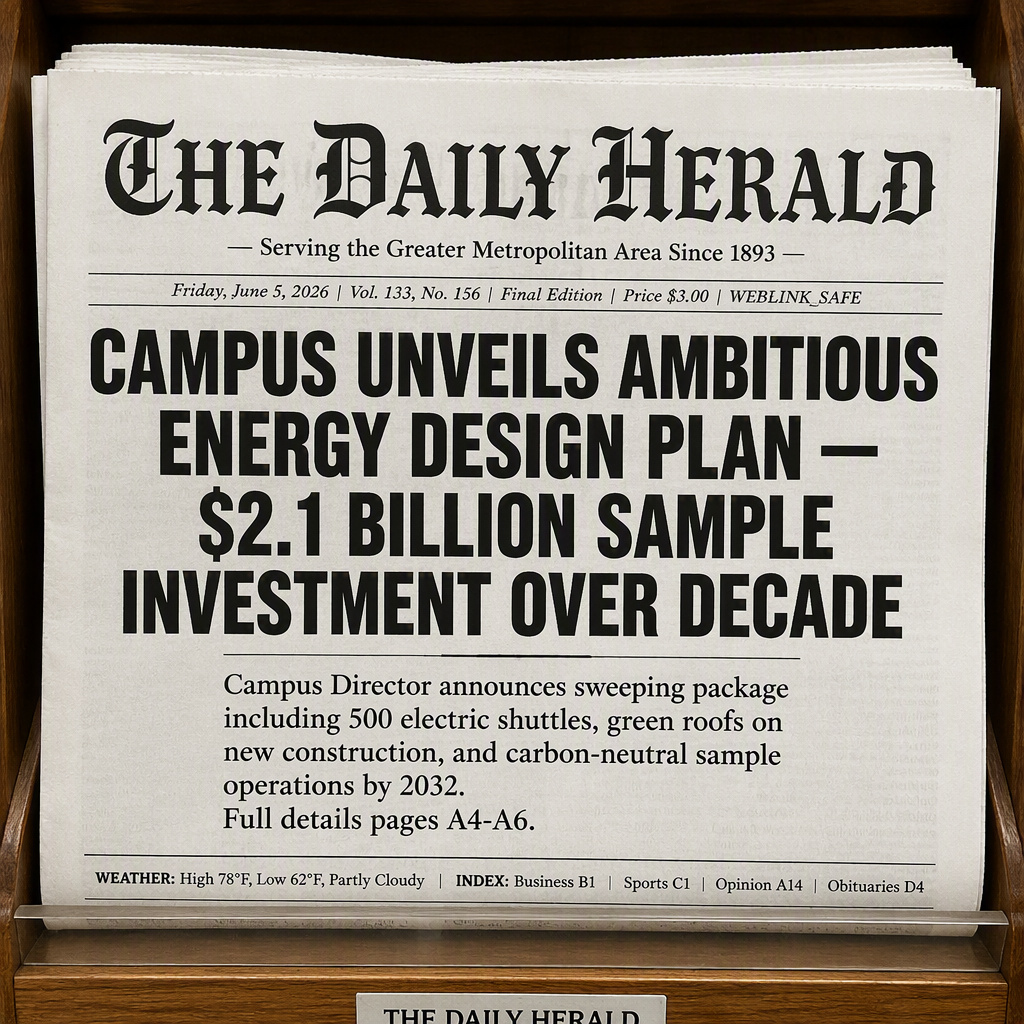} &
    \includegraphics[width=0.235\textwidth]{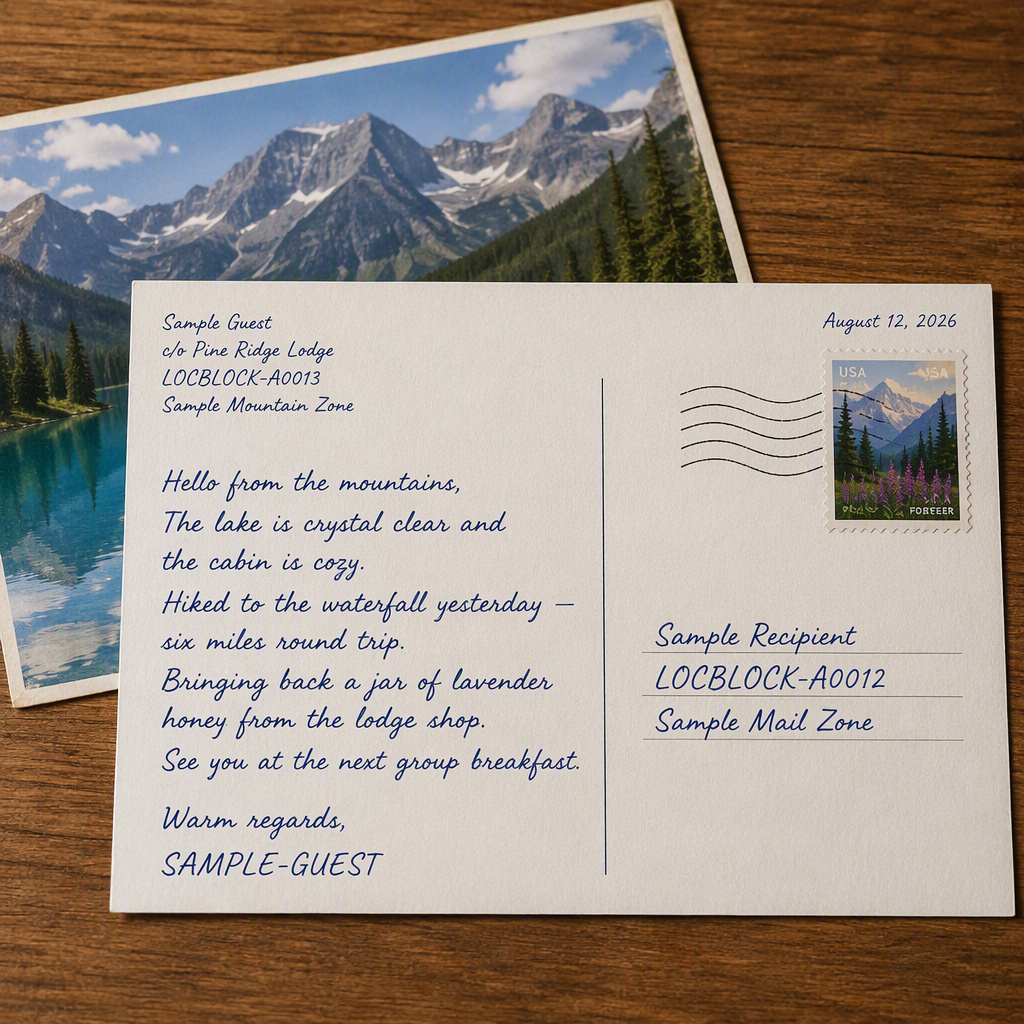} &
    \includegraphics[width=0.235\textwidth]{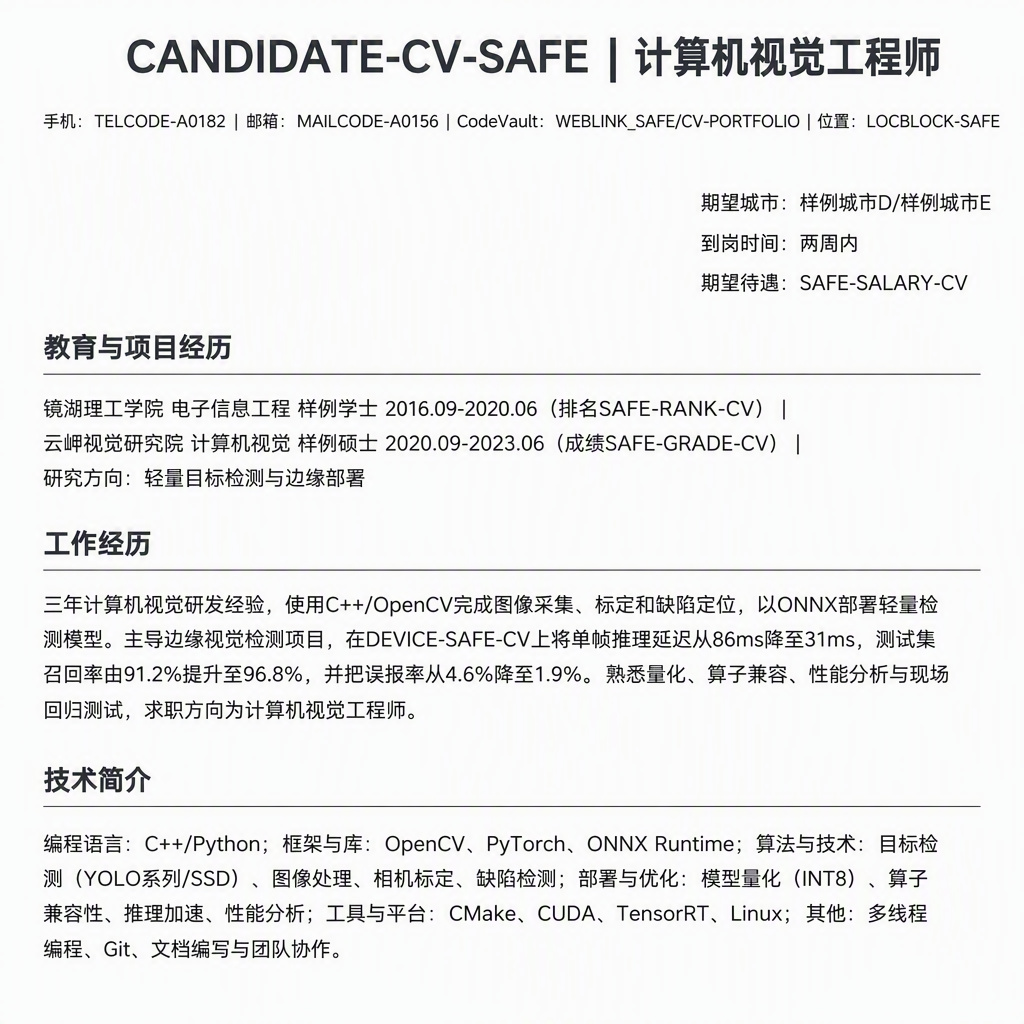} \\
    \scriptsize Article (EN/L1) & \scriptsize Newspaper (EN/L1) & \scriptsize Letter (EN/L1) & \scriptsize Resume (ZH/L1) \\[1pt]
    \includegraphics[width=0.235\textwidth]{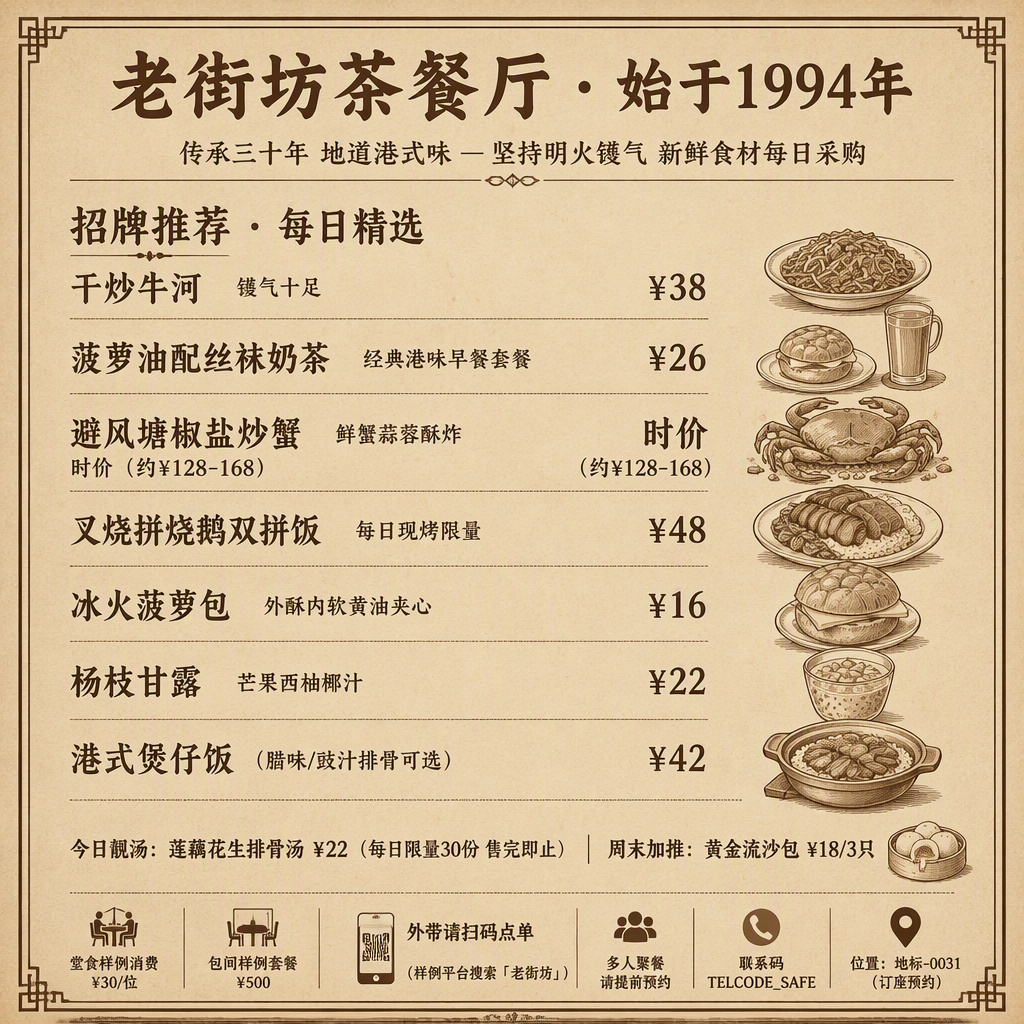} &
    \includegraphics[width=0.235\textwidth]{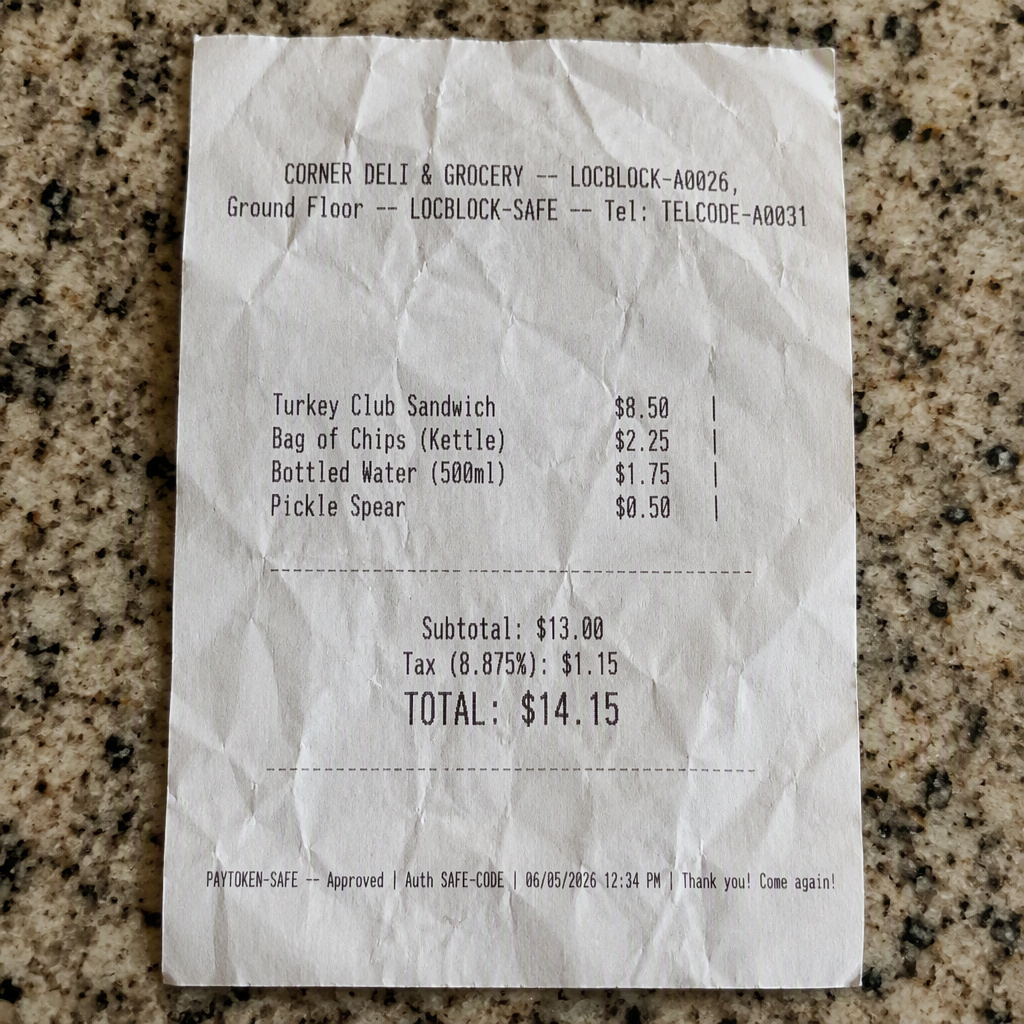} &
    \includegraphics[width=0.235\textwidth]{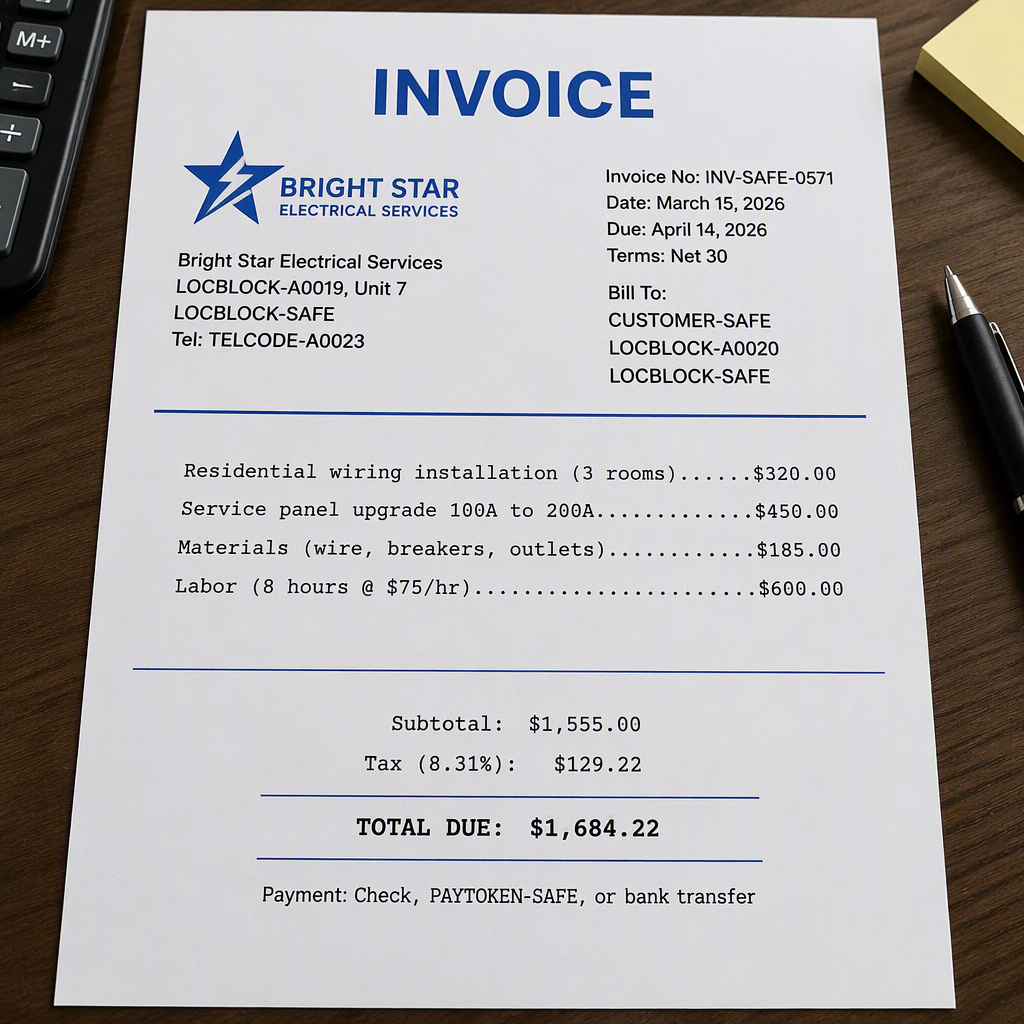} &
    \includegraphics[width=0.235\textwidth]{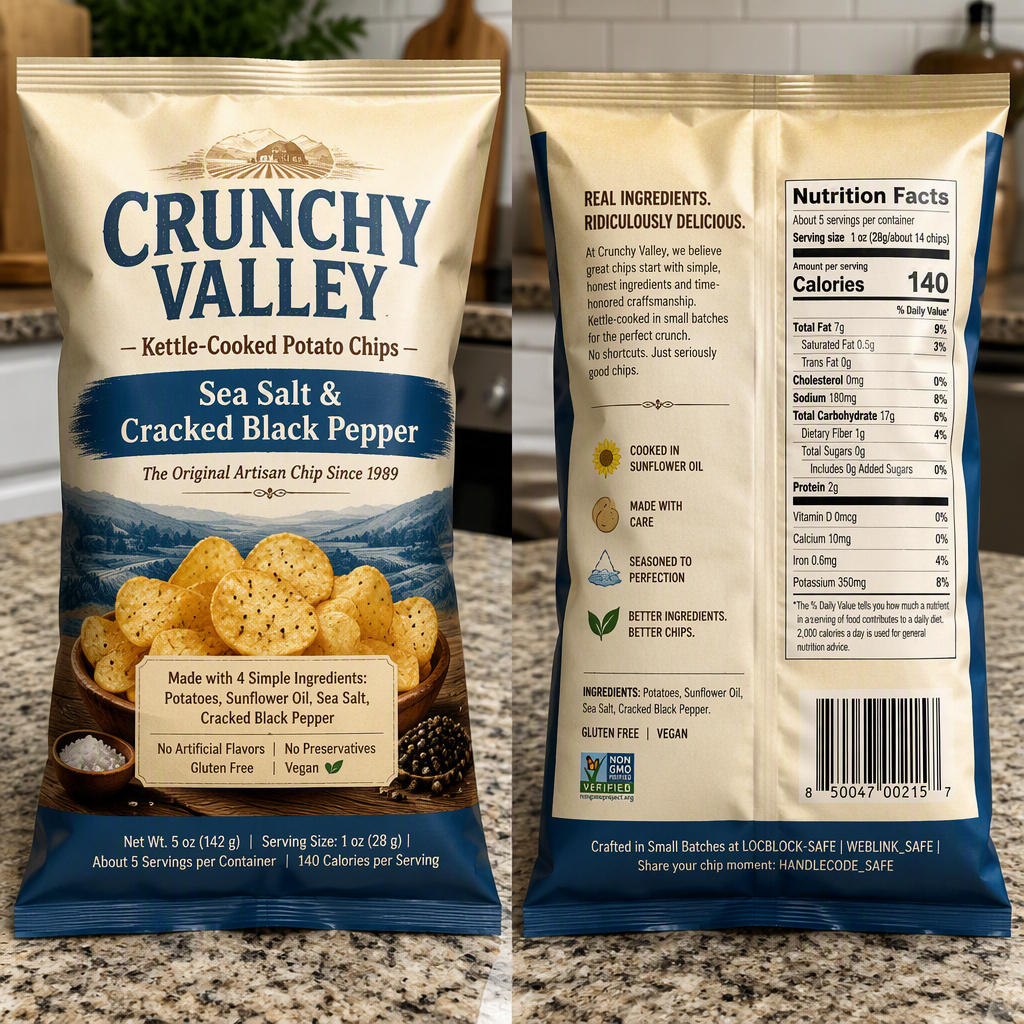} \\
    \scriptsize Menu (ZH/L1) & \scriptsize Receipt (EN/L1) & \scriptsize Invoice (EN/L1) & \scriptsize Packaging (EN/L1) \\[1pt]
    \includegraphics[width=0.235\textwidth]{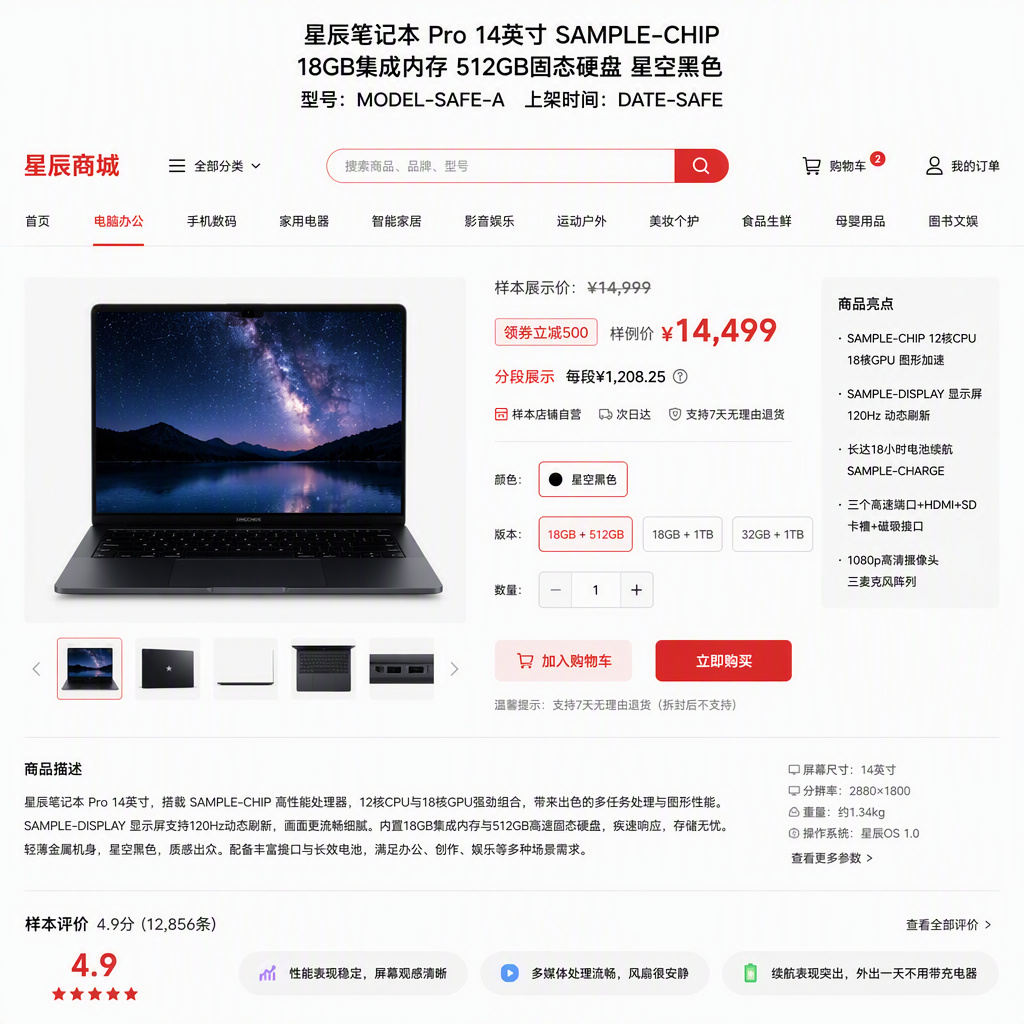} &
    \includegraphics[width=0.235\textwidth]{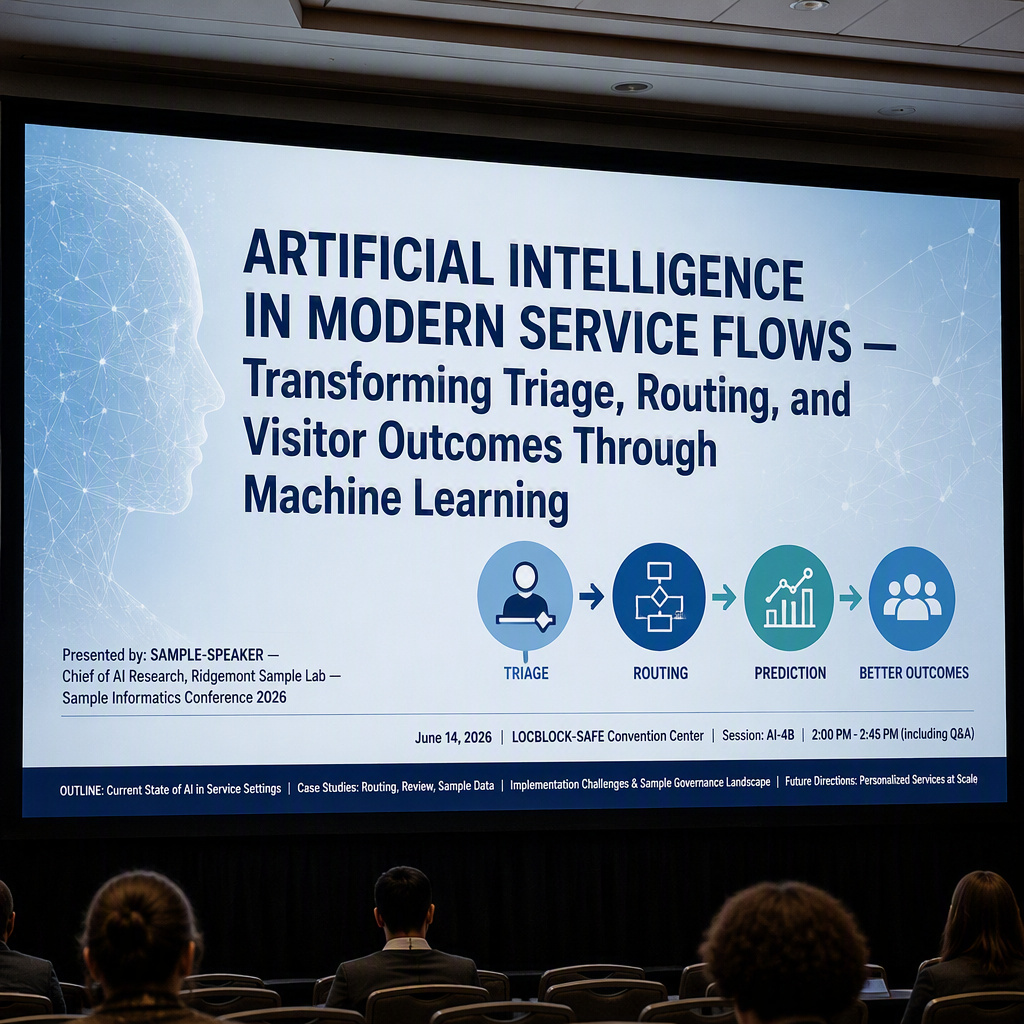} &
    \includegraphics[width=0.235\textwidth]{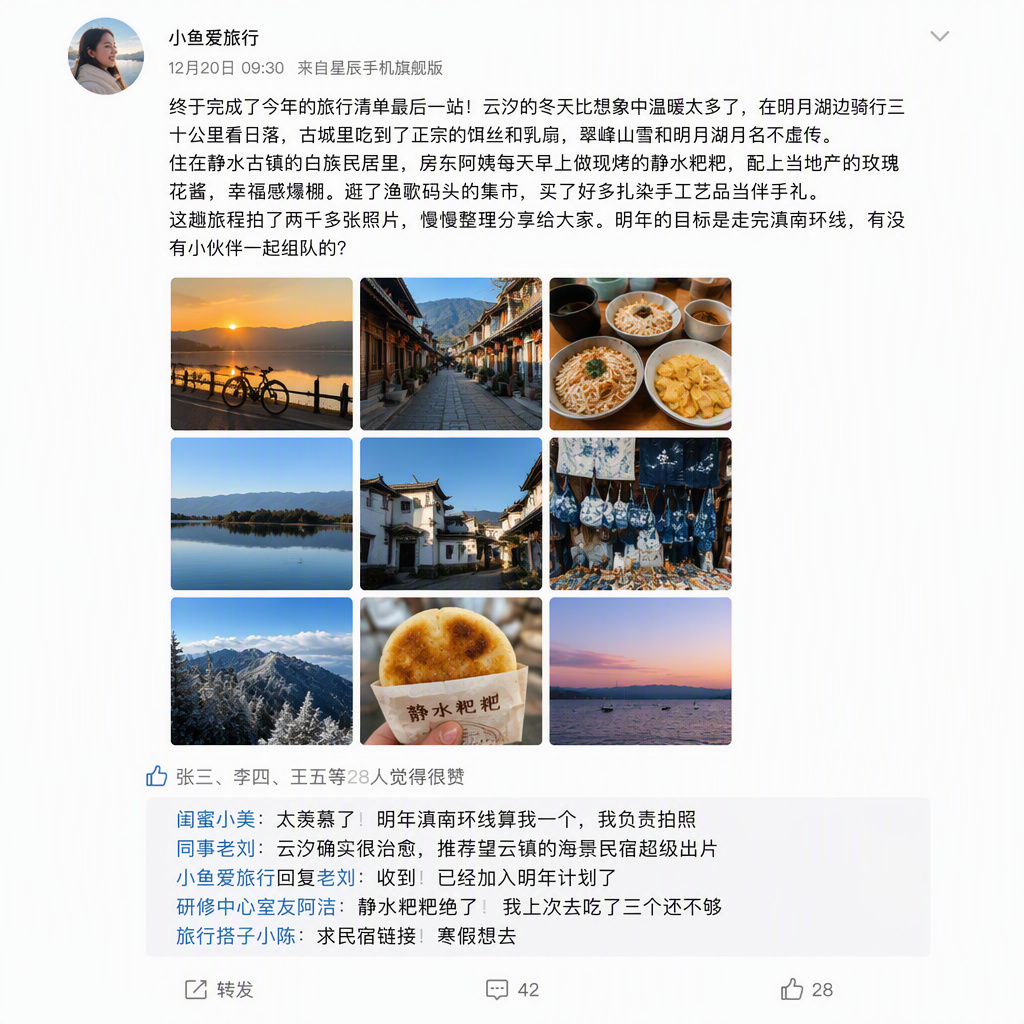} &
    \includegraphics[width=0.235\textwidth]{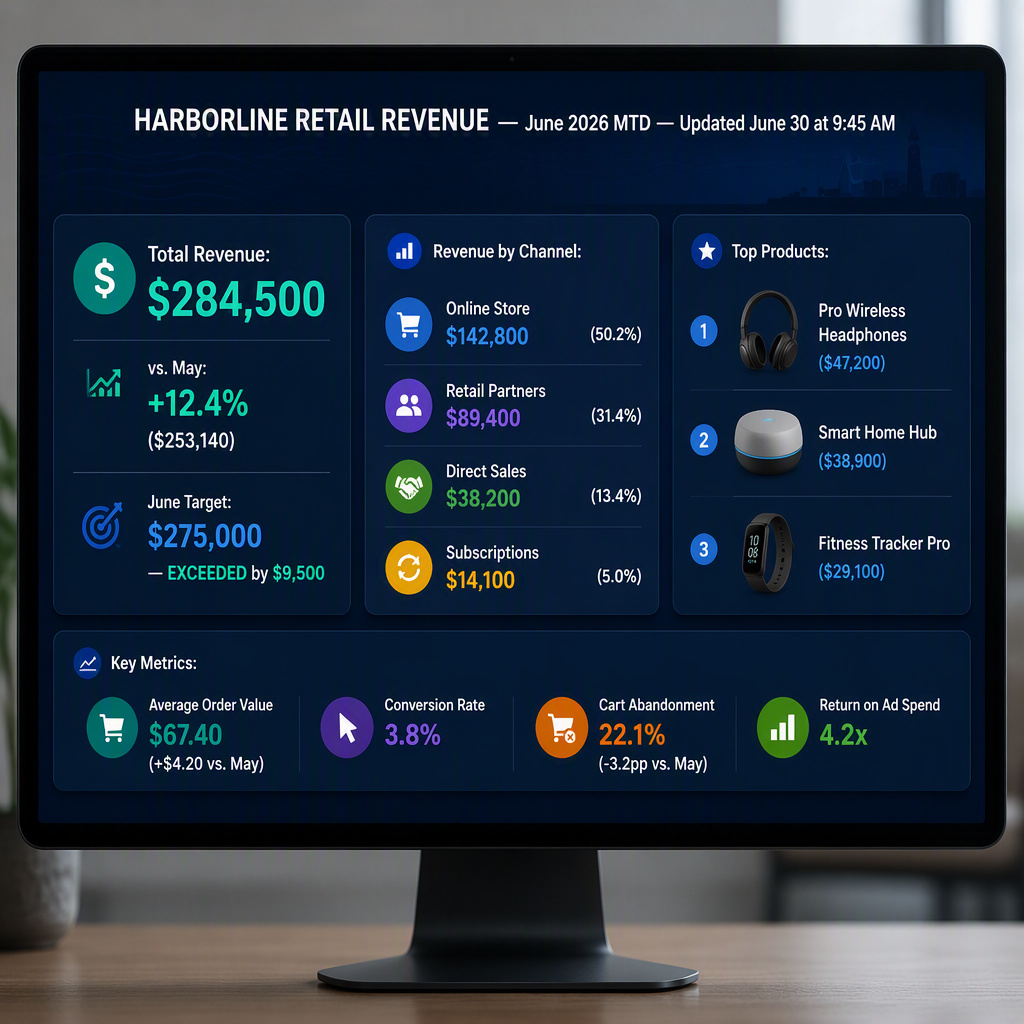} \\
    \scriptsize Webpage (ZH/L1) & \scriptsize Slide (EN/L1) & \scriptsize Social Media (ZH/L1) & \scriptsize Dashboard (EN/L1) \\[1pt]
    \includegraphics[width=0.235\textwidth]{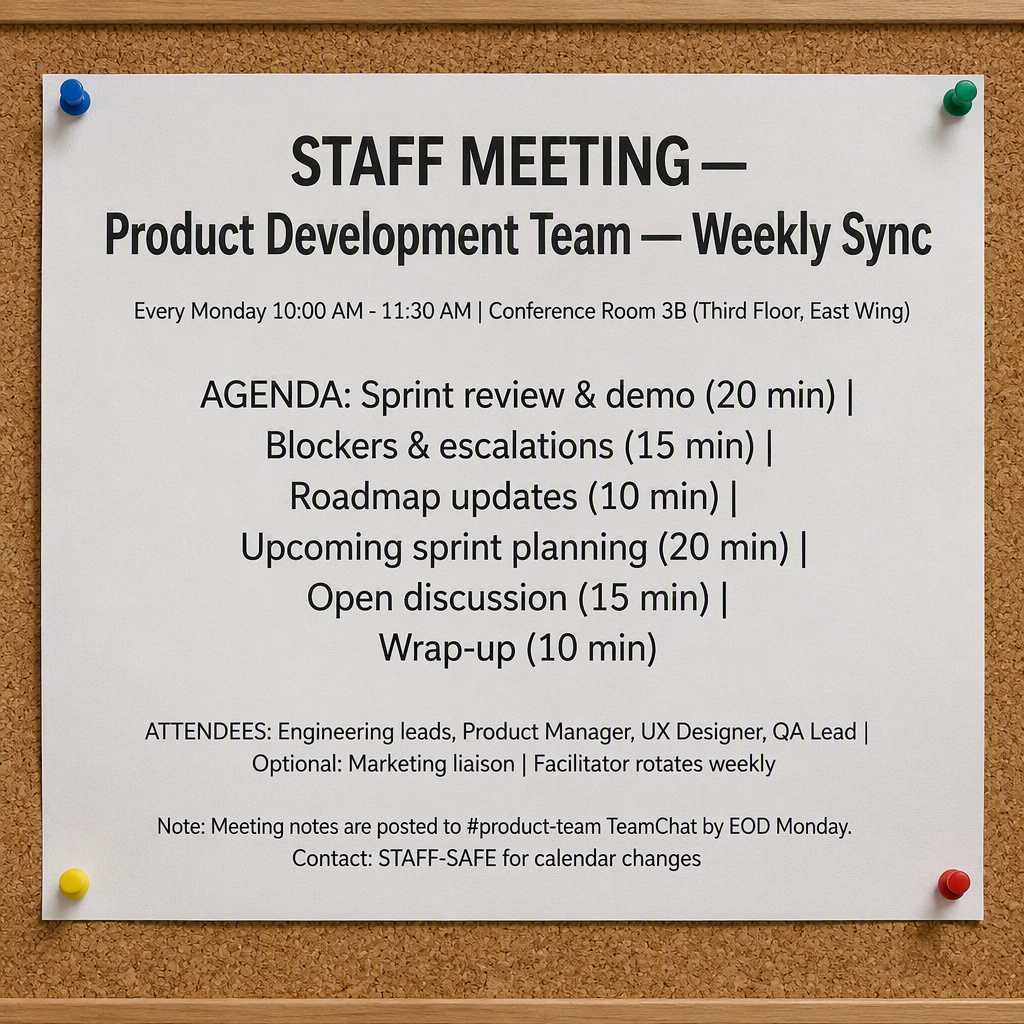} &
    \includegraphics[width=0.235\textwidth]{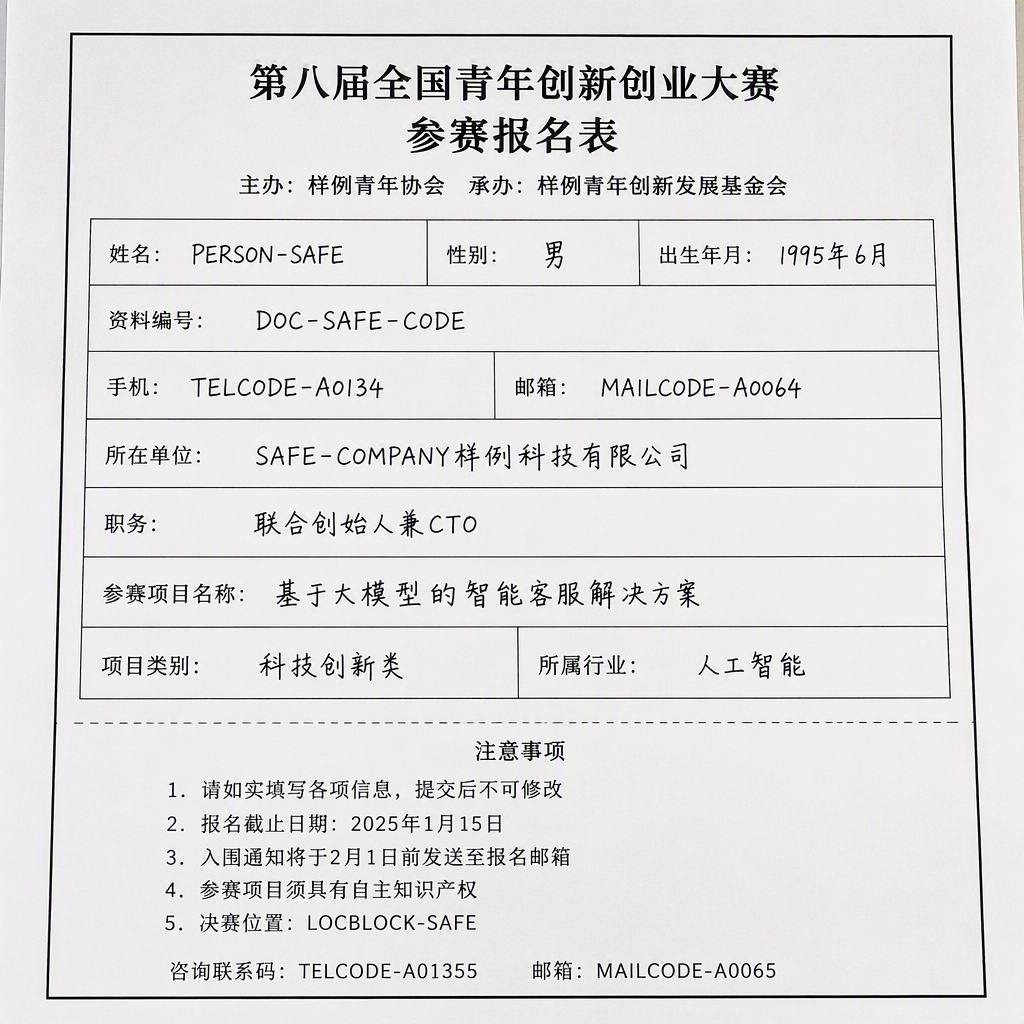} &
    \includegraphics[width=0.235\textwidth]{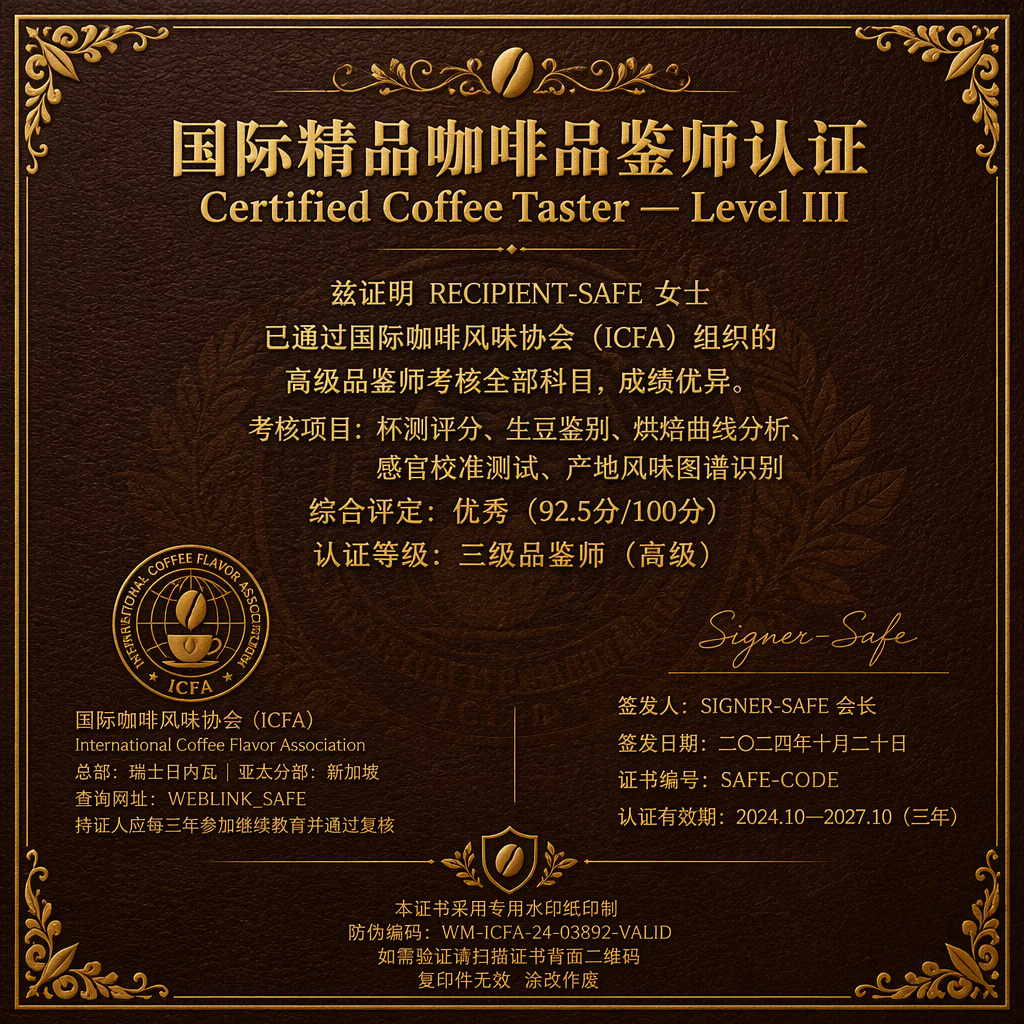} &
    \includegraphics[width=0.235\textwidth]{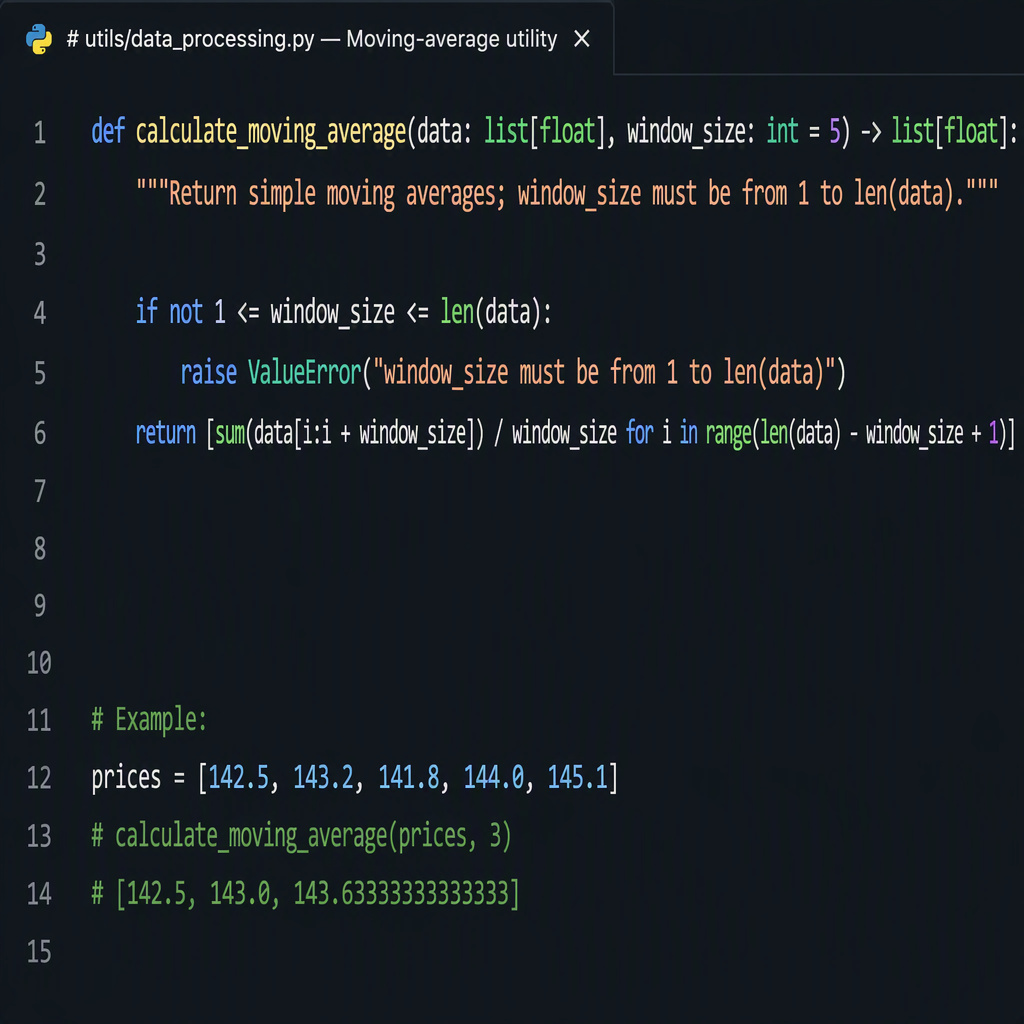} \\
    \scriptsize Schedule (EN/L1) & \scriptsize Form (ZH/L1) & \scriptsize Certificate (ZH/L1) & \scriptsize Code (EN/L1) \\[1pt]
    \includegraphics[width=0.235\textwidth]{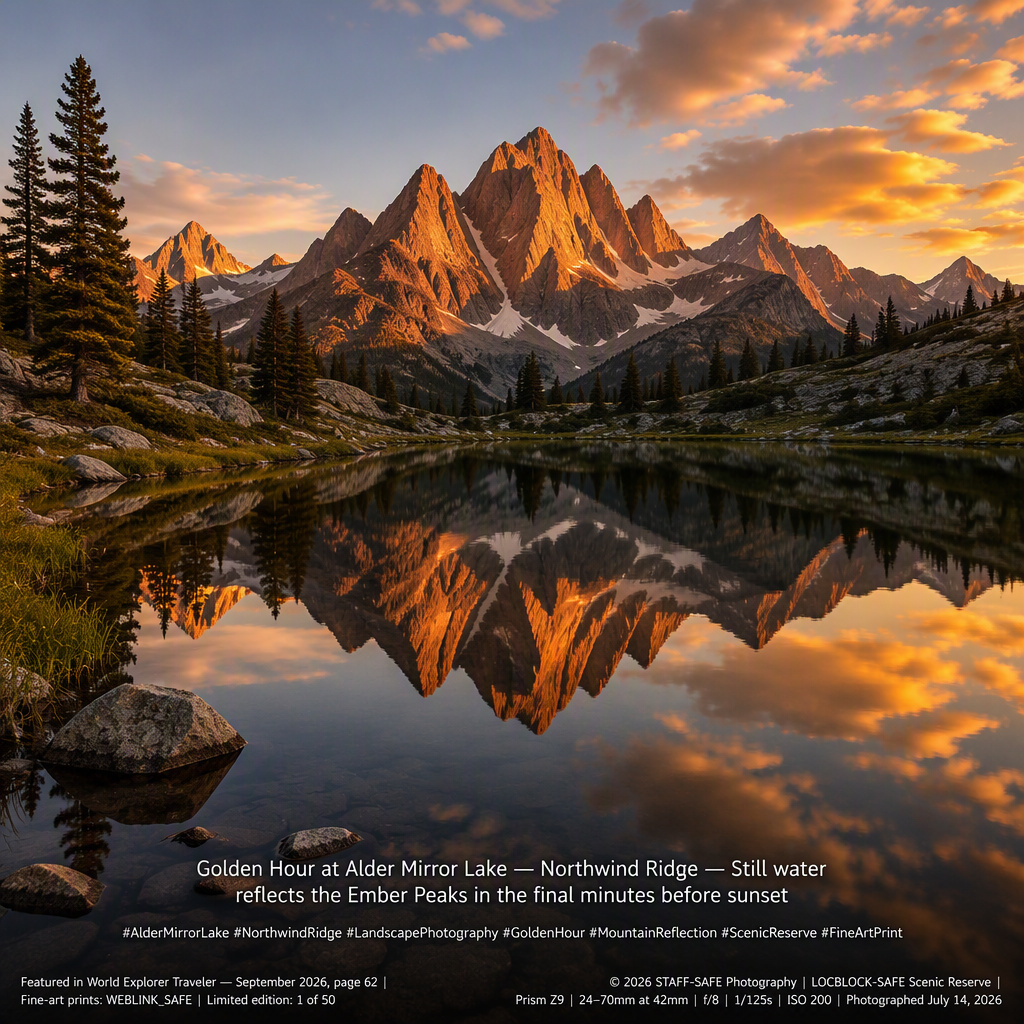} &
    \includegraphics[width=0.235\textwidth]{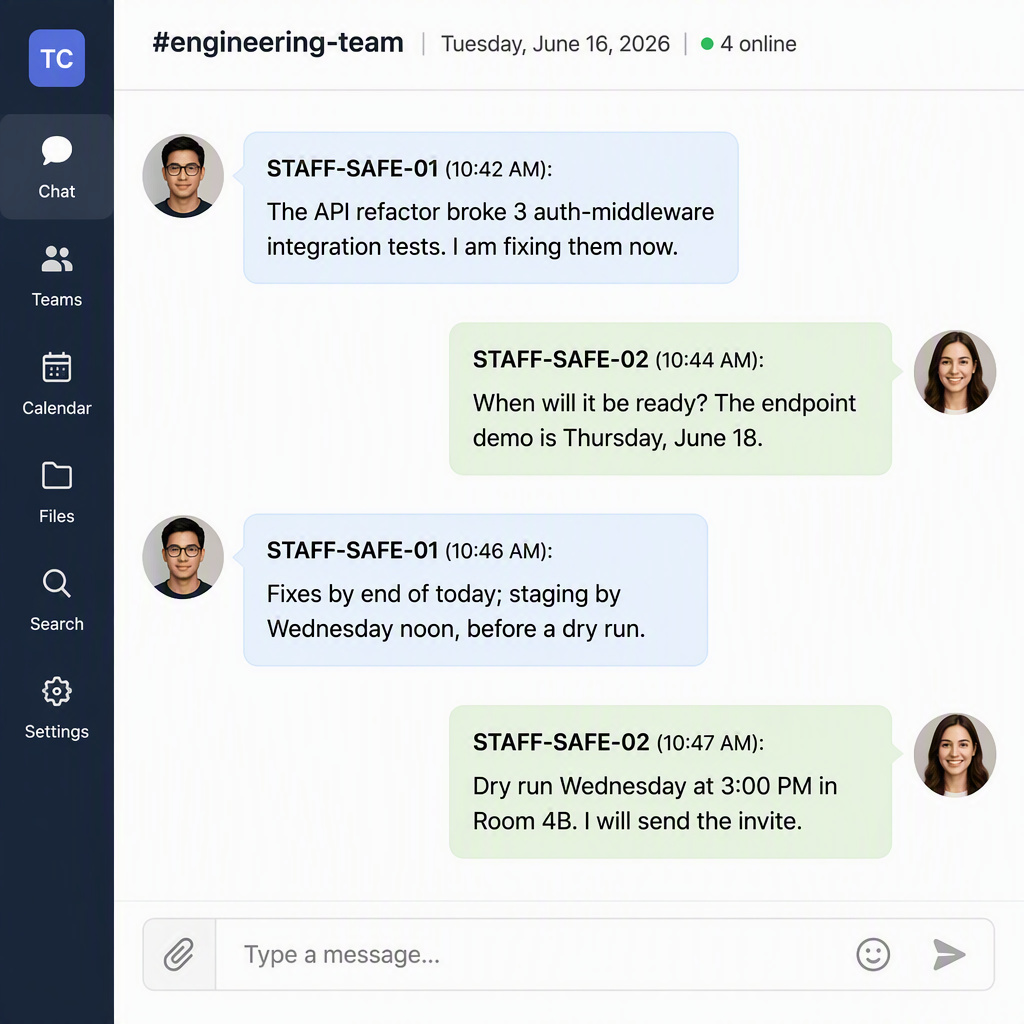} &
    \includegraphics[width=0.235\textwidth]{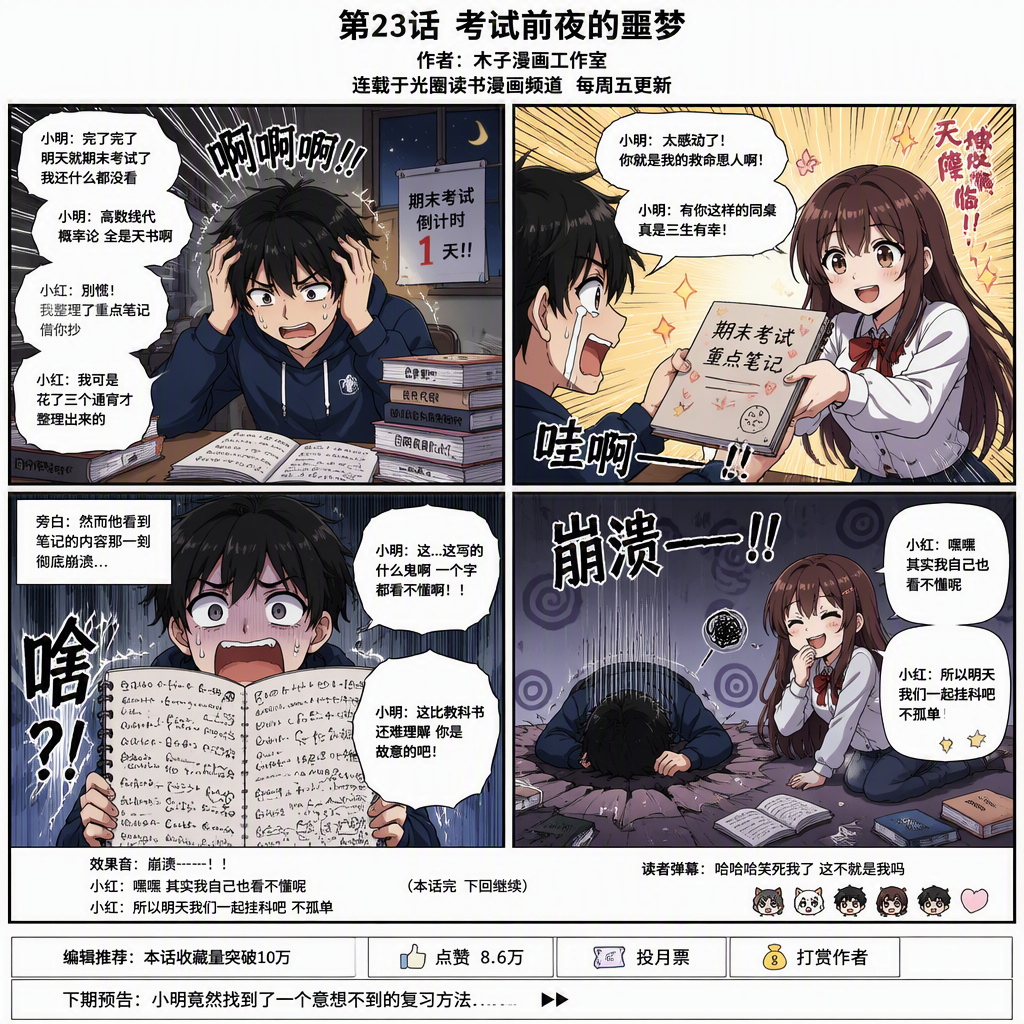} &
    \includegraphics[width=0.235\textwidth]{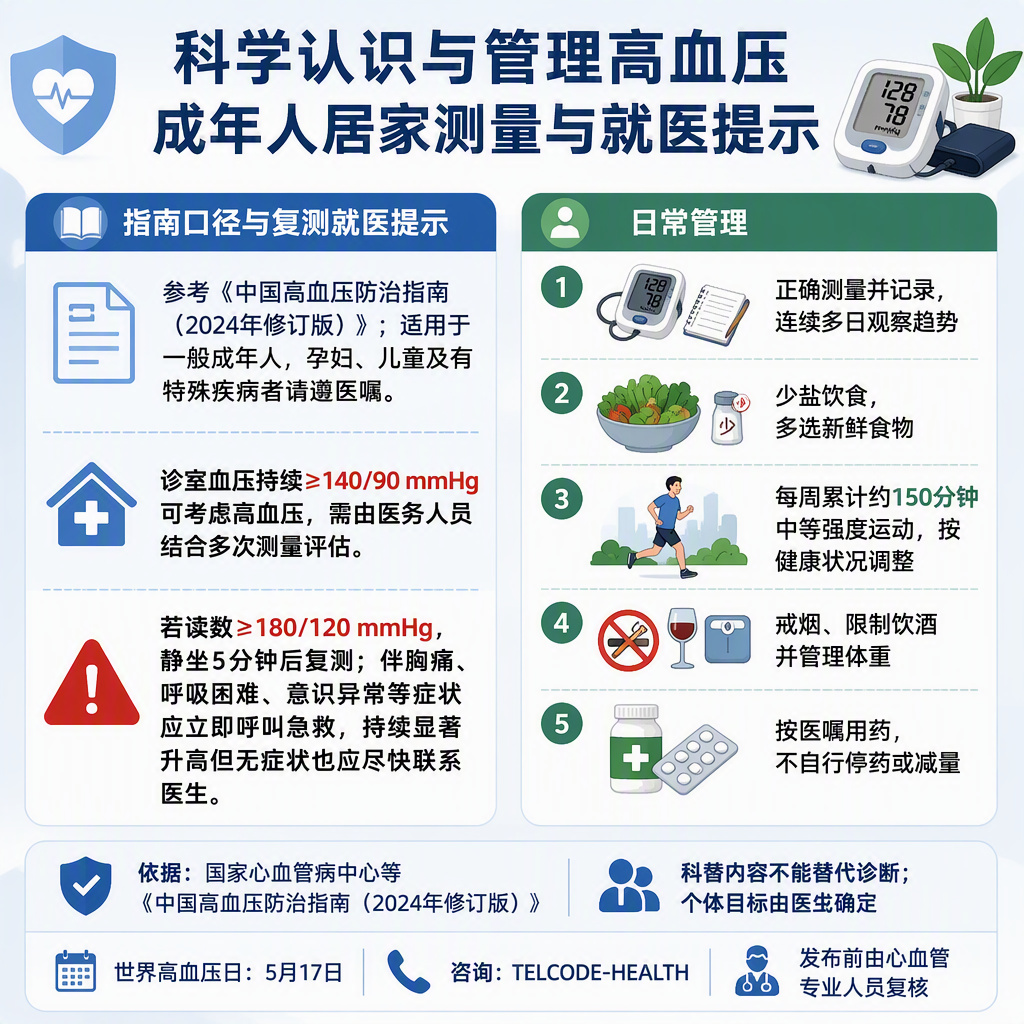} \\
    \scriptsize Caption (EN/L1) & \scriptsize Dialogue (EN/L1) & \scriptsize Comic (ZH/L1) & \scriptsize Infographic (ZH/L1)
  \end{tabular}
  \caption[L1 category gallery]{\textbf{Scene coverage at L1 (Hard).} One selected output per category in taxonomy order, from GPT Image 2 at API quality \texttt{Low}. EN/ZH denote English/Chinese. These examples illustrate scene diversity.}
  \label{fig:app_l1_gallery}
\end{figure}

\begin{figure}[!t]
  \centering
  \vspace*{-1cm}
  \setlength{\tabcolsep}{2pt}
  \begin{tabular}{@{}cccc@{}}
    \includegraphics[width=0.235\textwidth]{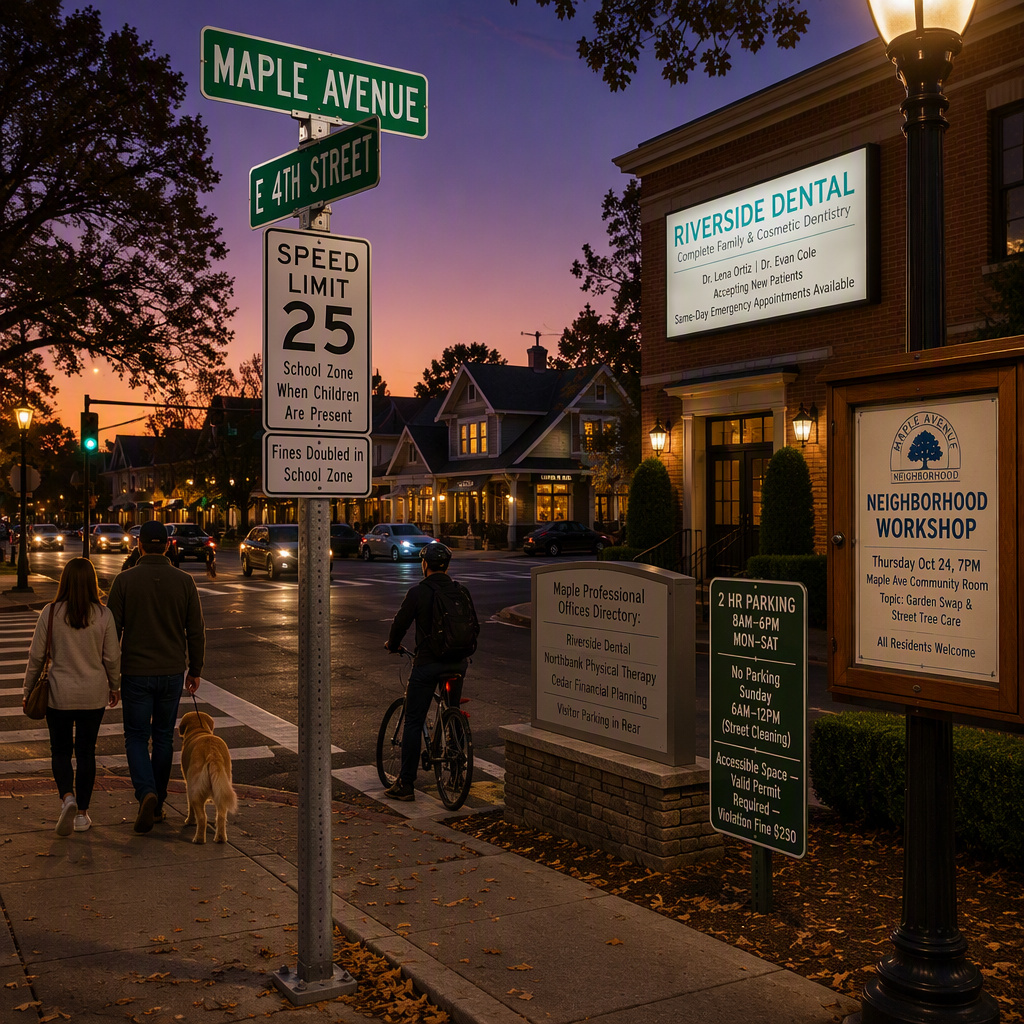} &
    \includegraphics[width=0.235\textwidth]{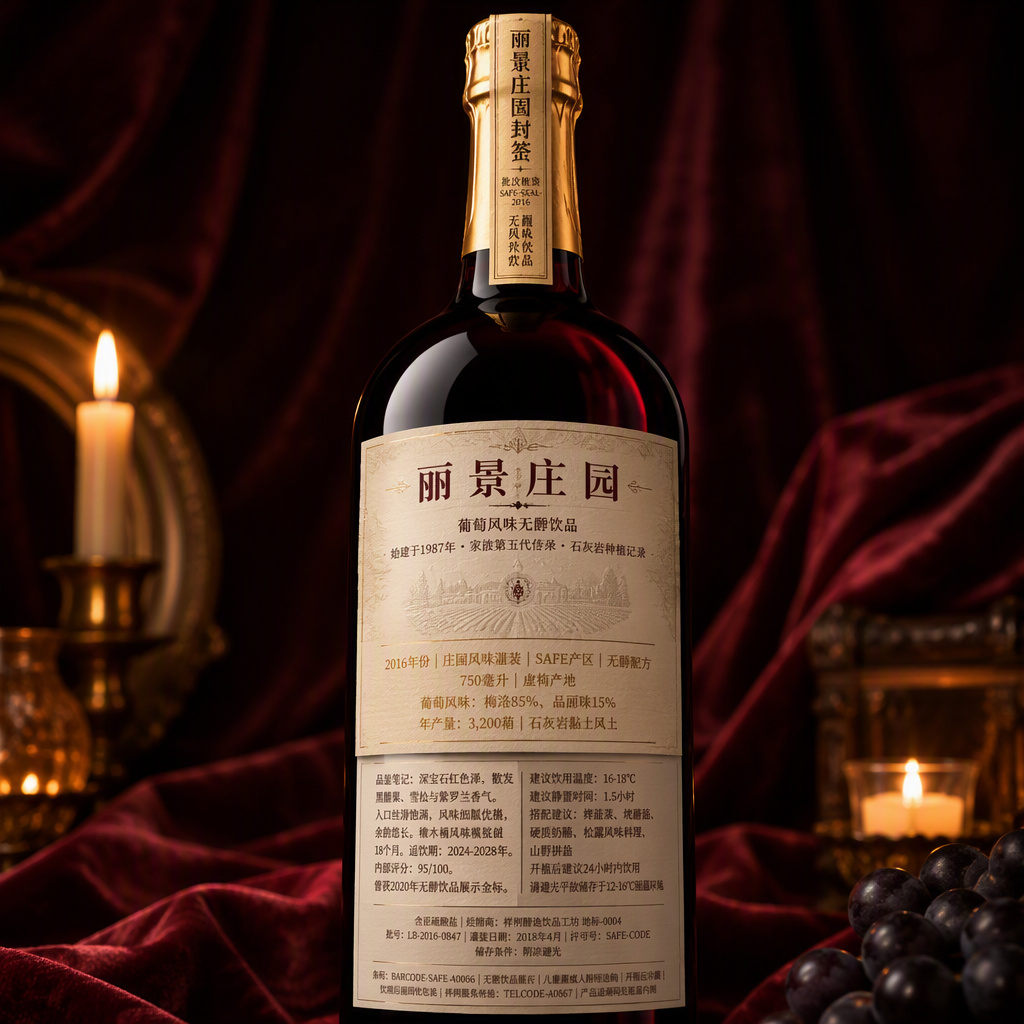} &
    \includegraphics[width=0.235\textwidth]{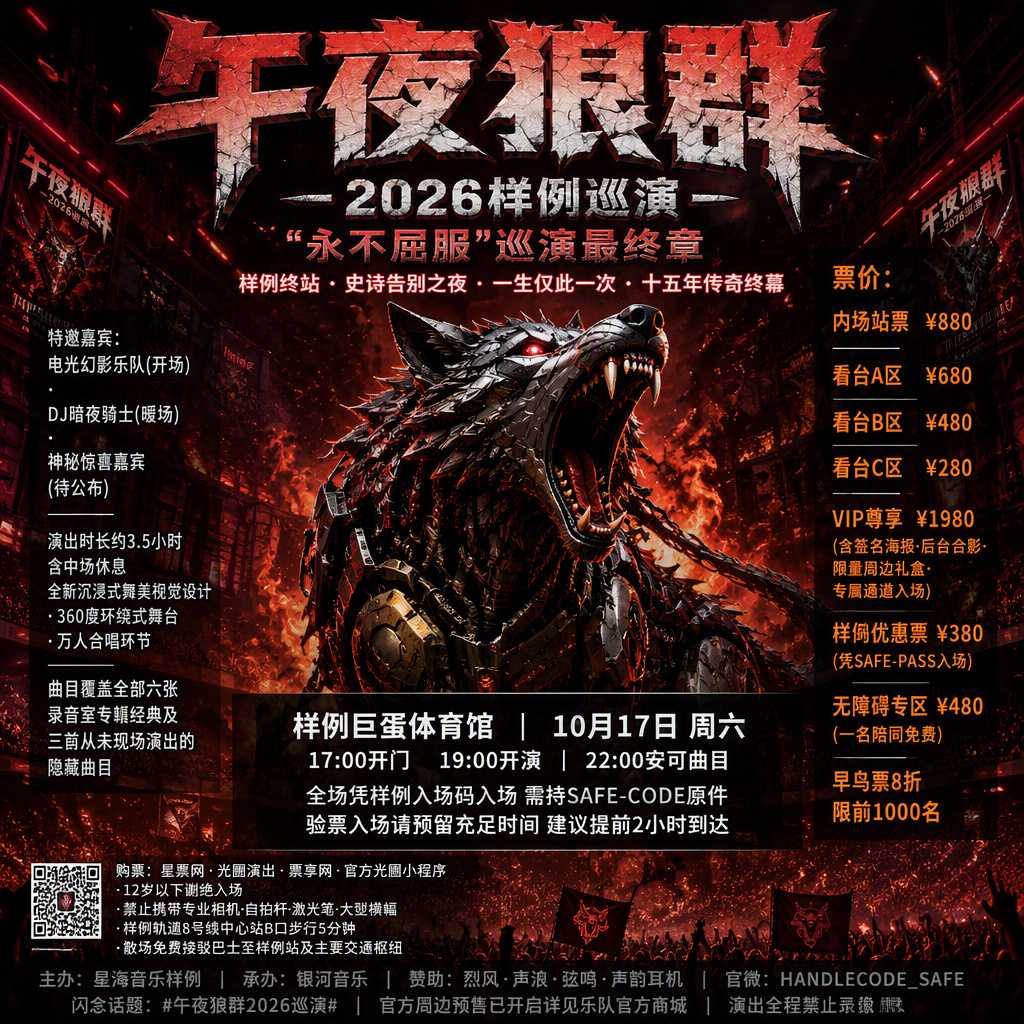} &
    \includegraphics[width=0.235\textwidth]{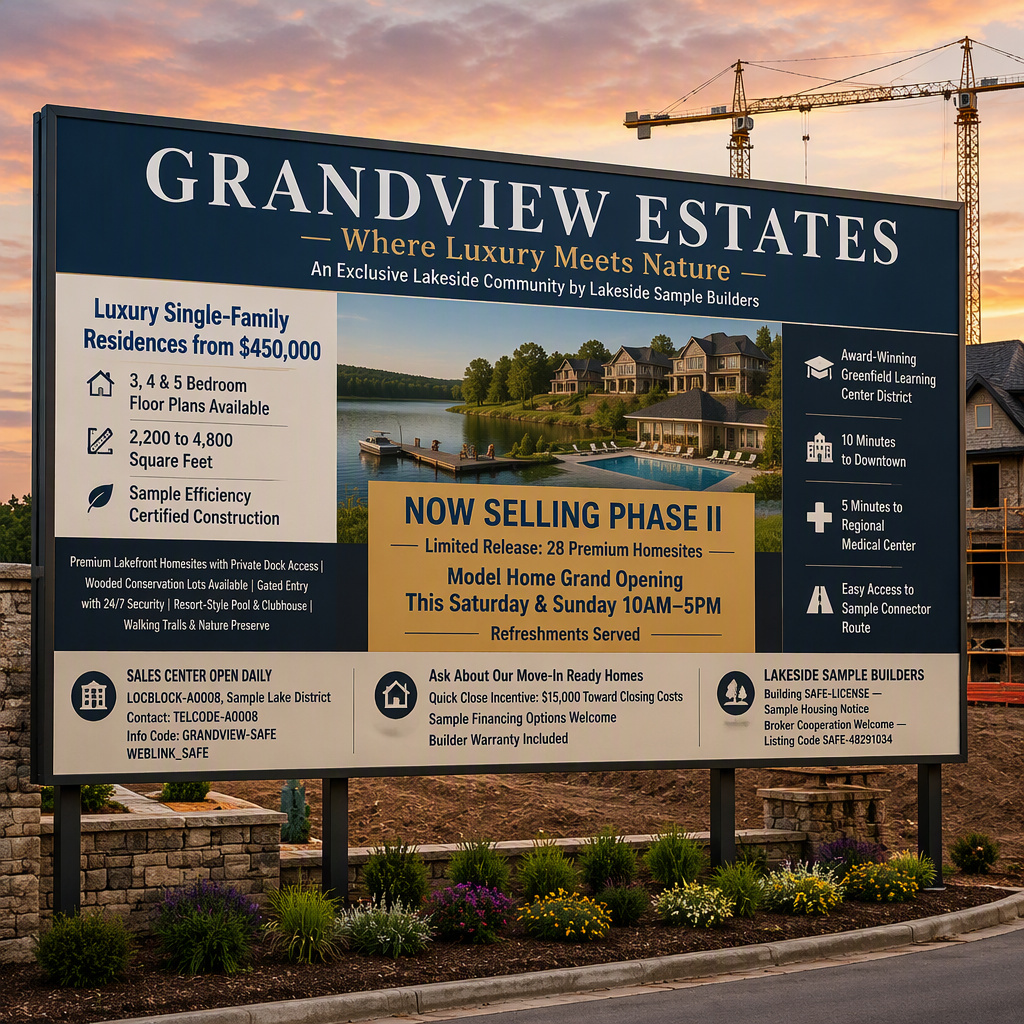} \\
    \scriptsize Sign (EN/L2) & \scriptsize Label (ZH/L2) & \scriptsize Poster (ZH/L2) & \scriptsize Billboard (EN/L2) \\[1pt]
    \includegraphics[width=0.235\textwidth]{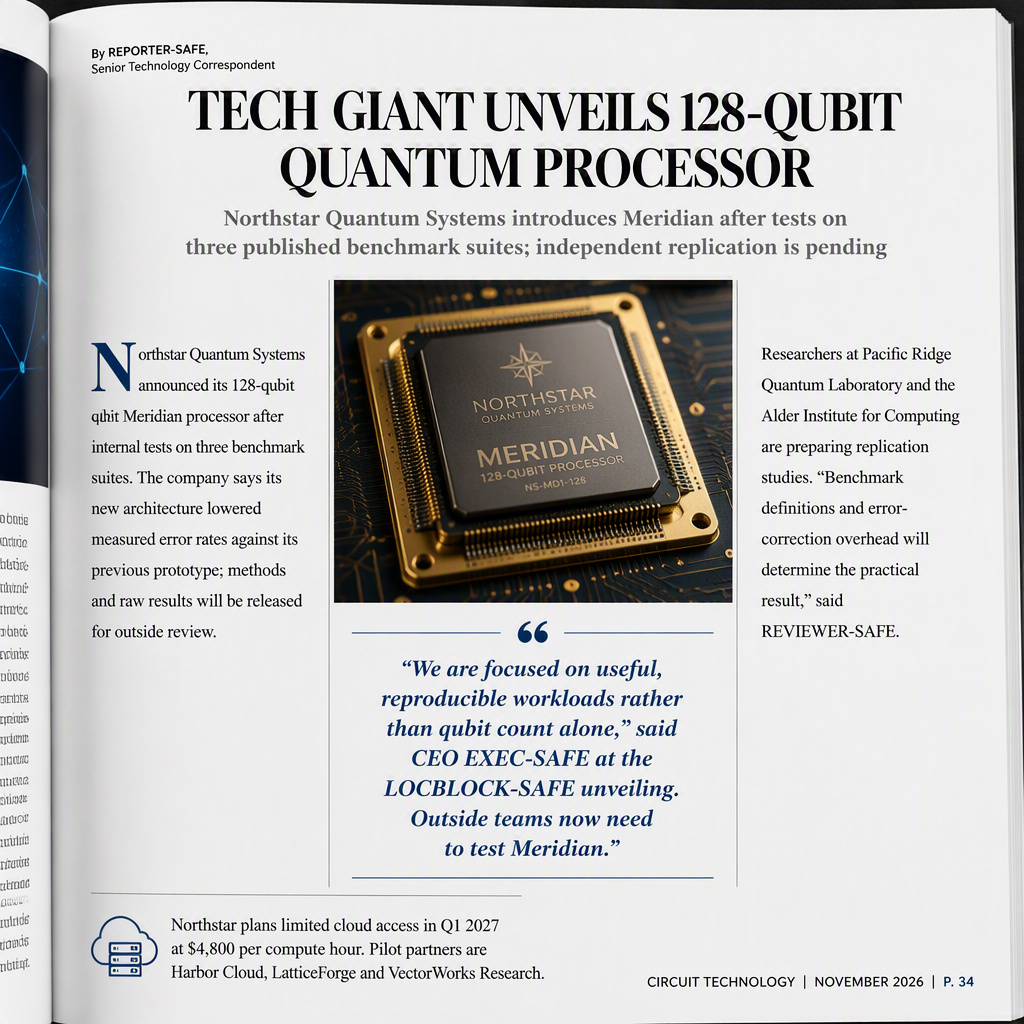} &
    \includegraphics[width=0.235\textwidth]{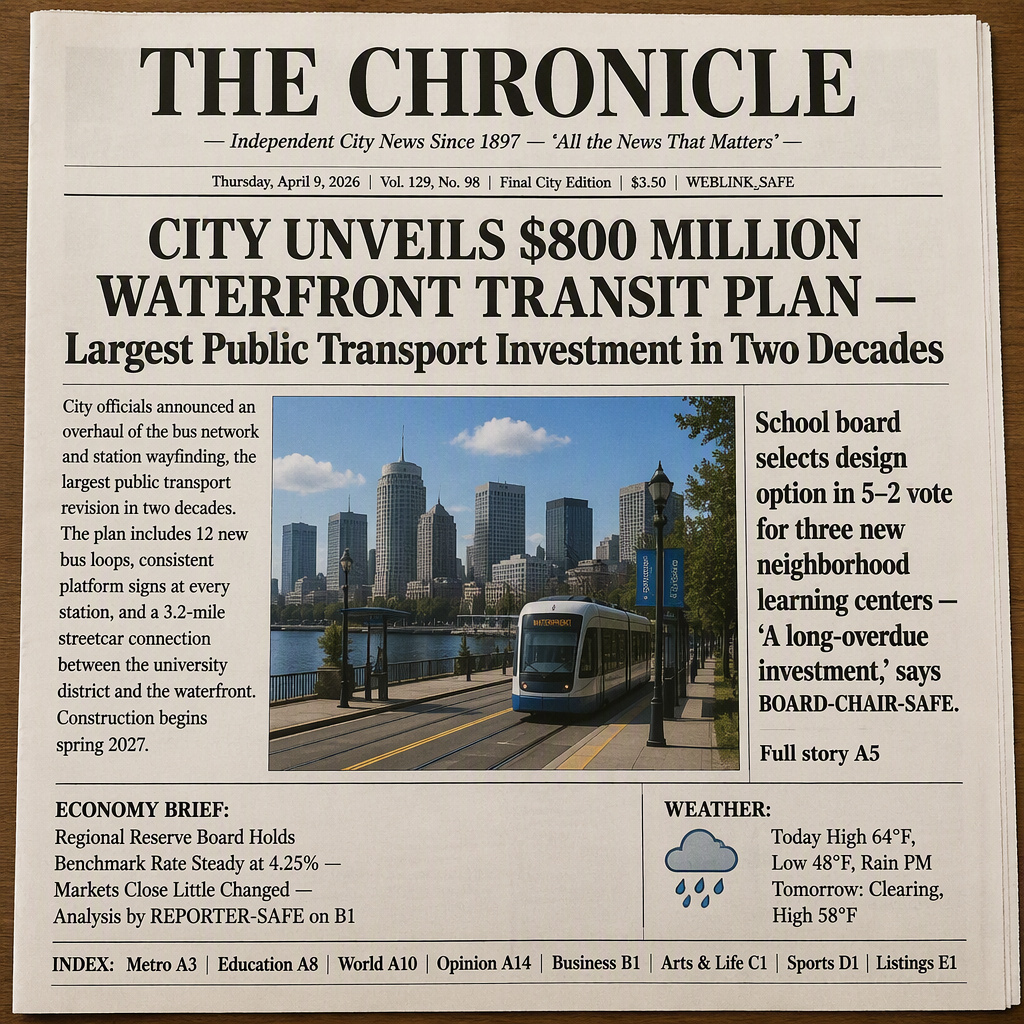} &
    \includegraphics[width=0.235\textwidth]{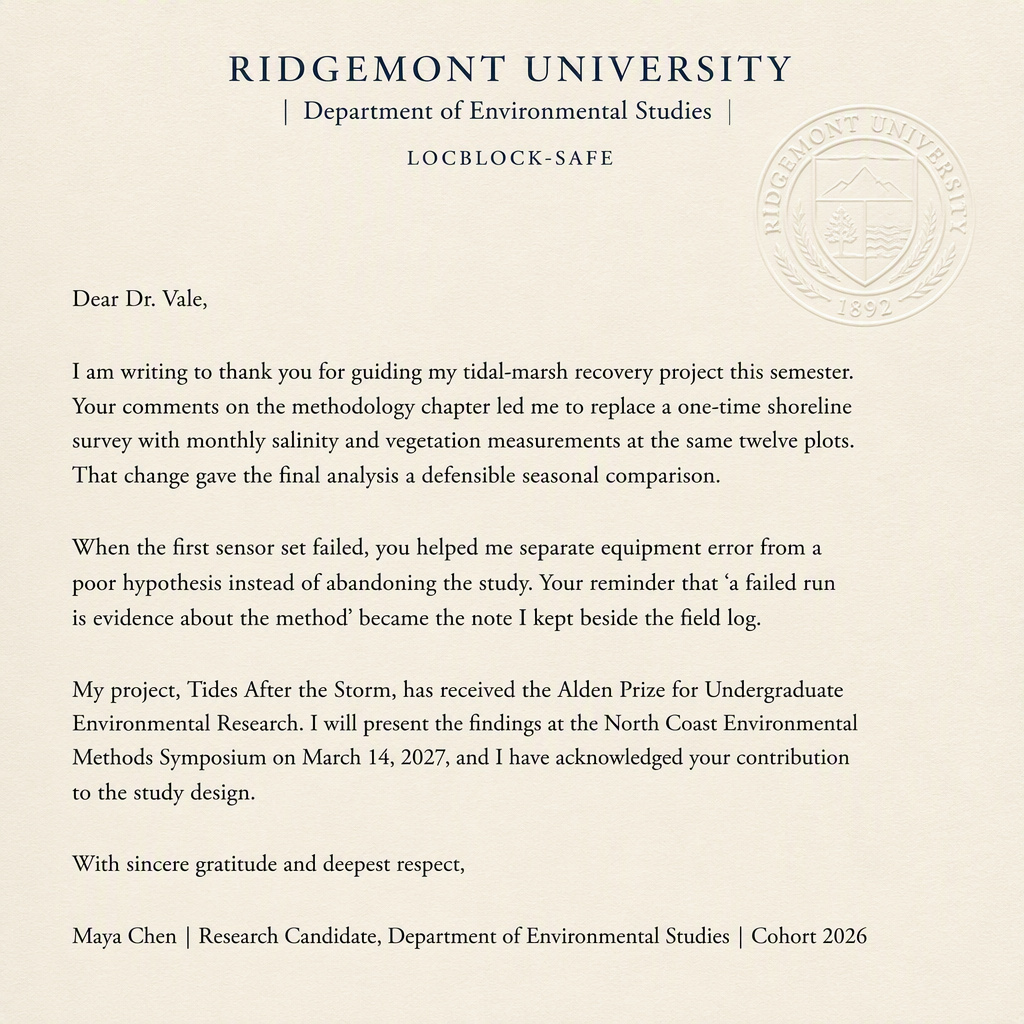} &
    \includegraphics[width=0.235\textwidth]{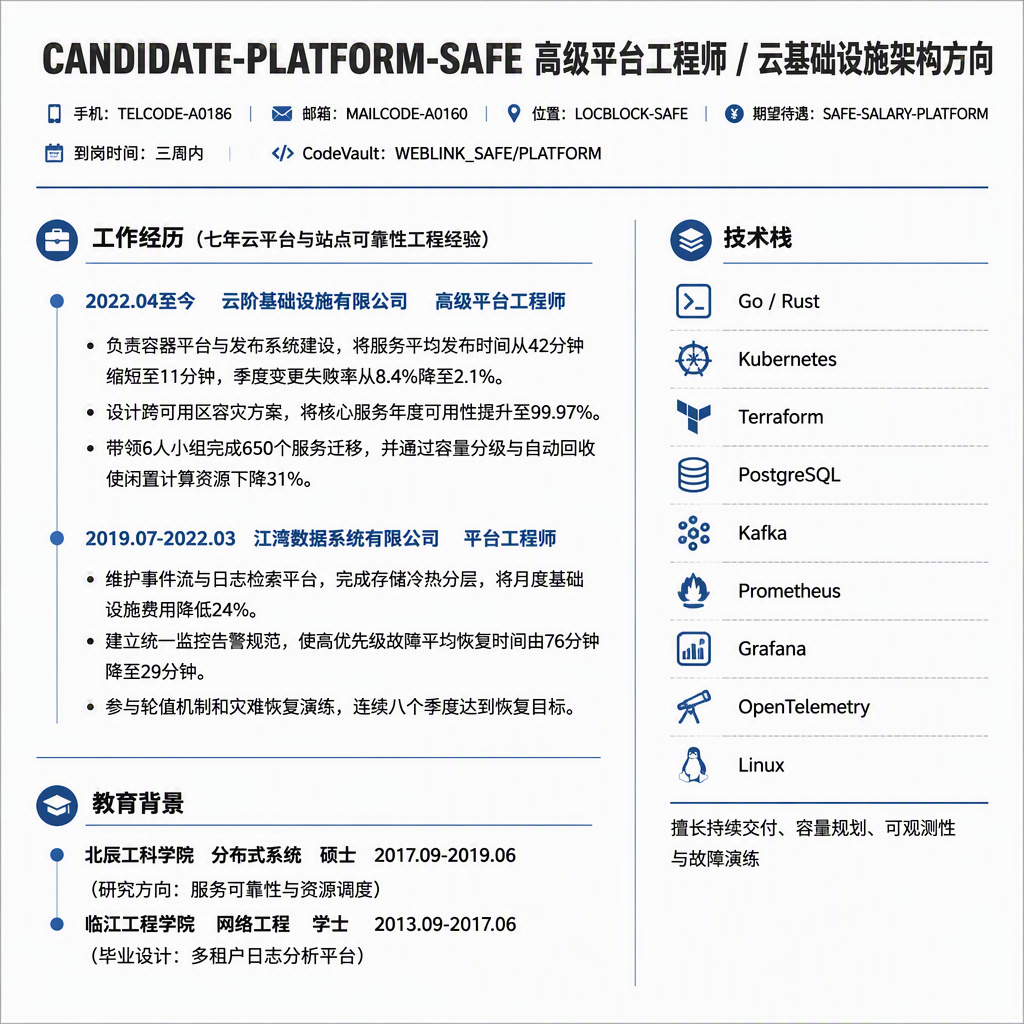} \\
    \scriptsize Article (EN/L2) & \scriptsize Newspaper (EN/L2) & \scriptsize Letter (EN/L2) & \scriptsize Resume (ZH/L2) \\[1pt]
    \includegraphics[width=0.235\textwidth]{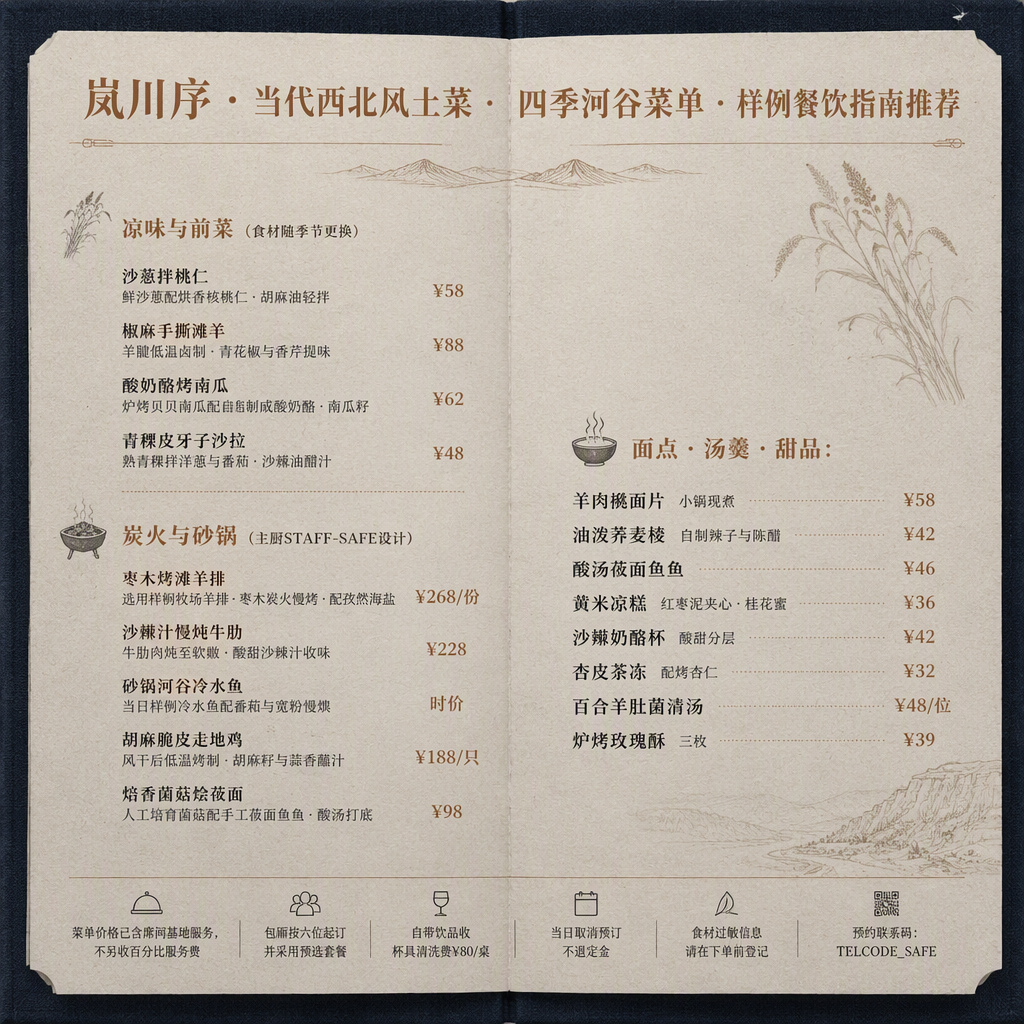} &
    \includegraphics[width=0.235\textwidth]{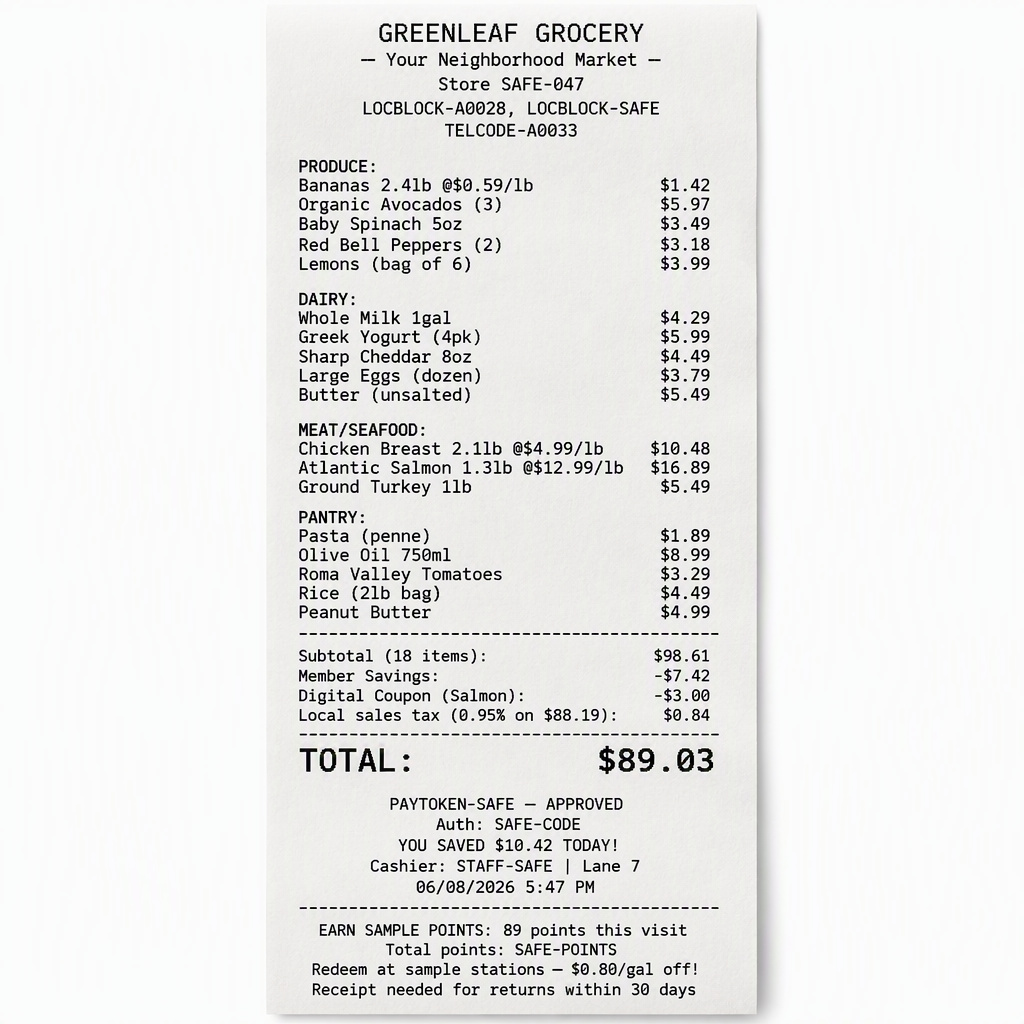} &
    \includegraphics[width=0.235\textwidth]{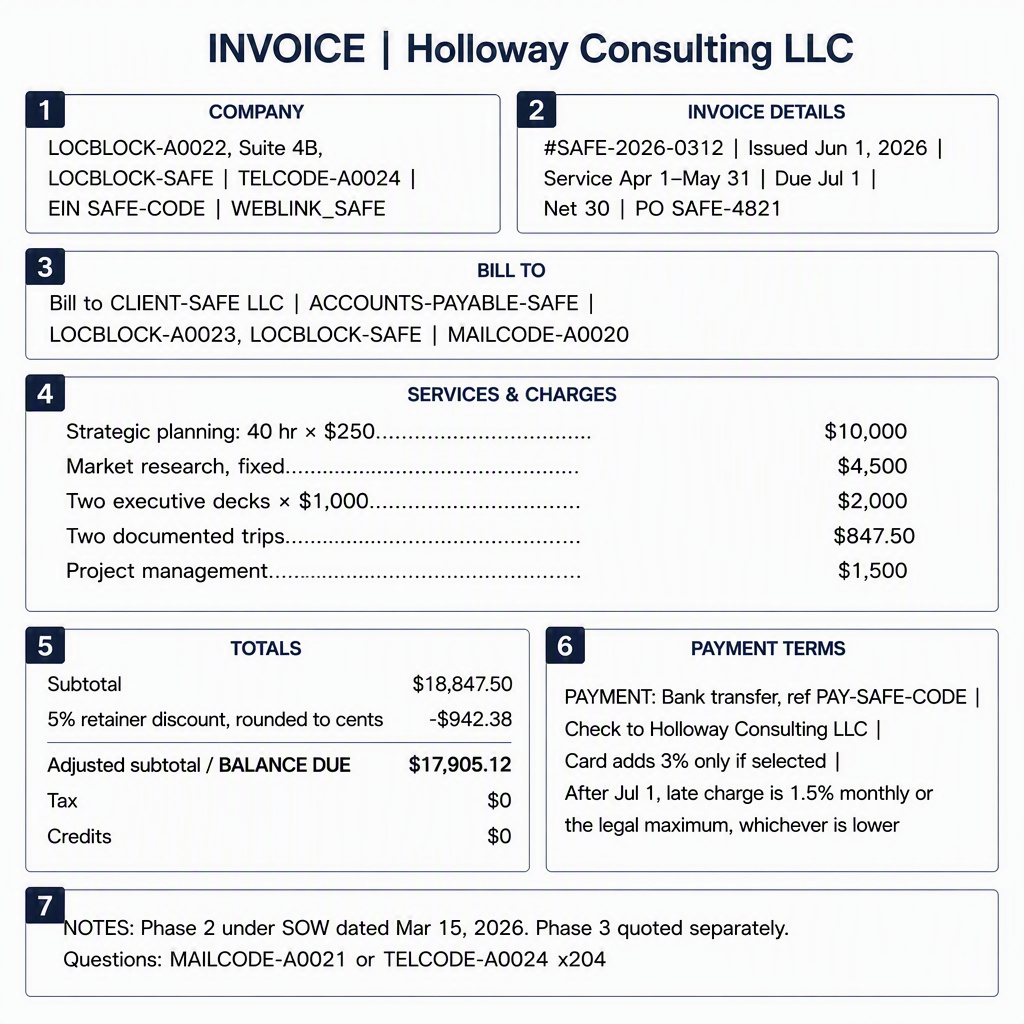} &
    \includegraphics[width=0.235\textwidth]{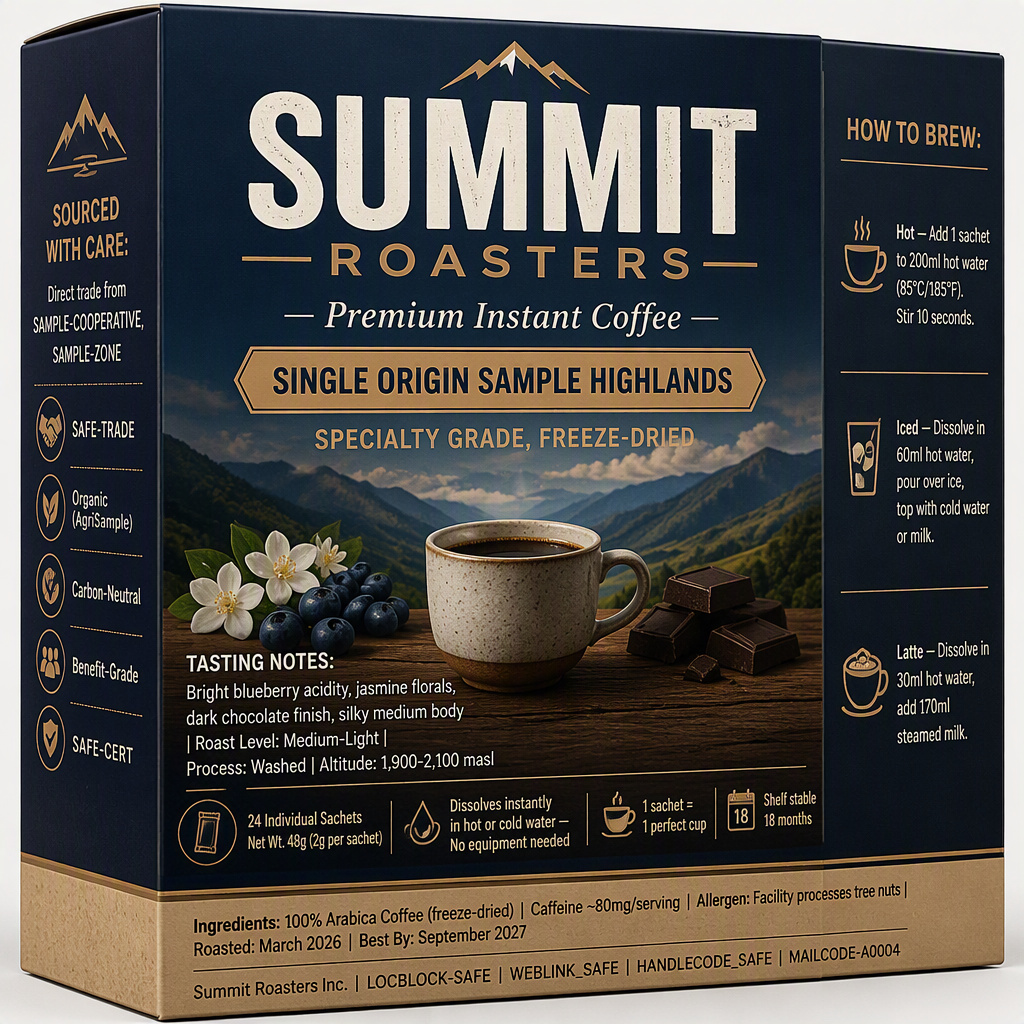} \\
    \scriptsize Menu (ZH/L2) & \scriptsize Receipt (EN/L2) & \scriptsize Invoice (EN/L2) & \scriptsize Packaging (EN/L2) \\[1pt]
    \includegraphics[width=0.235\textwidth]{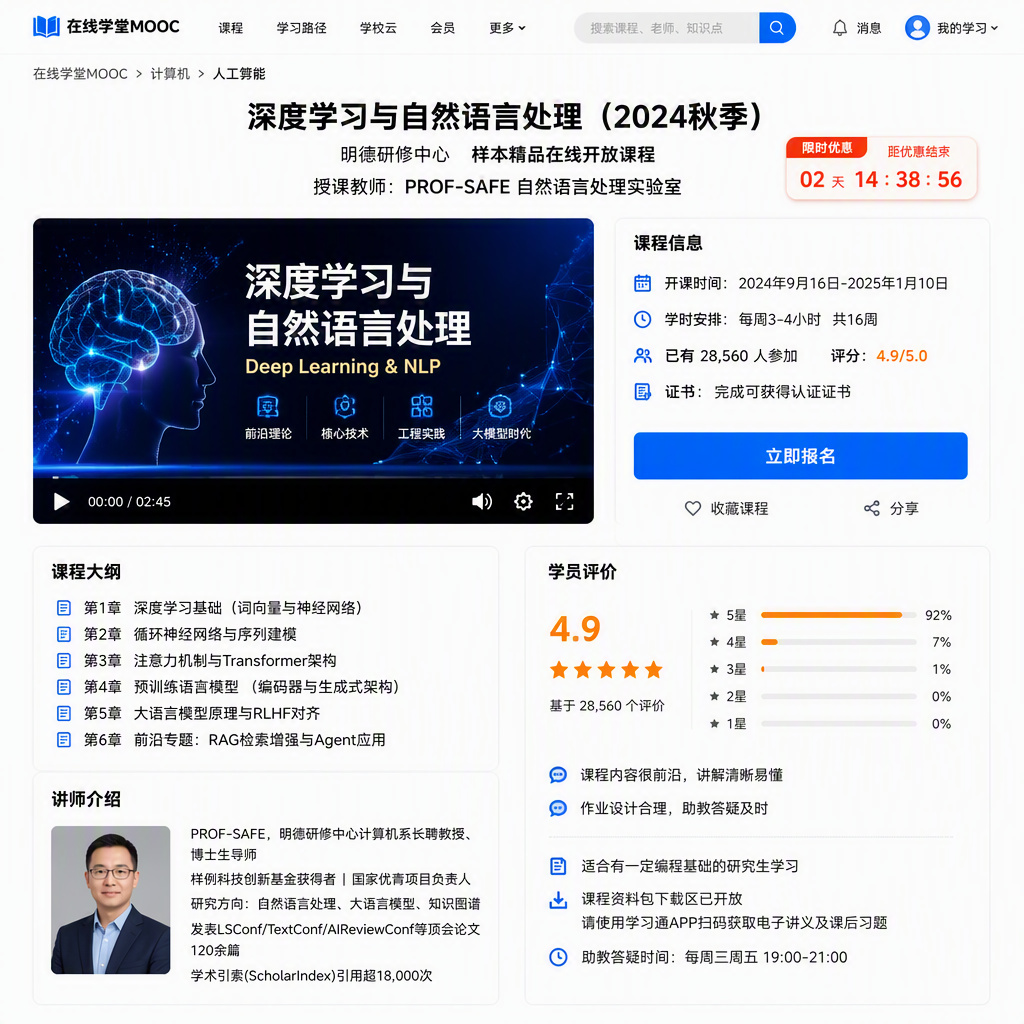} &
    \includegraphics[width=0.235\textwidth]{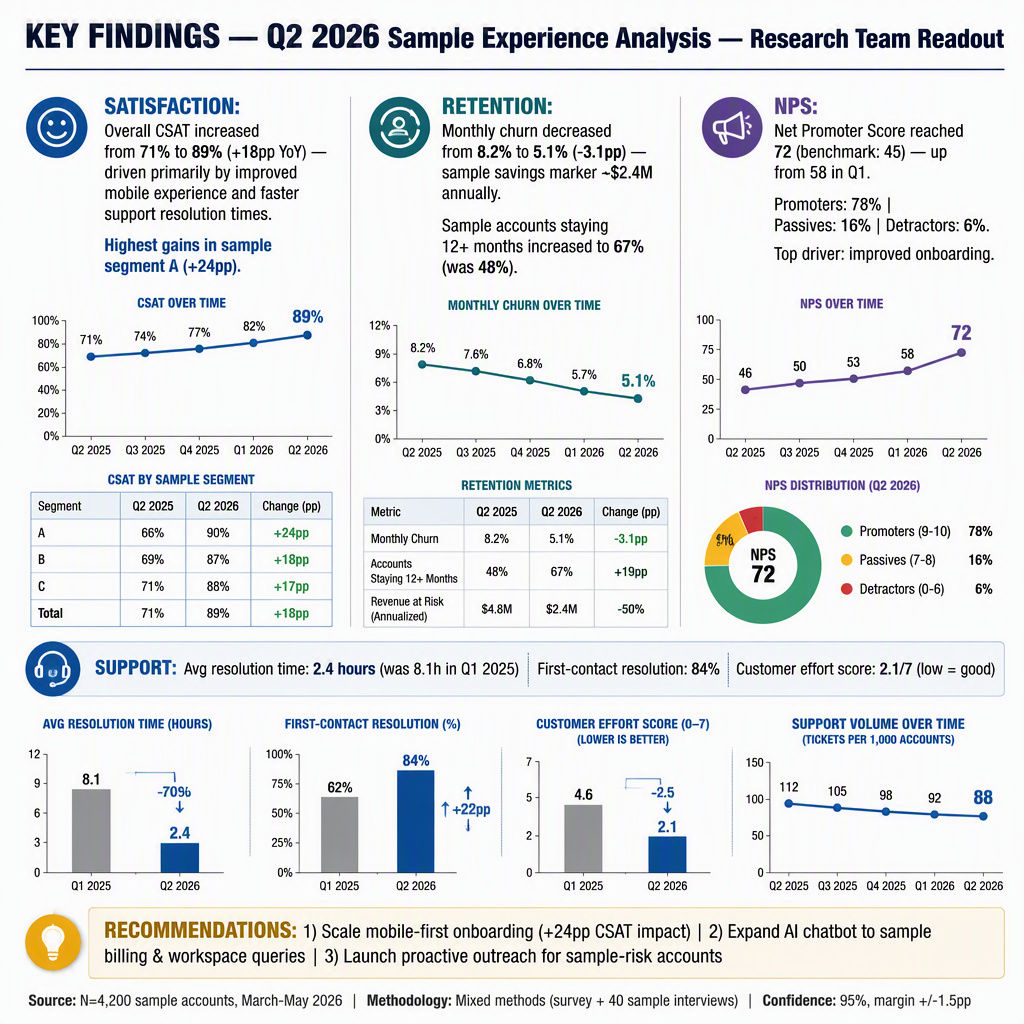} &
    \includegraphics[width=0.235\textwidth]{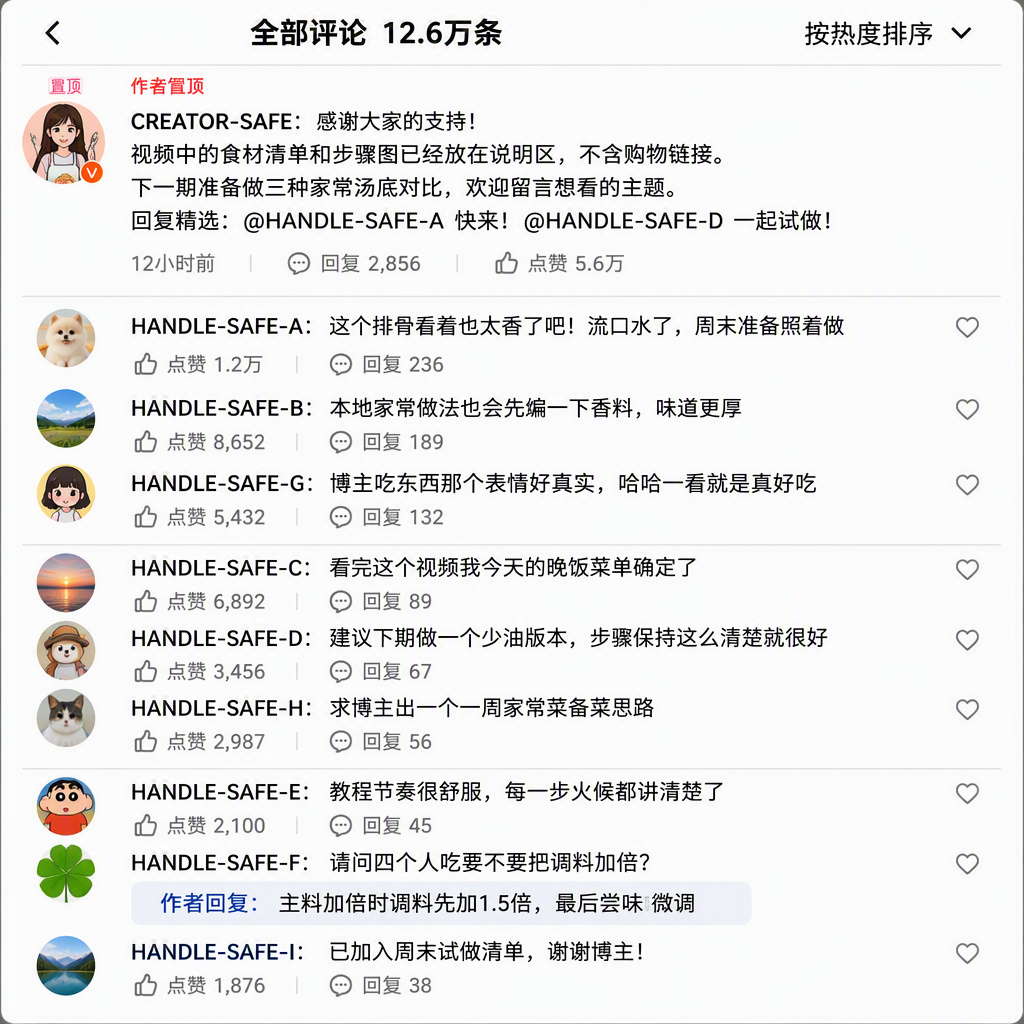} &
    \includegraphics[width=0.235\textwidth]{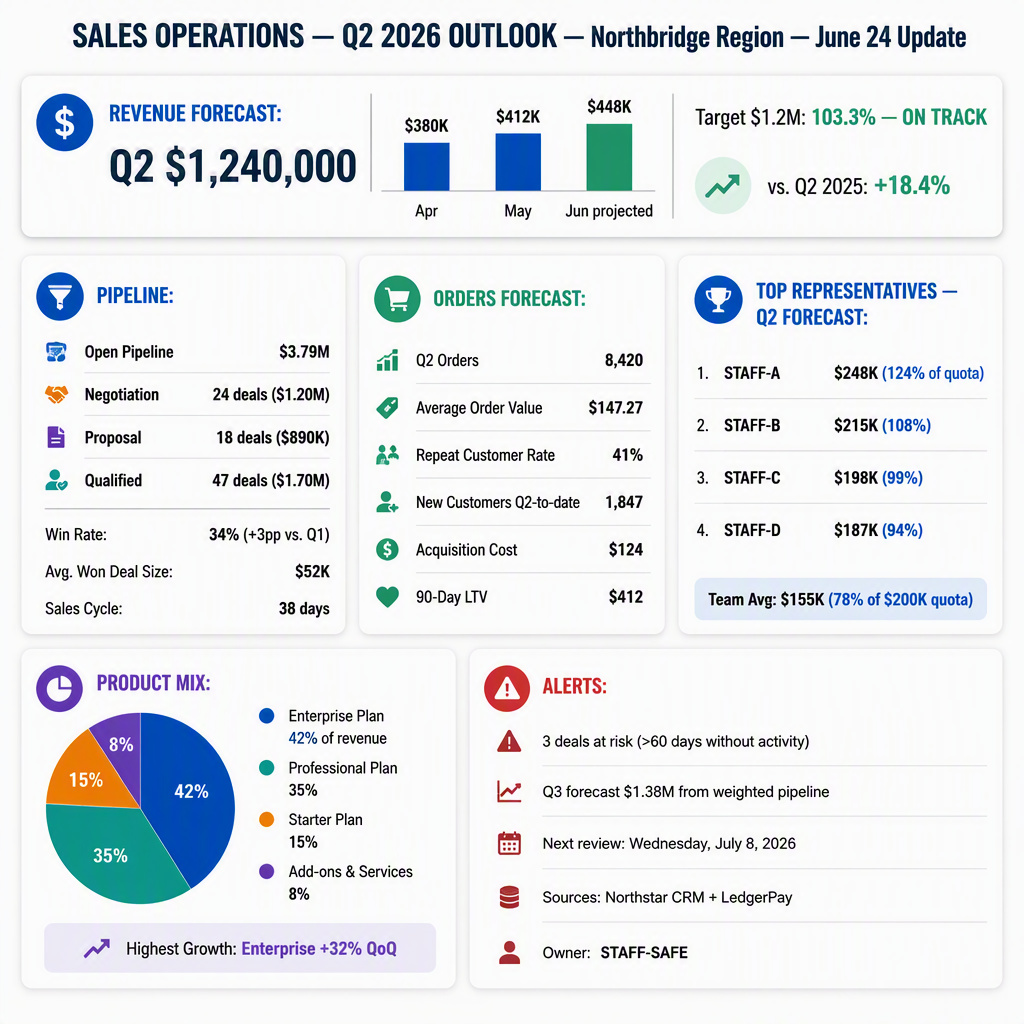} \\
    \scriptsize Webpage (ZH/L2) & \scriptsize Slide (EN/L2) & \scriptsize Social Media (ZH/L2) & \scriptsize Dashboard (EN/L2) \\[1pt]
    \includegraphics[width=0.235\textwidth]{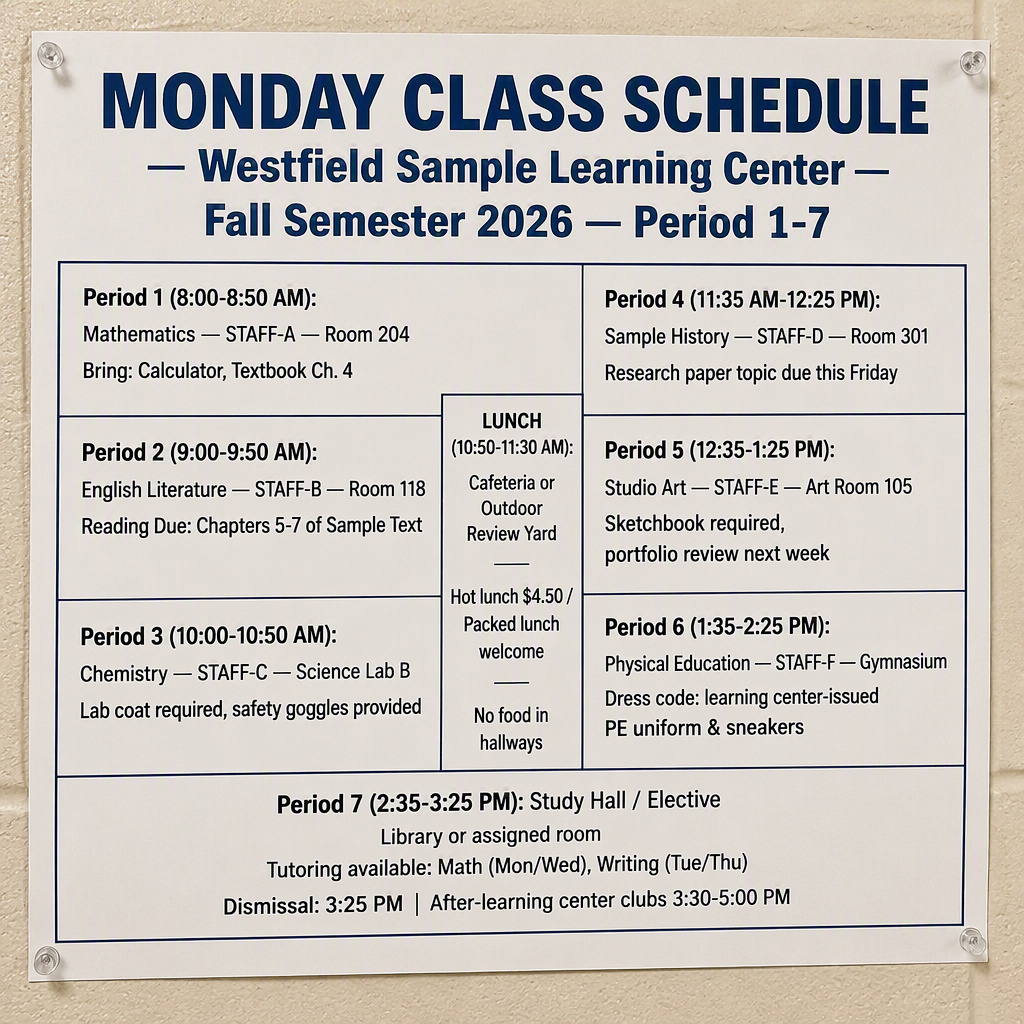} &
    \includegraphics[width=0.235\textwidth]{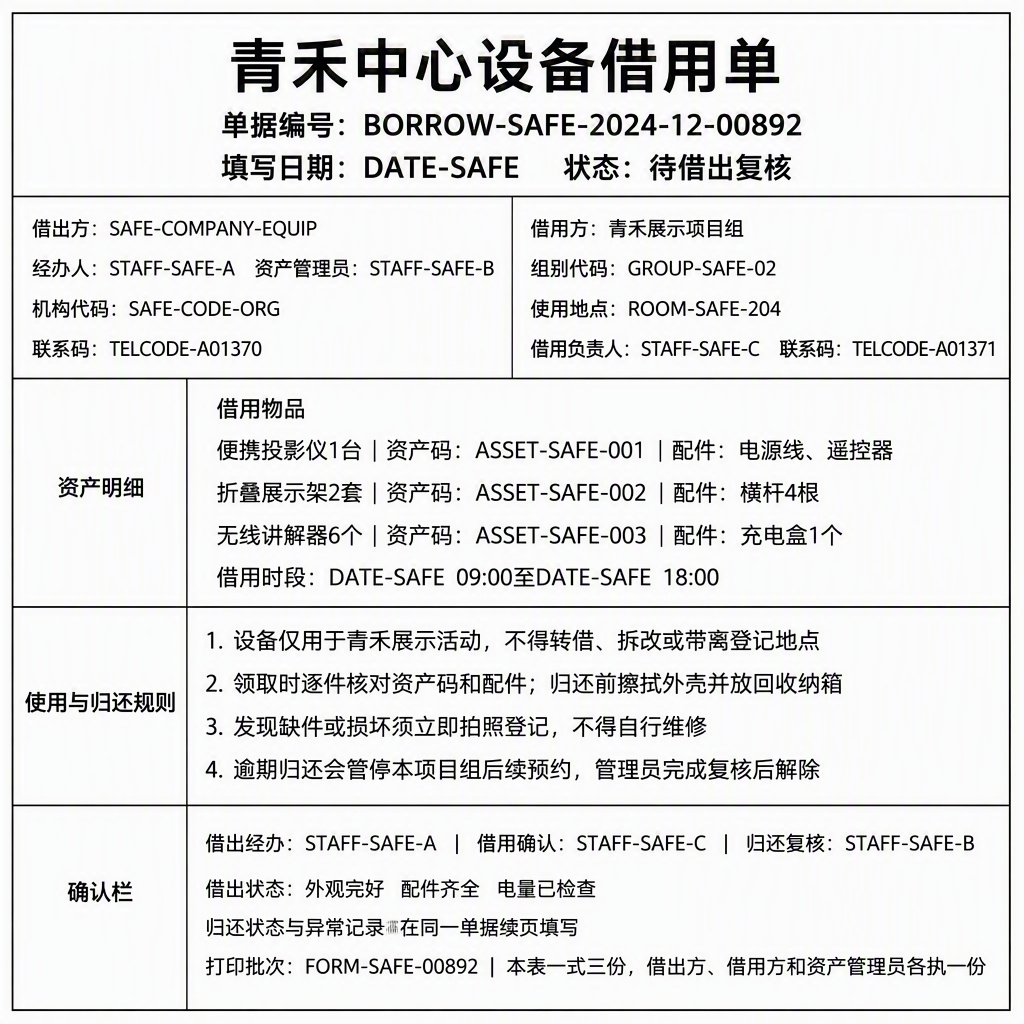} &
    \includegraphics[width=0.235\textwidth]{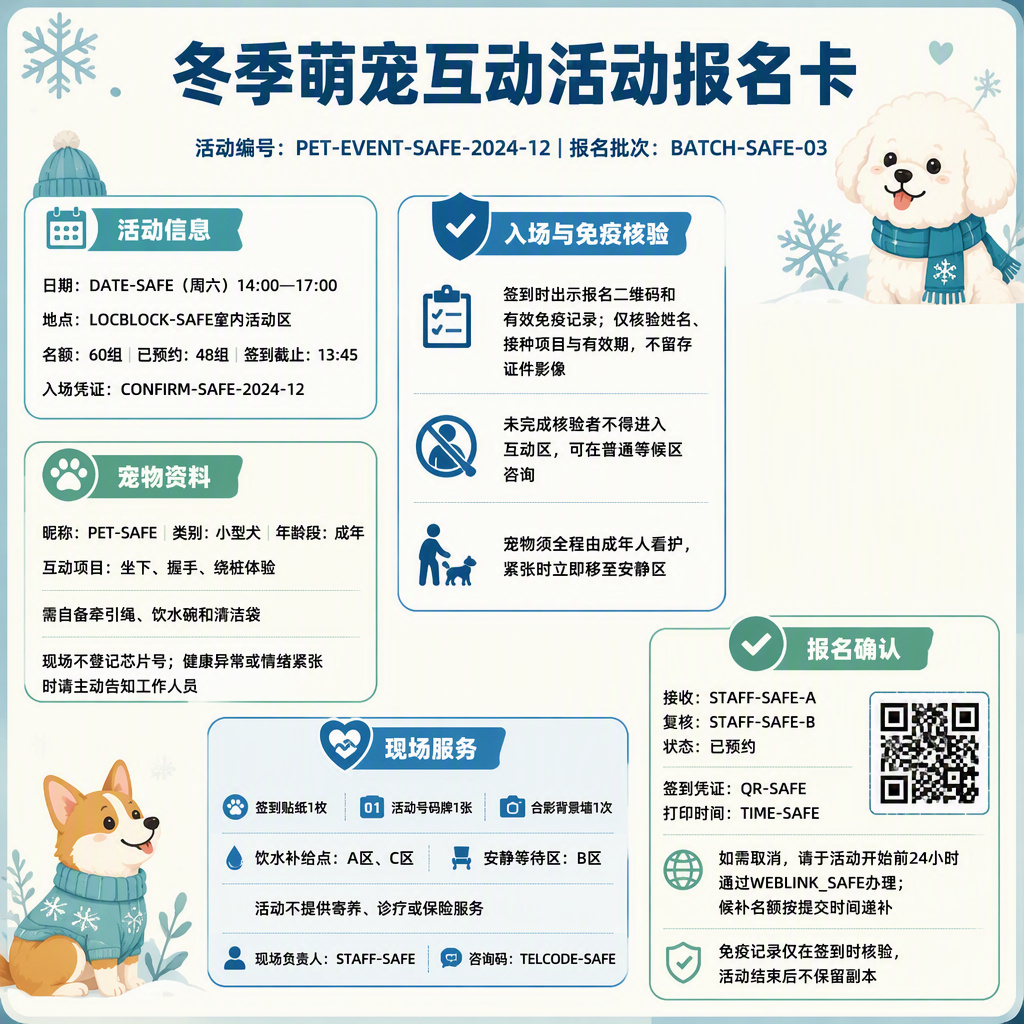} &
    \includegraphics[width=0.235\textwidth]{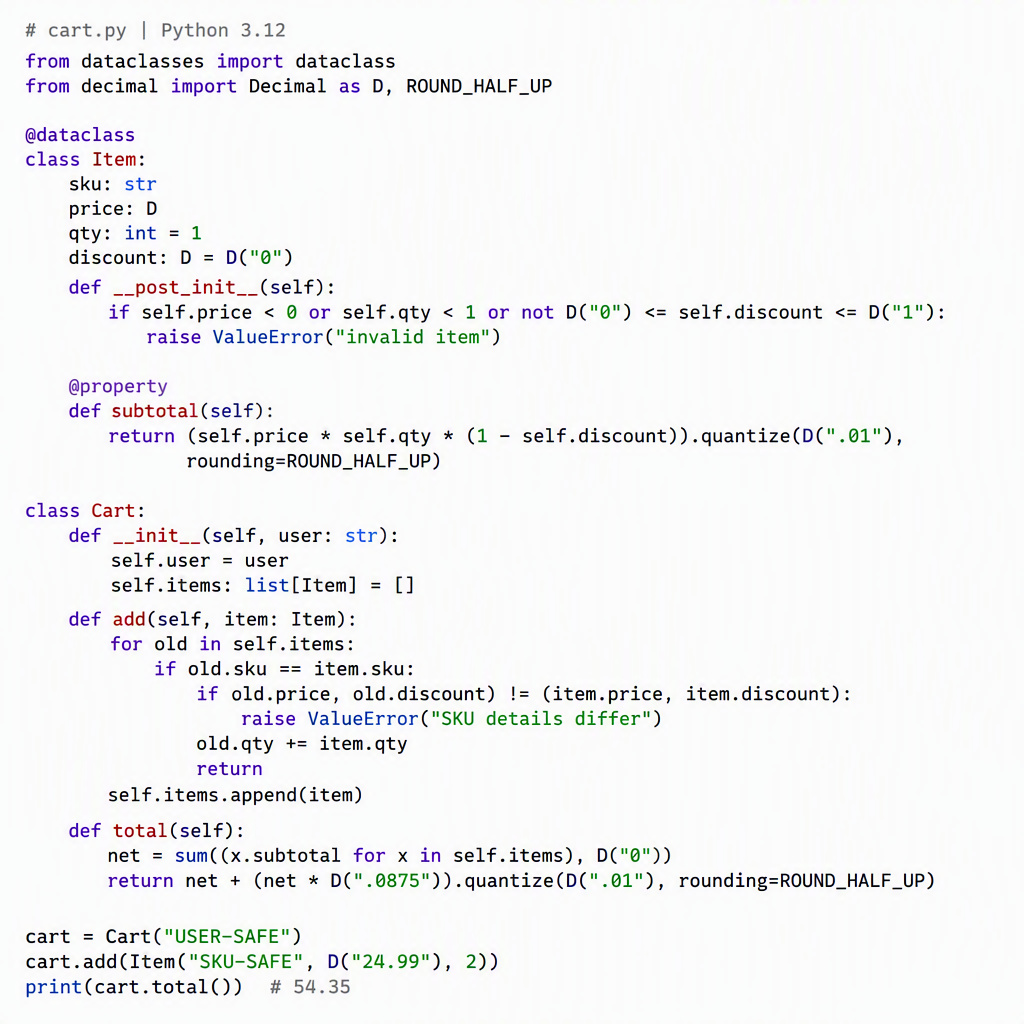} \\
    \scriptsize Schedule (EN/L2) & \scriptsize Form (ZH/L2) & \scriptsize Certificate (ZH/L2) & \scriptsize Code (EN/L2) \\[1pt]
    \includegraphics[width=0.235\textwidth]{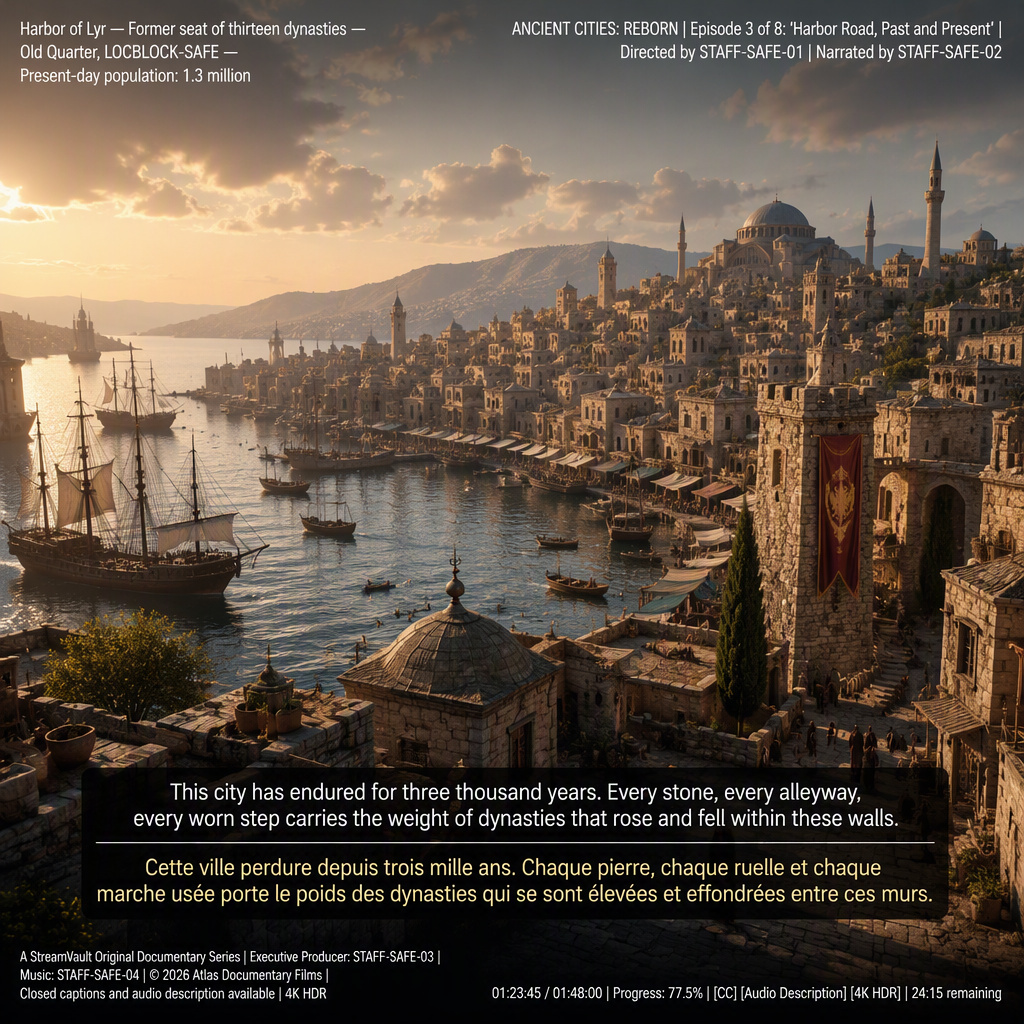} &
    \includegraphics[width=0.235\textwidth]{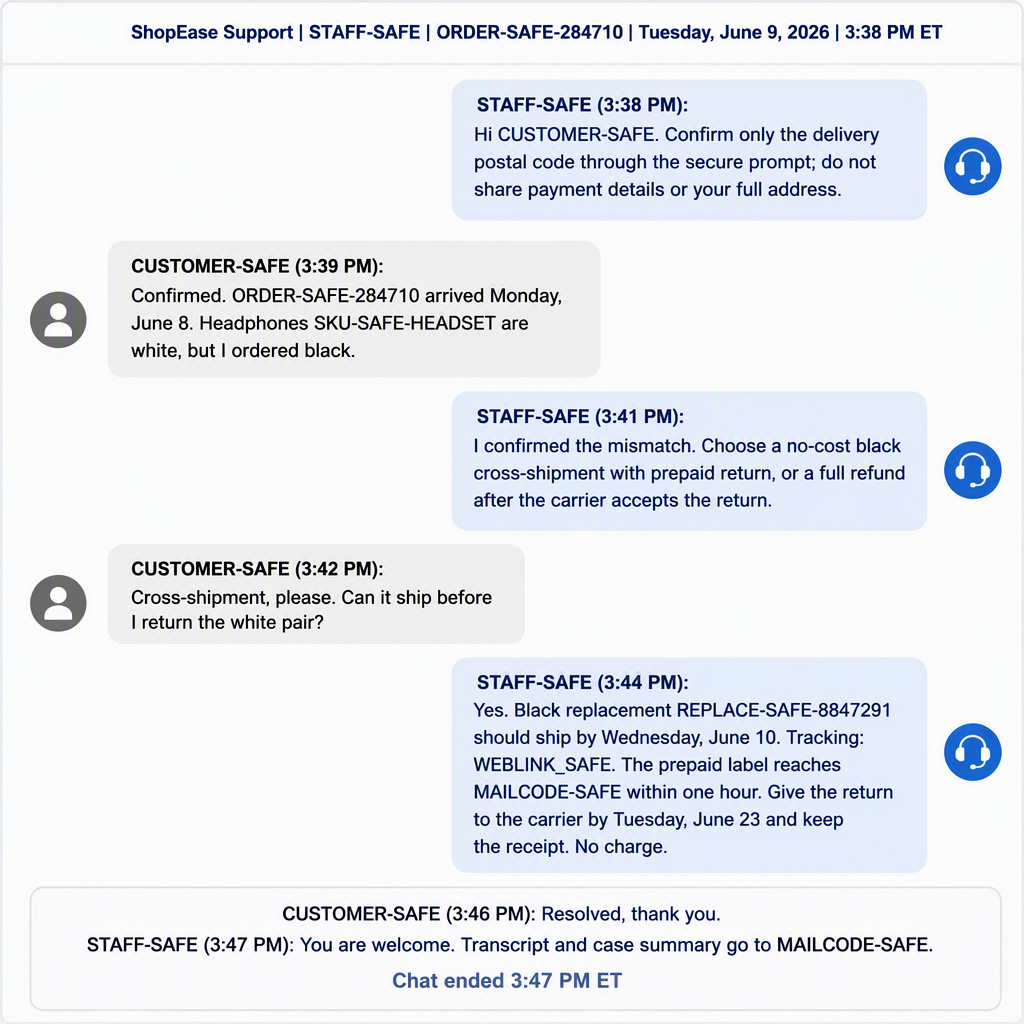} &
    \includegraphics[width=0.235\textwidth]{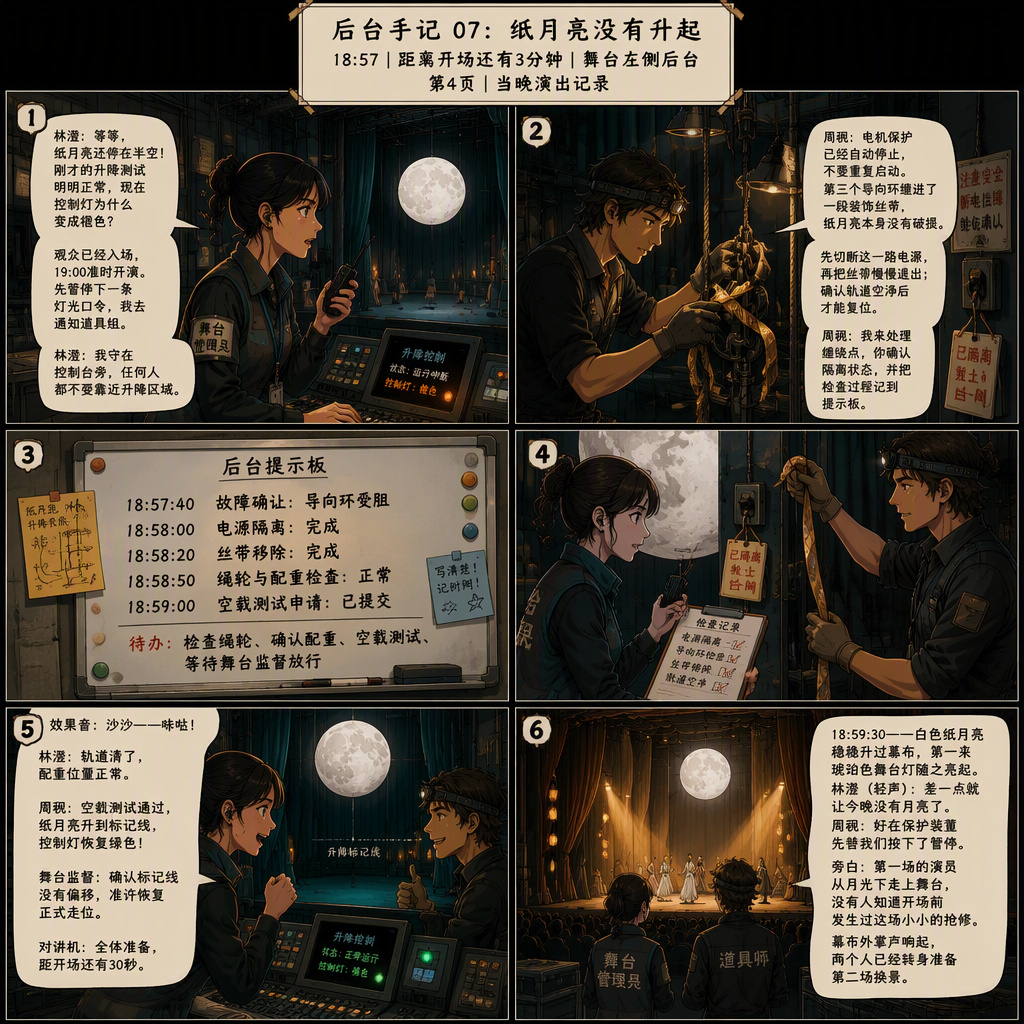} &
    \includegraphics[width=0.235\textwidth]{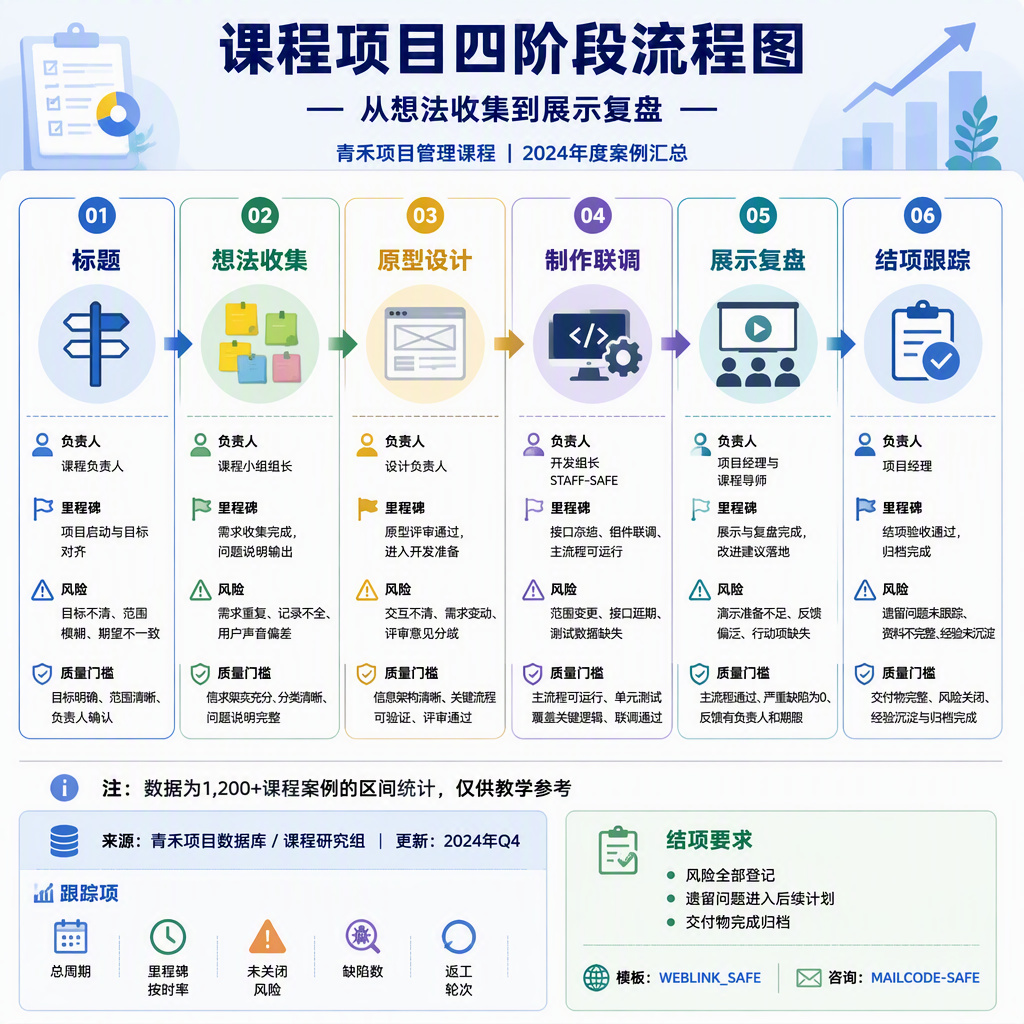} \\
    \scriptsize Caption (EN/L2) & \scriptsize Dialogue (EN/L2) & \scriptsize Comic (ZH/L2) & \scriptsize Infographic (ZH/L2)
  \end{tabular}
  \caption[L2 category gallery]{\textbf{Scene coverage at L2 (Very Hard).} One selected output per category in taxonomy order, from GPT Image 2 at API quality \texttt{Low}. EN/ZH denote English/Chinese. These examples illustrate scene diversity.}
  \label{fig:app_l2_gallery}
\end{figure}

\figref{fig:app_l1_gallery} and \figref{fig:app_l2_gallery} give one selected output per category at L1 and L2; the L3 gallery appears in \figref{fig:prompt_examples} in the main text.
The galleries show the range of scenes in the taxonomy. The enlarged cases above provide the detail needed to inspect individual strings and their carriers.
\clearpage

\clearpage

\FloatBarrier

\clearpage
\section{Released Materials and Their Limits}
\label{app:diagnostic_scope}
\paragraph{Material locations.}
The dataset and the evaluator are released in the benchmark code package; generated images and per-image judge records are released separately.
End-to-end reproduction requires a versioned public archive linking these materials to the evaluated model configurations.

\paragraph{Released evidence.}
The complete EN and ZH prompt records, together with the released evaluator, support
the dataset counts, the region inventories, the six raw metrics, strict response
validation, typed failure states, and the four-dimension composite.
The GPT Image 2 [Low] release adds images and per-image judge rows: 859/864
successful EN images over 215/216 prompts, and 864/864 successful ZH images
over 216/216 prompts. Four missing EN images belong to
\texttt{F3\_L1\_EN\_003} and one to \texttt{A4\_L2\_EN\_003}.
The recorded scores reproduce the bilingual prompt-macro Composite of 99.35.
All six raw dimensions equal 100 for 732/859 valid EN images and 708/864
valid ZH images, or 1,440/1,723 (83.6\%) together. These counts describe
automatic ratings, without assigning correctness labels to the images.
The cases shown in \appref{app:diagnostics} are drawn from these records.

\paragraph{Boundaries of the evidence.}
The GPT Image 2 cases selected here are qualitative examples rather than a
representative sample of error rates.
The released judge summaries identify \texttt{Qwen-Image-Bench}, but their
\texttt{model\_revision} and \texttt{model\_artifact\_sha256} fields are null.
The recorded identity hash does not fix the checkpoint weights. These records
therefore support checking score aggregation and image bindings, with incomplete
checkpoint provenance. The GPT Image 2 records do not establish the coverage or
score distribution of the other model configurations in the leaderboard.

\paragraph{Reporting consequence.}
Reproduction across all configurations requires their model revisions,
resolutions, sampling settings, API quality settings and evaluation dates,
together with images, raw judge responses, and the status of every planned row.
Image and prompt coverage should accompany each quality mean. Comparing only
shared prompts can assess how differences in coverage affect a pair of models;
uncertainty estimates should account for images sharing a prompt. These further
comparisons are not reported here.

The released dataset statistics, including the per-prompt character and region distributions, are reported in \tabref{tab:stats} and are not duplicated here.

\section{Human Alignment}
\label{app:human_alignment}

\paragraph{Participation and scope.}
Ten participants took part in human evaluation of the automatic scores.
The four dimensions assess content, readability, spatial organization, and scene integration (\tabref{tab:dimensions}).
No quantitative inter-rater or human--judge agreement is reported.

\paragraph{Qualitative comparisons.}
The selected cases in \appref{app:diagnostics} pair whole images with native-resolution crops and the relevant target text.
These views support different checks: the crop exposes character shapes, while the whole image shows placement and carrier fit.
The hours plate in \figref{fig:app_diag_position} appears at bottom-left rather than the requested bottom-right; the lettering in \figref{fig:app_diag_integration} follows the wooden sign's plane, lighting, and texture.
The cases also retain disagreements with the automatic scores.
Target-related chat text is visible in \figref{fig:app_diag_layout} despite zero accuracy and completeness ratings, while the code image in \figref{fig:app_diag_high_score} contains degraded small characters despite maximum ratings.
These selected observations do not estimate the frequency of scoring errors or establish the reliability of close leaderboard rankings.

\paragraph{Maximum ratings and their interpretation.}
A score of 100 is the upper end of the judge's rubric, not a measured percentage of correct characters.
Within a subset whose automatic scores are all 100, Pearson and Spearman correlations with those scores are undefined.
Assessing such a subset requires examining the human ratings and the rendered text directly.
A high human quality rating does not establish exact character reproduction, which requires transcription checks.

\end{CJK*}
\end{document}